\documentclass[10pt]{article}
\usepackage{graphicx} 
\usepackage[margin=1in]{geometry} 
\usepackage{url}
\usepackage{pgfplots}
\usepackage{csvsimple}
\usepackage[labelfont=bf,textfont=it]{caption}
\usepackage{subcaption}
\usepackage{tcolorbox}
\usepackage{helvet}
\usepackage{enumitem}
\usepackage{times}
\usepackage{titlesec}
\usepackage{lineno}

\titlespacing*{\paragraph}{0pt}{1.5ex plus 1ex minus .5ex}{1em}

\pgfplotsset{compat=1.11,
    /pgfplots/ybar legend/.style={
        /pgfplots/legend image code/.code={%
            \draw[##1,/tikz/.cd,bar width=6pt,yshift=-0.2em,bar shift=0pt]
            plot coordinates {(0cm,0.8em)};},
    },
}

\title{Geometry Matters: Benchmarking Scientific ML Approaches for Flow Prediction around Complex Geometries}

\author{Ali Rabeh\footnotemark[2], Ethan Herron\footnotemark[2], Aditya Balu, Soumik Sarkar, Chinmay Hegde, \\ Adarsh Krishnamurthy\footnotemark[1], Baskar Ganapathysubramanian\footnotemark[1]\thanks{Corresponding authors. $^{\dagger}$ Equal contribution. AR, EH, AB, SS, AK, and BG are with Iowa State University. CH is with NYU Tandon School of Engineering.}}
\date{}

\usepackage[numbers,sort&compress]{natbib}
\usepackage{multirow} 
\usepackage{array}    
\usepackage{booktabs} 
\usepackage{amsmath}  

\usepackage{tikz}
\usetikzlibrary{shapes, arrows.meta, positioning}

\usepackage{array}
\usepackage[bookmarks=false]{hyperref}
    \hypersetup{colorlinks,
      linkcolor=blue,
      citecolor=blue,
      urlcolor=blue}

\newcommand{\colref}[2]{\hyperref[#2]{#1~\ref*{#2}}}

\newcommand{\figref}[1]{\colref{Figure}{#1}}

\newcommand{\secref}[1]{\colref{Section}{#1}}
\newcommand{\tabref}[1]{\colref{Table}{#1}}
\newcommand{\appendixref}[1]{\colref{Appendix}{#1}}
\newcommand{\coloredref}[2]{\hyperref[#2]{#1~\ref*{#2}}}
\newcommand{\coloredsubref}[3]{\hyperref[#2]{#1~\ref*{#2}{#3}}}

\usepackage{cleveref} 

\begin{document}

\maketitle
\begin{abstract}
\noindent
Rapid and accurate simulations of fluid dynamics around complicated geometric bodies are critical in a variety of engineering and scientific applications, including aerodynamics and biomedical flows. However, while scientific machine learning (SciML) has shown considerable promise, most studies in this field are limited to simple geometries, and complex, real-world scenarios are underexplored. This paper addresses this gap by benchmarking diverse SciML models, including neural operators and vision transformer-based foundation models, for fluid flow prediction over intricate geometries. Using a high-fidelity dataset of steady-state flows across various geometries, we evaluate the impact of geometric representations --- Signed Distance Fields (SDF) and binary masks --- on model accuracy, scalability, and generalization. Central to this effort is the introduction of a novel, unified scoring framework that integrates metrics for global accuracy, boundary layer fidelity, and physical consistency to enable a robust, comparative evaluation of model performance. Our findings demonstrate that newer foundation models significantly outperform neural operators, particularly in data-limited scenarios, and that SDF representations yield superior results with sufficient training data. Despite these promises, all models struggle with out-of-distribution generalization, highlighting a critical challenge for future SciML applications. By advancing both evaluation models and modeling capabilities, our work paves the way for robust and scalable ML solutions for fluid dynamics across complex geometries.
\end{abstract}

\begin{figure}[!t]
    \centering
    \includegraphics[height=0.27\linewidth]{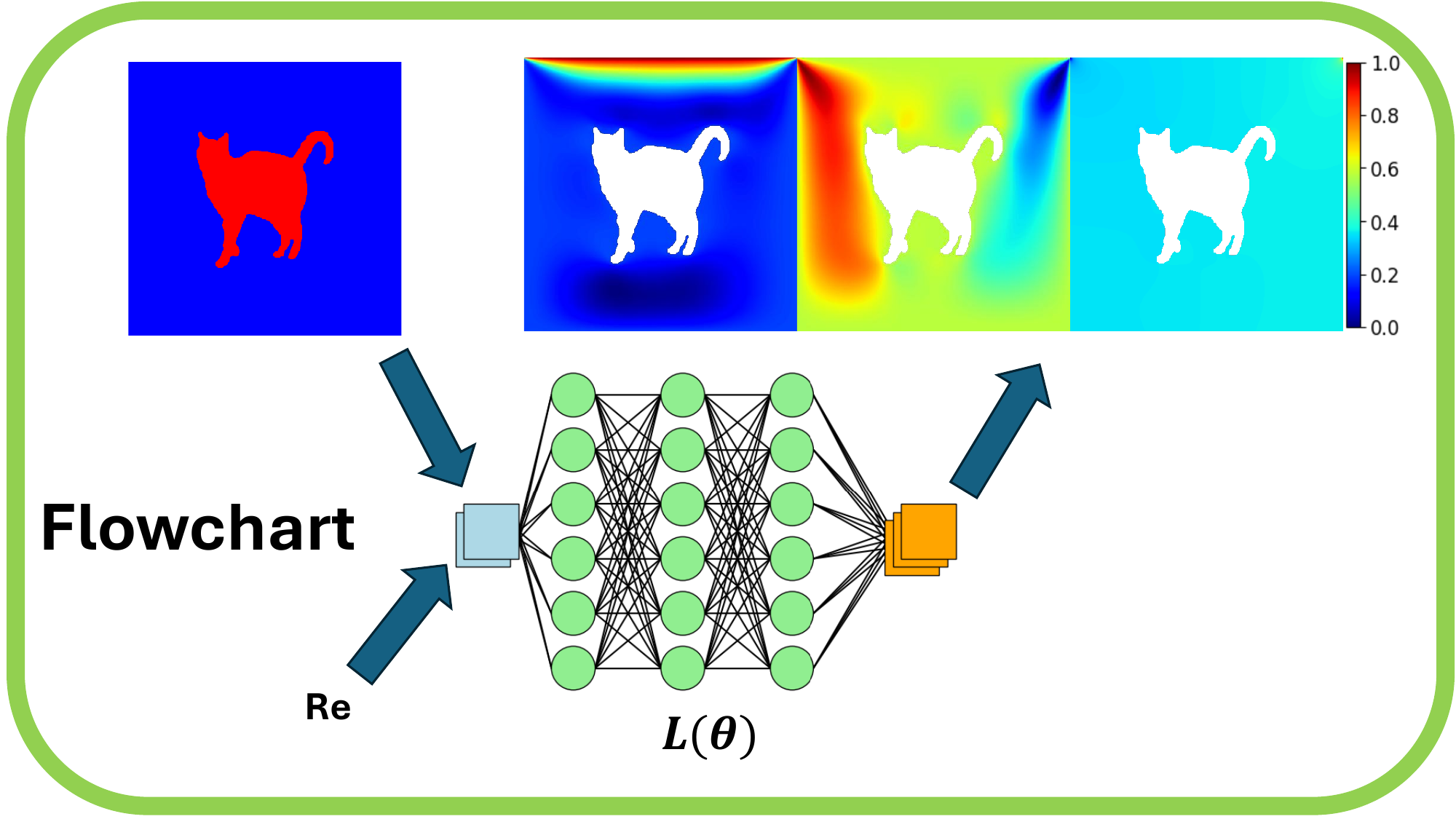}
    \includegraphics[height=0.27\linewidth]{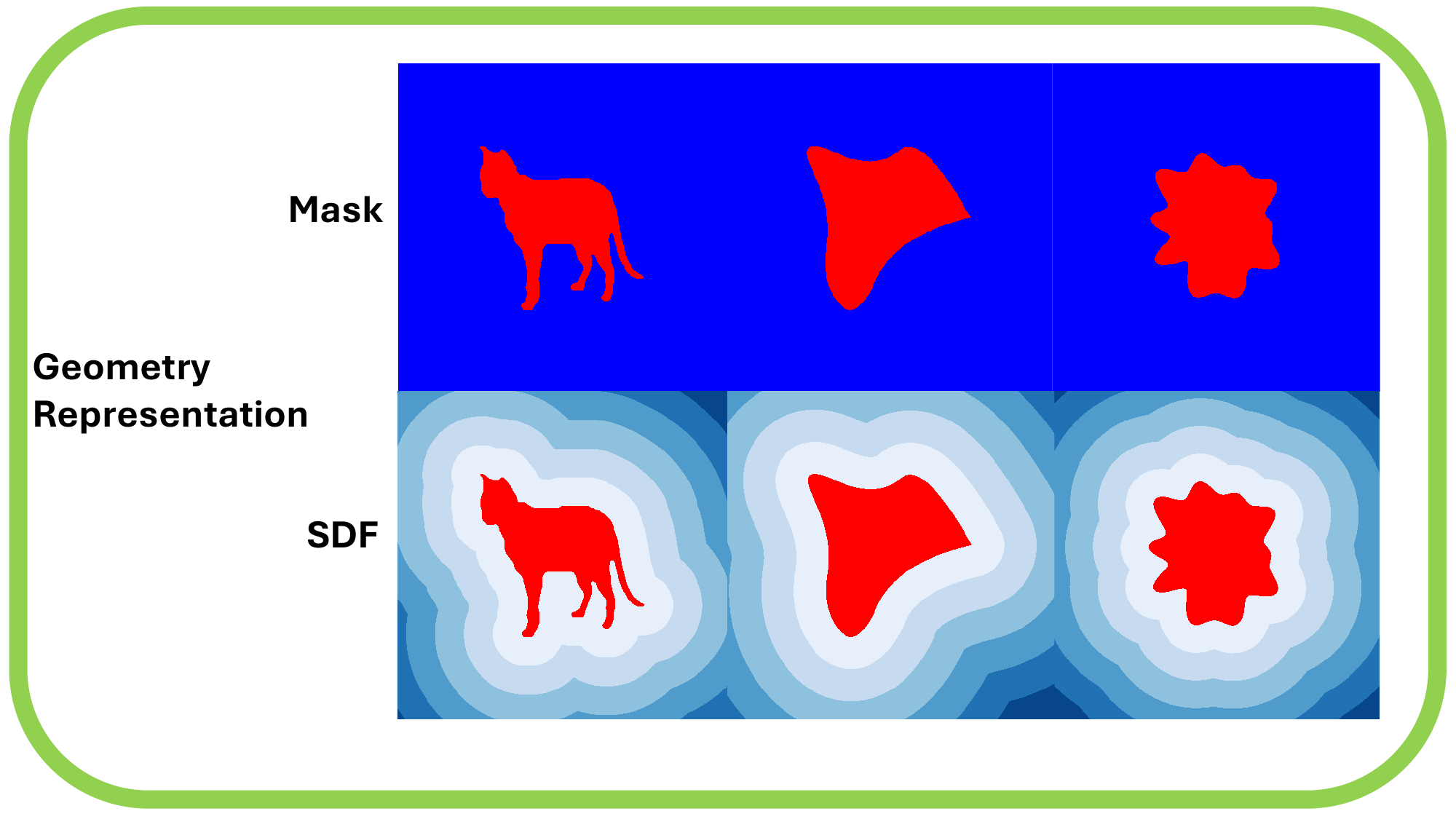}
    \includegraphics[height=0.27\linewidth, trim=0 180pt 0 0, clip]{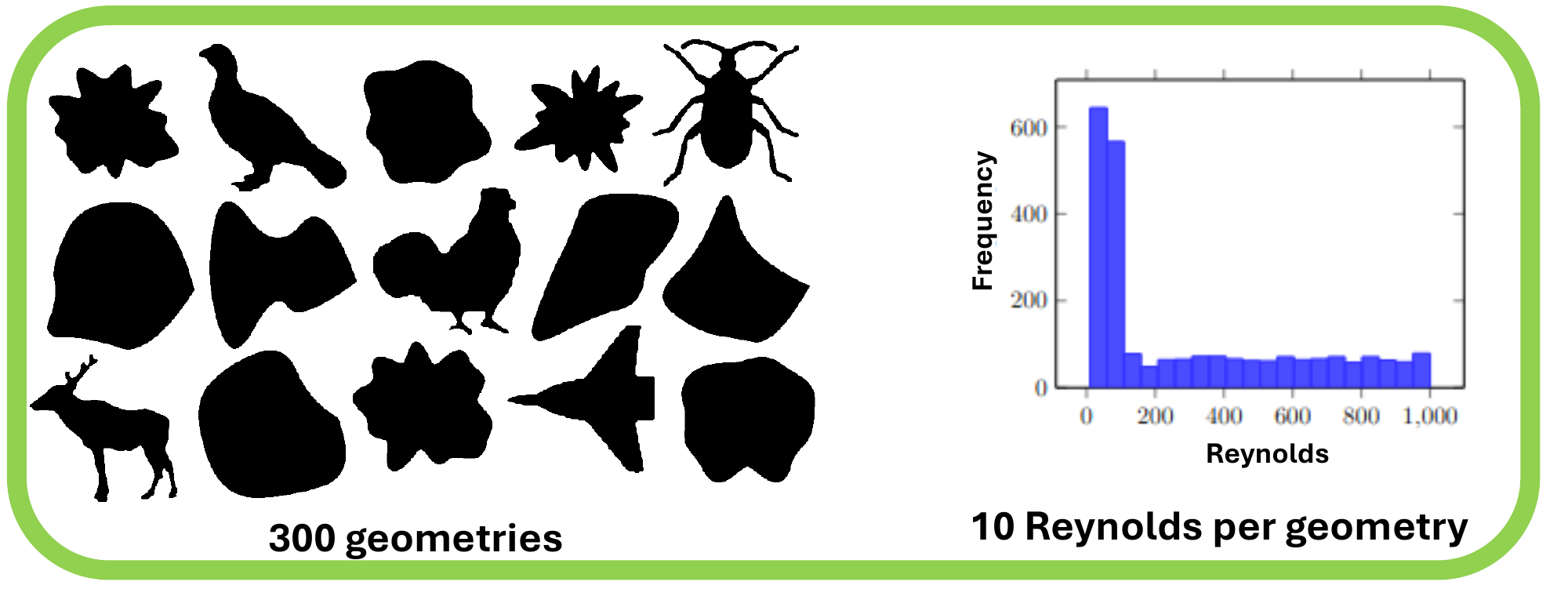}
    \caption{A data-driven evaluation framework for accelerating PDE solvers of fluid flow around complex geometries using scientific machine learning models. This figure illustrates the scientific ML framework that assesses neural operators and foundation models for fluid flow solvers. Flow simulations are performed for steady-state lid-driven cavity flow across various complex geometries and Reynolds numbers. The top left subpanel shows model inputs, including a randomly selected geometry with two inputs: the Reynolds number and a representation of the geometry. It also shows model outputs---x-velocity ($u$), y-velocity ($v$), and the pressure field. The bottom subpanel presents 15 randomly chosen geometries and a Reynolds number distribution bar chart of the training dataset. The top right subpanel contrasts two representations of a sample geometry: the binary mask and the signed distance field (SDF).} \label{fig:overview_sciML}
\end{figure}

\section{Introduction}
Accurate prediction and optimization of flows around complicated geometries are essential across various engineering disciplines, since fluid interactions with intricate shapes profoundly affect real-world behavior. In fields such as aerodynamics, fluid mechanics, and bioengineering, understanding flow patterns enables the analysis of fluid behavior around structures such as aircraft wings~\citep{Agarwal1999}, vehicle bodies~\citep{Iaccarino2005, Sudin2014}, cardiovascular flows~\citep{Balogh2017, Antiga2002}, and architectural designs~\citep{Wallisser2009}. Such insights are critical to advance drag reduction, lift generation, and heat transfer optimization. However, while accurate, traditional computational fluid dynamics (CFD) methods are often slow and computationally expensive, limiting their use in real-time or large-scale applications. This has led to the development of reduced-order models (ROMs) to serve as fast surrogates for expensive CFD simulations and address these limitations. 

However, ROMs are generally only relevant for applications similar to the specific problems for which they were designed. This is due to their reliance on dimensionality reduction techniques to manage complex parameter spaces~\citep{Pant2021}. This lack of generalizability to new simulations or changing parameters often restricts their practical utility. However, scientific machine learning (SciML) offers a promising alternative by leveraging data-driven techniques to model fluid dynamics with both speed and accuracy. The emergence of SciML has been driven by advancements in neural networks, which have demonstrated remarkable capabilities in learning complex, nonlinear mappings from high-dimensional data~\citep{Choudhary2022}. These models enable data-driven discovery, bypassing the need to solve governing equations explicitly while maintaining high levels of prediction accuracy~\citep{Balakrishnan2024}. As a result, SciML is increasingly being adopted in fields that require rapid prototyping~\citep{Kumar2023}, optimization~\citep{Waheed2023}, and quantification of uncertainty of fluid flow phenomena~\citep{Sun2023, Bhattacharjee2023, Psaros2023}.

Recent advances in SciML have enabled the development of models capable of approximating fluid dynamics with remarkable efficiency. These data-driven models can significantly reduce computational costs while maintaining high accuracy, making them attractive for applications where the deployment of traditional solvers is infeasible due to time constraints. However, despite these advances, SciML models have primarily been evaluated on flows over simple geometries~\citep{Bonnet2022, Luo2024, Xu2023}, limiting their relevance to real-world scenarios. Applications involving complex geometries, such as urban wind flows, biomedical fluid dynamics, and turbulent flows around vehicles, remain underexplored in SciML~\citep{Collins2023}. Therefore, it is necessary to carefully benchmark SciML models on challenging datasets involving intricate boundary interactions to uncover their true potential and limitations. Our study seeks to bridge this gap by evaluating the performance of a variety of SciML models in predicting fluid-flow problems using a high-fidelity dataset of steady-state flow, governed by the Navier-Stokes equations~\citep{Tali2024} and focusing on flows around complex geometries.

We focus on two distinct types of geometric representations: binary geometry masks and Signed Distance Fields (SDF). The geometry mask indicates whether a point lies inside or outside an object, effectively isolating the region of interest. In contrast, the SDF offers richer and smoother information by encoding the shortest distance from each point in the simulation domain to the surface of the complex geometry, and distinguishes interior from exterior points~\citep{Yang2024, Zhang2019, Lai2023}. By systematically evaluating these representations, our goal will be to understand how effectively SciML models can capture and simulate flow features around complex geometric objects. A

A promising feature of SciML models is the ability to extrapolate to out-of-sample distributions, a crucial requirement in computational fluid dynamics (CFD)\citep{Li2024unit, Gkimisis2023}. Unlike traditional physics solvers, designed for fixed boundary conditions and parameter settings, SciML models can generalize beyond their training distributions. However, real-world flows often involve variations in geometry, parameters, and boundary conditions, posing significant challenges for generalization~\citep{Kwak2005, Zhu2022}. Accurate predictions in extrapolatory regimes enable SciML models to serve as reliable alternatives to traditional methods, improving the feasibility of modeling unseen flow scenarios~\citep{Subramanian2024, Muckley2023, Goswami2024}. Understanding the limits of SciML models in these regimes is critical for designing architectures and training strategies capable of achieving robust generalization, even under significant distribution shifts.

Another essential consideration is data sufficiency, which refers to the training data required for optimal model performance. Sufficient training data enables models to capture critical flow characteristics correctly~\citep{Papamakarios2021, Shen2018}. However, obtaining large datasets is often computationally and practically challenging. This challenge is further compounded in high-fidelity CFD simulations, where generating datasets involves solving partial differential equations (PDEs) over millions of grid points, requiring significant computational resources~\citep{Rabeh2024}. By investigating the relationship between dataset size and model performance, this study aims to offer actionable insights into how data efficiency can be achieved without sacrificing prediction accuracy.

The FlowBench dataset~\cite{Tali2024} comprises more than 10,000 high-fidelity simulations featuring a variety of complex geometries and diverse flow conditions. This dataset is tailored for the benchmarking of scientific machine learning models, covering tasks associated with complex geometrical configurations and including both 2D and 3D simulations that capture steady and transient fluid dynamics across single and multiphysics scenarios. Specifically, our study employs the 2D Lid-driven Cavity subset from FlowBench to explore the following questions:

 \begin{itemize}[itemsep=0pt,topsep=0pt]
     \item What is the most effective representation of complex geometries in the context of predicting fluid flow around these geometries?
     \item How does training dataset size correlate with performance, and what is the minimum data requirement for SciML models to predict fluid flow around complex geometries?
     \item How do SciML models' accuracy respond to distribution shifts between training and testing datasets, particularly in extrapolatory regimes (of both geometry and Reynolds number)?
 \end{itemize}

To evaluate the performance of SciML models across these tasks, we utilize three metrics: global mean squared error (MSE), near-boundary MSE, and PDE residual~\citep{Khara2024, NeuFENetKhara2024}. These metrics assess the ability of the SciML model to replicate flow fields, capture boundary effects, and adhere to physical laws (PDEs). The inclusion of PDE residual as a metric is particularly appealing, as it directly measures the ability of the model to satisfy governing equations, offering insights into the physical consistency of their predictions. Since it is cognitively challenging to evaluate models across different metrics, we define a single unified score normalized to the range of 0 to 100. This score is calculated based on the logarithmic scale of MSE values, where $\text{MSE}_{\max} = 1$ corresponds to a meaningless prediction (e.g., predicting zero everywhere) and $\text{MSE}_{\min} = 10^{-6}$ reflects the numerical accuracy of the CFD simulations. This scoring system ensures a meaningful comparison by aligning with both the worst-case scenario (score of 0) and the expected precision of the ground truth (score of 100). Further details of the scoring scale are provided in \secref{subsec:metrics}. \secref{sec:experimentation_and_results} describes the dataset, details the SciML models employed, and presents the experimental results and their analysis. \secref{sec:discussion_and_conclusion} summarizes the findings, identifies open questions, and discusses potential directions for future research.

\section{Results}\label{sec:experimentation_and_results}

\subsection{Geometric Representation}\label{subsec:geometric_representation}

\begin{figure}[!t]
    \centering
    \begin{subfigure}[b]{0.25\linewidth}
        \centering
        \includegraphics[width=\linewidth]{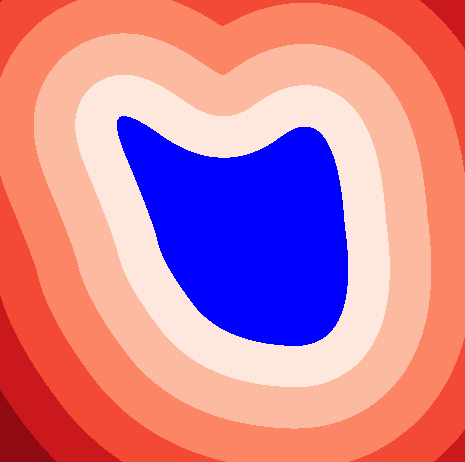}
        \caption{SDF - geometry 1}
        \label{fig:SDF_nurbs}
    \end{subfigure}
    \hspace{0.05\linewidth}
    \begin{subfigure}[b]{0.25\linewidth}
        \centering
        \includegraphics[width=\linewidth]{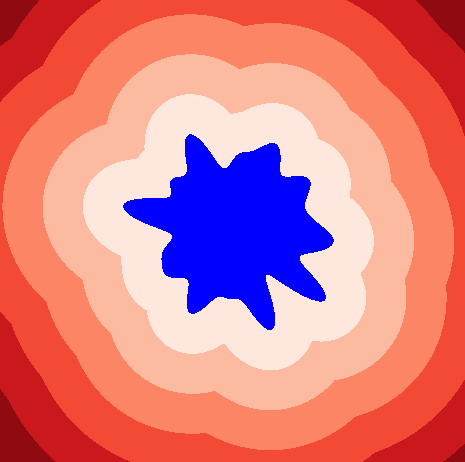}
        \caption{SDF - geometry 2}
        \label{fig:SDF_harmonics}
    \end{subfigure}
    \hspace{0.05\linewidth}
    \begin{subfigure}[b]{0.25\linewidth}
        \centering
        \includegraphics[width=\linewidth]{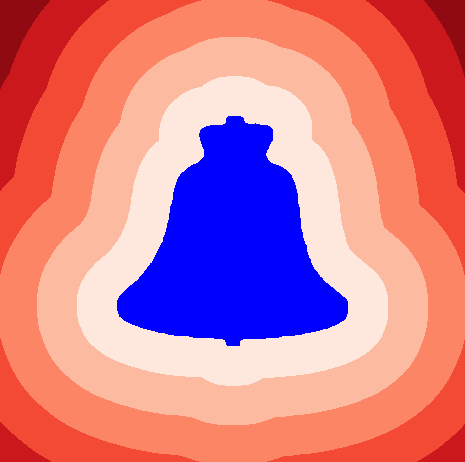}
        \caption{SDF - geometry 3}
        \label{fig:SDF_skelneton}
    \end{subfigure}    
    \begin{subfigure}[b]{0.25\linewidth}
        \centering
        \includegraphics[width=\linewidth]{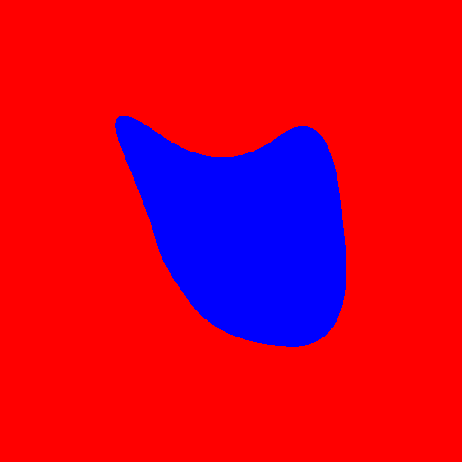}
        \caption{Mask - geometry 1}
        \label{fig:mask_nurbs}
    \end{subfigure}
    \hspace{0.05\linewidth}
    \begin{subfigure}[b]{0.25\linewidth}
        \centering
        \includegraphics[width=\linewidth]{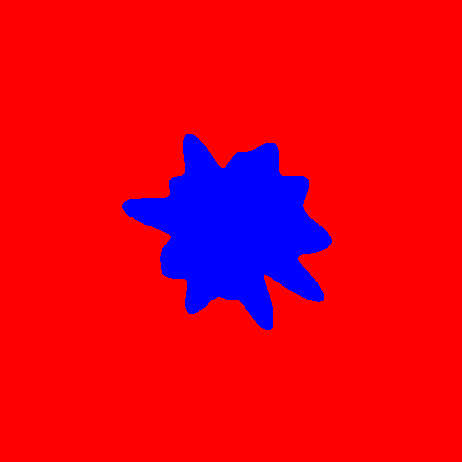}
        \caption{Mask - geometry 2}
        \label{fig:mask_harmonics}
    \end{subfigure}
    \hspace{0.05\linewidth}
    \begin{subfigure}[b]{0.25\linewidth}
        \centering
        \includegraphics[width=\linewidth]{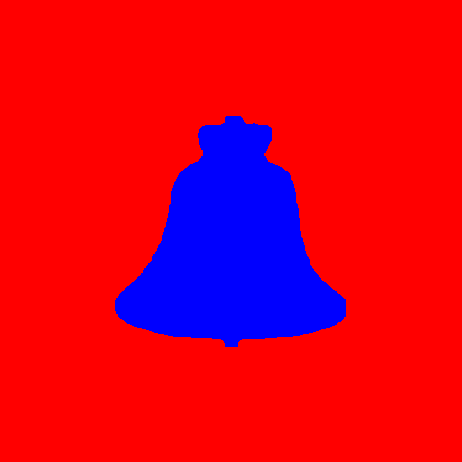}
        \caption{Mask - geometry 3}
        \label{fig:mask_skelneton}
    \end{subfigure}
    \caption{Comparison of geometry representations for different geometries: (a-c) Signed Distance Field (SDF) representations and (d-f) binary mask representations.}
    \label{fig:geometry_comparison}
\end{figure}

We evaluate two geometric representations: SDF and binary masks. SDFs are a scalar field indicating the shortest distance from each point in the prediction domain to the object's boundary. The signed distance field represents the shortest distance from a given point in space to the surface of a geometric shape. It takes negative values inside the object, positive values outside the object, and zero values on the boundary surface. In contrast, the binary mask represents geometry as a binary field, with 0 inside the object and 1 outside, offering a more straightforward, less informative structure regarding relative distances to the boundary layer. An example of the SDF and binary mask for three sample geometries from our dataset is shown in \figref{fig:geometry_comparison}. We aim to assess whether there is added value in using a continuous representation of distance from object boundary versus a simple binary mask on capturing fluid behavior around objects.

To quantitatively evaluate the impact of geometric representation on the performance of the SciML model, we utilize a unified scoring scale. This scale presents error metrics in an understandable range between 0 and 100. A score of 0 indicates the least favorable outcome, where the models predict zero across all fields, and a score of 100 corresponds to model predictions that align with the high precision of computational fluid dynamics (CFD) simulations. More details on this logarithmic scale scoring methodology are presented in \secref{subsec:metrics}.

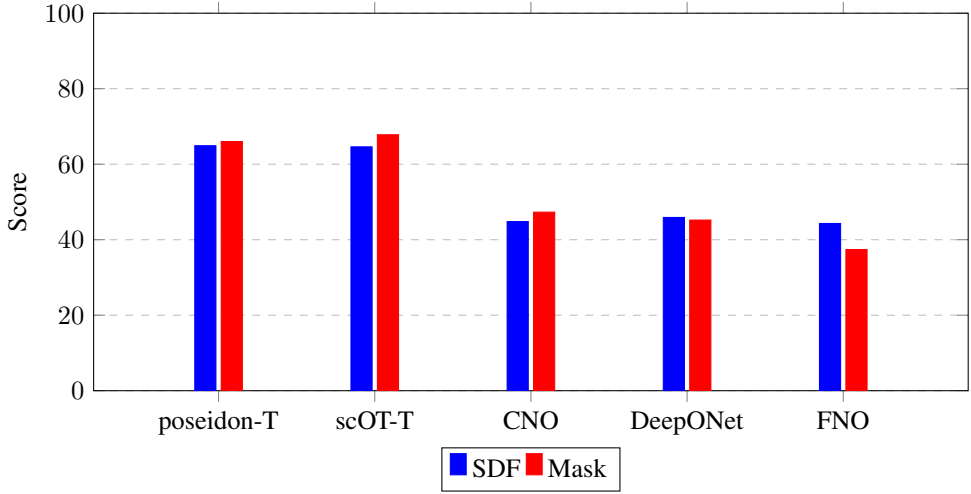
\begin{figure}[t!]
    \centering
    \begin{tikzpicture}
        \begin{axis}[
            ybar,
            bar width=8pt,
            width=0.8\linewidth,
            height=0.4\linewidth,
            enlarge x limits=0.2,  
            symbolic x coords={poseidon-T, scOT-T, CNO, DeepONet, FNO},
            xtick=data,
            ylabel={\text{Score}},
            legend style={at={(0.5,-0.15)},anchor=north,legend columns=-1},
            ymin= 0,
            ymax=100,
            log origin=infty,  
            ymajorgrids=true,
            grid style=dashed
        ]
            \addplot+[ybar, fill=blue] coordinates {(poseidon-T,64.9) (scOT-T,64.6) (CNO,44.8) (DeepONet,45.9) (FNO,44.3)};
            \addplot+[ybar, fill=red] coordinates {(poseidon-T,66.0) (scOT-T,67.8) (CNO,47.3) (DeepONet,45.2) (FNO,37.4)};
            
            \legend{SDF, Mask}
        \end{axis}
    \end{tikzpicture}
    \caption{Comparison of score values for different models using Signed Distance Field (SDF) and binary mask representations. The bar plot shows the score for each model, indicating the performance difference between SDF and mask representations.}
    \label{fig:log_error_SDF-vs-mask}
\end{figure}

We assess the effect of geometry representation on SciML prediction error. The dataset, consisting of 3000 samples, is randomly divided into an 80-20 train/test split. The test dataset, containing 600 samples, is held constant across all experiments to ensure consistent evaluation of model performance. As shown in \figref{fig:log_error_SDF-vs-mask}, scOT-T, poseidon-T, and CNO achieve higher scores with the mask representation, while other neural operators tend to perform better with the SDF. Additionally, \Cref{tab:mse-512-sdf,tab:mse-512-mask} shows that scOT and Poseidon models outperform the other neural operators by roughly an order of magnitude. Sample field predictions and error comparisons between SDF and mask representations for the velocity in the y-direction of a random test sample are displayed in \figref{fig:sdf-vs-mask} in \appendixref{sec:field-predictions}. The error plot indicates that the mask representation produces lower error for the poseidon-T and CNO models, whereas the SDF yields lower error for the geometric-DeepONet model. This difference suggests that scOT, Poseidon, and CNO models benefit from the sharpness of the binary mask, while other neural operators perform better when using the continuous boundary information provided by the SDF. This observation is not intuitive as SDF is a richer field that provides information on how close the object boundary is versus simple in/out information through the binary mask.

\begin{tcolorbox}[colframe=white!70!black, colback=blue!10, coltitle=black, left=6pt, title=Key Takeaways on the Choice of Geometric Representations]
\textbf{Impact on Model Performance:} Vision transformer-based models such as scOT-T and poseidon-T, along with the CNO model, exhibit improved accuracy when using binary mask representations, while other neural operators perform better using the SDF.

\textbf{Performance Comparison:} scOT and Poseidon models outperform other neural operators scoring 20 points higher in performance metrics (order of magnitude lower MSE).
\end{tcolorbox}

\begin{table}[!t] 
\centering
\small
\setlength\extrarowheight{2pt}
\caption{The score of SciML models trained on the full dataset using the signed distance field at two different difficulty levels (random and extrapolatory). All errors are reported on the testing dataset.}
\label{tab:mse-512-sdf}
\begin{tabular}{c c c c c c c}
\hline
 & \multicolumn{3}{c}{\textbf{Random}} & \multicolumn{3}{c}{\textbf{Extrapolatory}} \\ \hline
\textbf{Model} & \textbf{M1} & \textbf{M2} & \textbf{M3} & \textbf{M1} & \textbf{M2} & \textbf{M3} \\ \hline
\textbf{poseidon-L} & $55.9$ & $65.2$ & $24.1$ & $22.5$ & $41.6$ & $0.0$ \\ \hline
\textbf{poseidon-B} & $58.7$ & $69.4$ & $23.6$ & $26.5$ & $41.2$ & $28.2$ \\ \hline
\textbf{poseidon-T} & $\mathbf{64.9}$ & $\mathbf{73.3}$ & $24.2$ & $27.0$ & $\mathbf{42.2}$ & $26.6$ \\ \hline
\textbf{scOT-L} & $60.0$ & $68.6$ & $23.6$ & $21.9$ & $38.8$ & $29.5$ \\ \hline
\textbf{scOT-B} & $58.3$ & $67.0$ & $24.1$ & $23.6$ & $38.4$ & $28.1$ \\ \hline
\textbf{scOT-T} & $64.6$ & $71.4$ & $23.5$ & $23.7$ & $38.9$ & $29.3$ \\ \hline
\textbf{CNO} & $44.8$ & $54.5$ & $28.2$ & $24.1$ & $36.4$ & $24.9$ \\ \hline
\textbf{FNO} & $44.3$ & $59.2$ & $20.4$ & $18.2$ & $32.2$ & $45.0$ \\ \hline
\textbf{WNO} & $24.1$ & $41.3$ & $27.7$ & $13.6$ & $28.4$ & $7.7$ \\ \hline
\textbf{Deeponet} & $45.9$ & $53.0$ & $\mathbf{33.5}$ & $\mathbf{28.0}$ & $35.6$ & $27.7$ \\ \hline
\textbf{geometric-deeponet} & $53.0$ & $59.9$ & $30.2$ & $25.0$ & $37.4$ & $\mathbf{30.8}$ \\ \hline
\end{tabular}
\end{table}

\begin{table}[!t]
\centering
\small
\setlength\extrarowheight{2pt}
\caption{The score of SciML models trained on the full dataset using the binary mask at two different difficulty levels (random and extrapolatory). All errors are reported on the testing dataset.}
\label{tab:mse-512-mask}
\begin{tabular}{c c c c c c c}
\hline
 & \multicolumn{3}{c}{\textbf{Random}} & \multicolumn{3}{c}{\textbf{Extrapolatory}} \\ \hline
\textbf{Model} & \textbf{M1} & \textbf{M2} & \textbf{M3} & \textbf{M1} & \textbf{M2} & \textbf{M3} \\ \hline
\textbf{poseidon-L} & $65.6$ & $74.7$ & $24.1$ & $20.8$ & $36.3$ & $26.4$ \\ \hline
\textbf{poseidon-B} & $63.0$ & $73.1$ & $23.7$ & $25.1$ & $\mathbf{37.9}$ & $28.4$ \\ \hline
\textbf{poseidon-T} & $66.0$ & $\mathbf{76.1}$ & $23.5$ & $23.4$ & $37.1$ & $29.8$ \\ \hline
\textbf{scOT-L} & $67.1$ & $75.6$ & $24.0$ & $20.1$ & $34.7$ & $32.1$ \\ \hline
\textbf{scOT-B} & $62.0$ & $71.8$ & $24.2$ & $20.4$ & $33.8$ & $31.5$ \\ \hline
\textbf{scOT-T} & $\mathbf{67.8}$ & $75.1$ & $24.3$ & $23.1$ & $36.8$ & $31.5$ \\ \hline
\textbf{CNO} & $47.3$ & $59.8$ & $27.1$ & $\mathbf{26.5}$ & $\mathbf{37.9}$ & $28.4$ \\ \hline
\textbf{FNO} & $37.4$ & $58.5$ & $25.6$ & $15.6$ & $29.6$ & $39.2$ \\ \hline
\textbf{WNO} & $24.8$ & $41.2$ & $\mathbf{29.5}$ & $11.6$ & $25.0$ & $\mathbf{40.9}$ \\ \hline
\textbf{Deeponet} & $45.2$ & $53.6$ & $28.7$ & $20.7$ & $32.4$ & $34.5$ \\ \hline
\textbf{geometric-deeponet} & $47.4$ & $54.8$ & $28.6$ & $20.9$ & $34.5$ & $35.1$ \\ \hline
\end{tabular}
\end{table}

\subsection{Data Sufficiency} \label{subsec:data_sufficiency}

Evaluating the impact of training dataset size is critical for understanding the practical feasibility of deploying SciML models, especially in scenarios where generating large datasets is computationally expensive or time-intensive. To investigate the role of training dataset size in the performance of SciML models, we conduct a series of experiments using subsets of the FlowBench dataset. The baseline experiment uses the full training dataset of 2,400 samples, representing the complete FlowBench dataset. Four additional experiments are performed with subsets of 1,200, 800, 400, and 240 samples, keeping the test dataset constant at 600 samples for consistency. These experiments address key questions regarding data sufficiency: How much data is required to achieve reasonable performance? Are there general trends in data requirements across the SciML models? Do certain models demonstrate greater data efficiency than others, and does the choice of geometry representation (e.g., SDF vs. mask) influence these trends?

\begin{figure}[!t]
    \centering
    \begin{subfigure}[b]{0.49\linewidth}
        \centering
        \begin{tikzpicture}
            \begin{axis}[
                width=0.9\linewidth,
                height=0.72\linewidth,
                xlabel={Training Size},
                ylabel={Score},
                legend style={at={(0.99,0.99)}, anchor=north east, draw, fill=white, font=\tiny},
                xtick={2400, 1200, 800, 400, 240},
                xticklabels={2400, 1200, 800, 400, 240},
                ymin=0, ymax=100,
                xticklabel style={rotate=45, font=\footnotesize},
                grid=none
            ]
            \addplot[color=blue, mark=triangle, line width=1pt] coordinates {
                (2400, 44.8) (1200, 36.5) (800, 37.8) (400, 36.4) (240, 32.7)
            };
            \addlegendentry{CNO}
            \addplot[color=purple, mark=diamond, line width=1pt] coordinates {
                (2400, 44.3) (1200, 37.6) (800, 34.2) (400, 32.4) (240, 32)
            };
            \addlegendentry{FNO}
            \addplot[color=green, mark=*, line width=1pt] coordinates {
                (2400, 53) (1200, 47.1) (800, 45.2) (400, 40.3) (240, 34.9)
            };
            \addlegendentry{Geo-DeepONet}
            \end{axis}
        \end{tikzpicture}
        \caption{Neural Operators (SDF)}
    \end{subfigure}
    \hfill
    \begin{subfigure}[b]{0.49\linewidth}
        \centering
        \begin{tikzpicture}
            \begin{axis}[
                width=0.9\linewidth,
                height=0.72\linewidth,
                xlabel={Training Size},
                ylabel={Score},
                legend style={at={(0.99,0.99)}, anchor=north east, draw, fill=white, font=\tiny},
                xtick={2400, 1200, 800, 400, 240},
                xticklabels={2400, 1200, 800, 400, 240},
                ymin=0, ymax=100,
                xticklabel style={rotate=45, font=\footnotesize},
                grid=none
            ]
            \addplot[color=orange, mark=o, line width=1pt] coordinates {
                (2400, 64.9) (1200, 58.6) (800, 56.3) (400, 43.5) (240, 43.8)
            };
            \addlegendentry{poseidon-T}
            \addplot[color=red, mark=square, line width=1pt] coordinates {
                (2400, 58.3) (1200, 48.6) (800, 50.8) (400, 45.7) (240, 43.1)
            };
            \addlegendentry{scOT-B}
            \addplot[color=pink, mark=square*, line width=1pt] coordinates {
                (2400, 64.6) (1200, 56.5) (800, 54.4) (400, 45.1) (240, 42.7)
            };
            \addlegendentry{scOT-T}
            \end{axis}
        \end{tikzpicture}
        \caption{Foundation Models (SDF)}
    \end{subfigure}
    
    \begin{subfigure}[b]{0.49\linewidth}
        \centering
        \begin{tikzpicture}
            \begin{axis}[
                width=0.9\linewidth,
                height=0.72\linewidth,
                xlabel={Training Size},
                ylabel={Score},
                legend style={at={(0.99,0.99)}, anchor=north east, draw, fill=white, font=\tiny},
                xtick={2400, 1200, 800, 400, 240},
                xticklabels={2400, 1200, 800, 400, 240},
                ymin=0, ymax=100,
                xticklabel style={rotate=45, font=\footnotesize},
                grid=none
            ]
            \addplot[color=blue, mark=triangle, line width=1pt] coordinates {
                (2400, 47.3) (1200, 40.0) (800, 42.5) (400, 32.7) (240, 30.6)
            };
            \addlegendentry{CNO}
            \addplot[color=purple, mark=diamond, line width=1pt] coordinates {
                (2400, 37.4) (1200, 34.1) (800, 32.3) (400, 30.9) (240, 29.8)
            };
            \addlegendentry{FNO}
            \addplot[color=green, mark=*, line width=1pt] coordinates {
                (2400, 47.4) (1200, 45.2) (800, 45.5) (400, 40) (240, 38.6)
            };
            \addlegendentry{Geo-DeepONet}
            \end{axis}
        \end{tikzpicture}
        \caption{Neural Operators (Binary Mask)}
    \end{subfigure}
    \hfill
    \begin{subfigure}[b]{0.49\linewidth}
        \centering
        \begin{tikzpicture}
            \begin{axis}[
                width=0.9\linewidth,
                height=0.72\linewidth,
                xlabel={Training Size},
                ylabel={Score},
                legend style={at={(0.99,0.99)}, anchor=north east, draw, fill=white, font=\tiny},
                xtick={2400, 1200, 800, 400, 240},
                xticklabels={2400, 1200, 800, 400, 240},
                ymin=0, ymax=100,
                xticklabel style={rotate=45, font=\footnotesize},
                grid=none
            ]
            \addplot[color=orange, mark=o, line width=1pt] coordinates {
                (2400, 66) (1200, 65.7) (800, 55.6) (400, 54.7) (240, 50.8)
            };
            \addlegendentry{poseidon-T}
            \addplot[color=red, mark=square, line width=1pt] coordinates {
                (2400, 62) (1200, 57) (800, 53.1) (400, 47.8) (240, 46.4)
            };
            \addlegendentry{scOT-B}
            \addplot[color=pink, mark=square*, line width=1pt] coordinates {
                (2400, 67.8) (1200, 63.7) (800, 52.3) (400, 45.1) (240, 38.7)
            };
            \addlegendentry{scOT-T}
            \end{axis}
        \end{tikzpicture}
        \caption{Foundation Models (Binary Mask)}
    \end{subfigure}
    
    \caption{Comparison of score values vs. sample size for different scientific machine learning models using SDF and binary mask representations. The top row compares neural operators (CNO, FNO, Geo-DeepONet), while the bottom row compares foundation models (scOT-B, scOT-T, poseidon-T) across varying sample sizes (2400, 1200, 800, 400, and 240).}
    \label{fig:MSE_vs_Sample_Size_Comparison}
\end{figure}
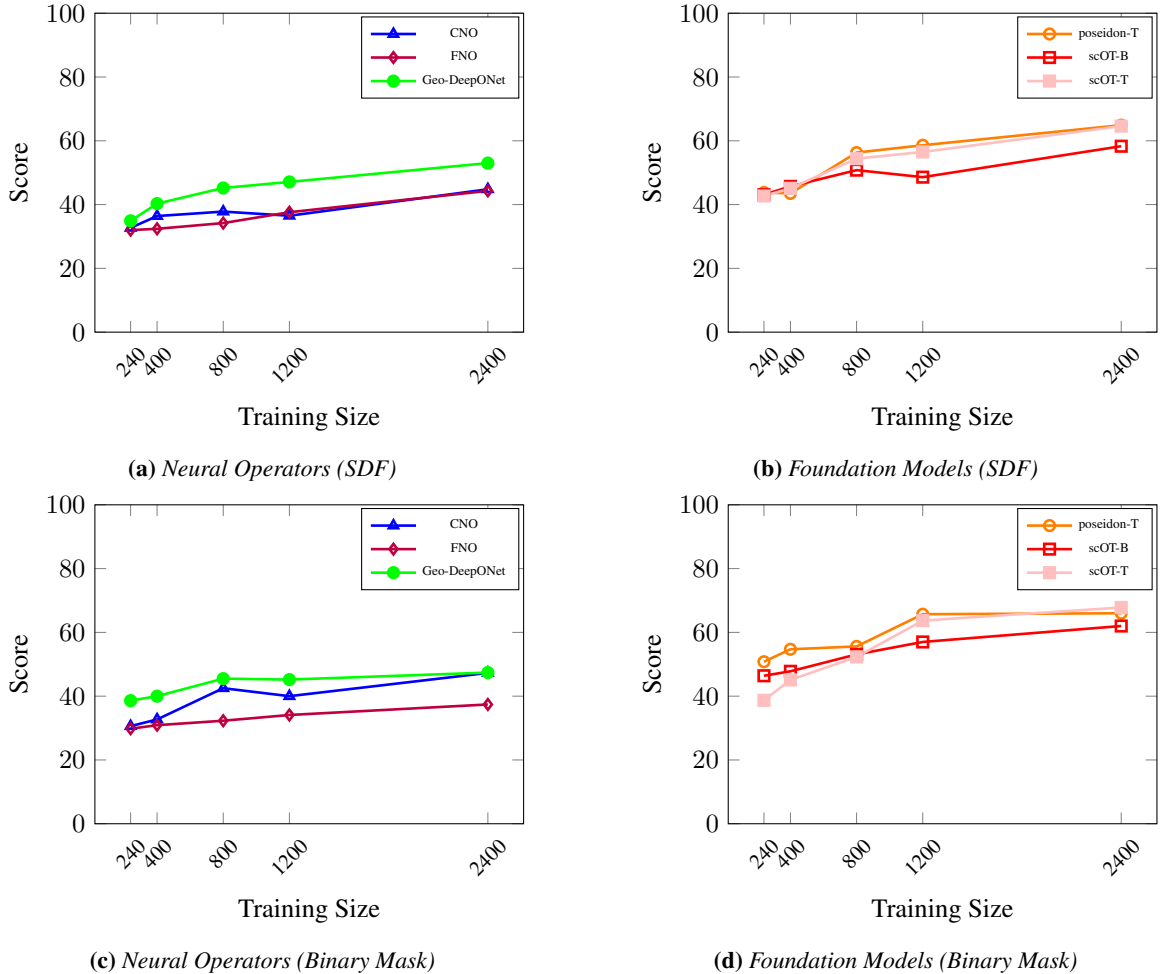

 In \figref{fig:MSE_vs_Sample_Size_Comparison}, model performance varies substantially based on the geometry representation, with notable differences between the Signed Distance Field (SDF) and binary mask. Across all models—poseidon-T, scOT-T, scOT-B, CNO, FNO, and Geometric-DeepONet—larger sample sizes consistently lead to progressively higher score values. However, neural operators reach an asymptotic error limit of around 800 samples in the mask representation, where additional data has minimal impact on further increasing score values. In contrast, scOT and Poseidon models maintain improvements up to 1200 samples, demonstrating their capacity to effectively utilize larger data sizes. This trend underscores the influence of a smooth geometry representation, such as SDF, in enhancing model learning and highlights the ability of scOT and Poseidon models to leverage larger sample sizes due to their larger architectures.

This pattern reveals that when trained with the simpler binary mask, neural operators reach their data utility limit at approximately 800 samples. In contrast, with the SDF representation, these models continue to improve with more data. Poseidon-T, in particular, highlights the benefits of pretraining in data-limited scenarios. When trained on fewer than 800 samples, Poseidon significantly outperforms scOT, achieving an MSE that is an order of magnitude lower (see \figref{fig:MSE_vs_Sample_Size_Comparison}). This performance advantage is especially notable compared to other neural operators, as poseidon-T and scOT-T achieve MSE values around \(10^{-4}\) (score = 50) in data-sparse scenarios, emphasizing their efficiency.

The training dataset often does not fully represent the target distribution for which the model is designed. To address this, we assess the model's ability to extrapolate and make out-of-distribution predictions ("extrapolatory") by employing a test dataset that includes field solutions for lid-driven cavity flows with Reynolds numbers either in the top or bottom 10\% of the range while restricting the training dataset to Reynolds numbers from the middle 80\%. While the scOT and Poseidon models consistently outperform other neural operators across both geometry representations, their performance remains stable in the extrapolatory split regardless of dataset size. This observation suggests that in out-of-distribution scenarios, the ability to extrapolate relies more on the inherent robustness of the model architectures than on the volume of training data. Detailed results for all models at smaller dataset sizes are provided in \appendixref{sec:data-sufficiency} (\Cref{tab:mse-512-sdf-half,tab:mse-512-mask-half,tab:mse-512-sdf-third,tab:mse-512-mask-third,tab:mse-512-sdf-sixth,tab:mse-512-mask-sixth,tab:mse-512-sdf-tenth,tab:mse-512-mask-tenth}). Additionally, sample field predictions and error comparisons for models trained on 240 versus 800 samples, specifically for y-velocity of an example sample, are shown in \figref{fig:1000-vs-300} in \appendixref{sec:field-predictions}.

\begin{tcolorbox}[colframe=white!70!black, colback=blue!10, coltitle=black, title=Key Takeaways on Data Sufficiency]
\textbf{Impact of Sample Size on Performance:} Neural operators benefit from increased data sizes when using the SDF representation, showing continuous improvement, whereas their performance saturates around 800 samples with the binary mask. 

\textbf{Performance in Data-Limited Scenarios:} pre-trained model Poseidon-T demonstrates superior accuracy in data-limited scenarios, reaching an MSE around \(10^{-4}\) (score=50) with fewer than 800 samples.
\end{tcolorbox}

\subsection{Extrapolation Capabilities}\label{subsec:train-test_dataset_distribution}

\begin{figure}[!b]
    \centering
    \begin{subfigure}{0.45\linewidth}
        \centering
        \begin{tikzpicture}
            \begin{axis}[
                width=0.9\linewidth,
                height=0.7\linewidth,
                ybar,
                xlabel={Reynolds Number},
                ylabel={Frequency},
                title={Train - random},
                ymin=0,
                bar width=5pt,
                xticklabel style={/pgf/number format/.cd, fixed, precision=0},
                yticklabel style={/pgf/number format/.cd, fixed, precision=0}
            ]
                \addplot+[
                    hist={bins=20},
                    fill=blue!70,
                    draw=blue!80
                ] table [y=Easy, col sep=comma] {figures/reynolds_numbers_train.csv};
            \end{axis}
        \end{tikzpicture}
    \end{subfigure}
    \begin{subfigure}{0.45\linewidth}
        \centering
        \begin{tikzpicture}
            \begin{axis}[
                width=0.9\linewidth,
                height=0.7\linewidth,
                ybar,
                xlabel={Reynolds Number},
                title={Train - extrapolatory},
                ymin=0,
                bar width=5pt,
                xticklabel style={/pgf/number format/.cd, fixed, precision=0},
                yticklabel style={/pgf/number format/.cd, fixed, precision=0}
            ]
                \addplot+[
                    hist={bins=20},
                    fill=red!70,
                    draw=red!80
                ] table [y=Hard, col sep=comma] {figures/reynolds_numbers_train.csv};
            \end{axis}
        \end{tikzpicture}
    \end{subfigure}

    \begin{subfigure}{0.45\linewidth}
        \centering
        \begin{tikzpicture}
            \begin{axis}[
                width=0.9\linewidth,
                height=0.7\linewidth,
                ybar,
                xlabel={Reynolds Number},
                ylabel={Frequency},
                title={Test - random},
                ymin=0,
                bar width=5pt,
                xticklabel style={/pgf/number format/.cd, fixed, precision=0},
                yticklabel style={/pgf/number format/.cd, fixed, precision=0}
            ]
                \addplot+[
                    hist={bins=20},
                    fill=blue!70,
                    draw=blue!80
                ] table [y=Easy, col sep=comma] {figures/reynolds_numbers_test.csv};
            \end{axis}
        \end{tikzpicture}
    \end{subfigure}
    \begin{subfigure}{0.45\linewidth}
        \centering
        \begin{tikzpicture}
            \begin{axis}[
                width=0.9\linewidth,
                height=0.7\linewidth,
                ybar,
                xlabel={Reynolds Number},
                title={Test - extrapolatory},
                ymin=0,
                bar width=5pt,
                xticklabel style={/pgf/number format/.cd, fixed, precision=0},
                yticklabel style={/pgf/number format/.cd, fixed, precision=0}
            ]
                \addplot+[
                    hist={bins=20},
                    fill=red!70,
                    draw=red!80
                ] table [y=Hard, col sep=comma] {figures/reynolds_numbers_test.csv};
            \end{axis}
        \end{tikzpicture}
    \end{subfigure}

    \caption{Histogram of Reynolds numbers for Train and Test splits in random and extrapolatory cases.}
    \label{fig:reynolds_numbers_comparison}
\end{figure}
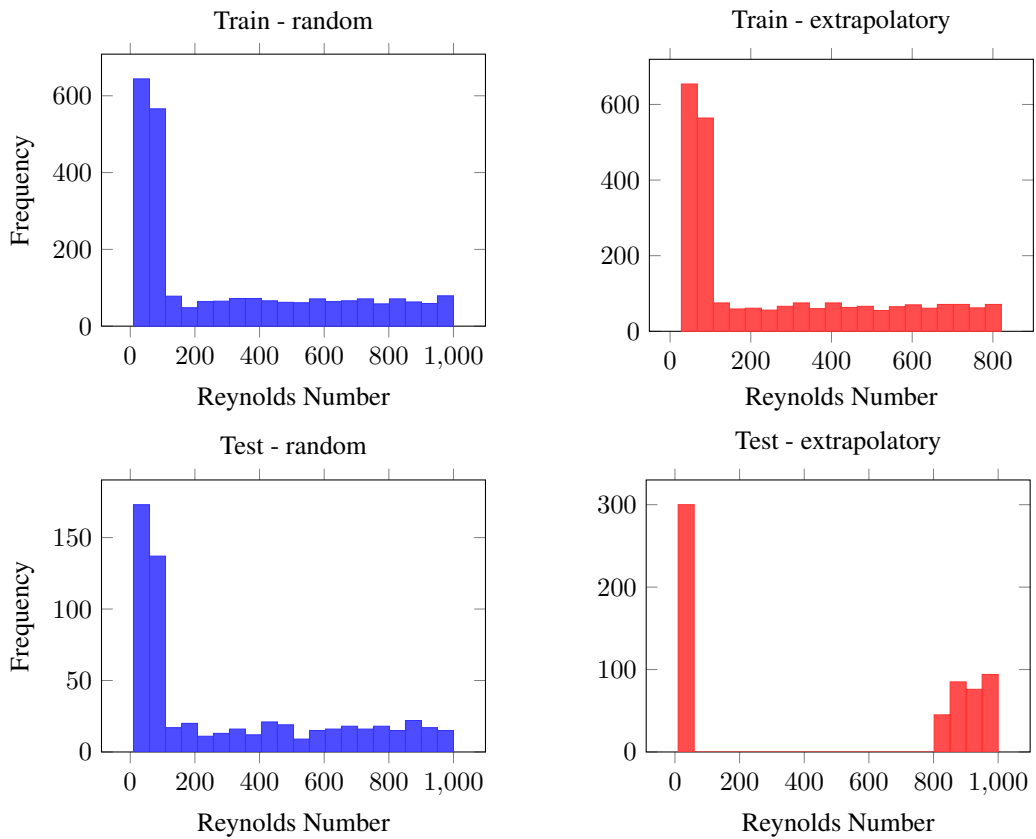

The training dataset often fails to fully represent the target distribution for which the model is intended or may not encompass its entire range. We design two train-test splitting strategies to evaluate the model's ability to make out-of-distribution predictions. For the out-of-distribution experiment, the test dataset comprises field solutions for lid-driven cavity flows with Reynolds numbers in the top or bottom 10\% of the range, while the training dataset is restricted to Reynolds numbers from the middle 80\%. In contrast, the baseline experiment employs a random train-test split, ensuring that both datasets contain samples spanning the entire distribution of Reynolds numbers. The distributions of Reynolds numbers for both splitting strategies are shown in \figref{fig:reynolds_numbers_comparison}.

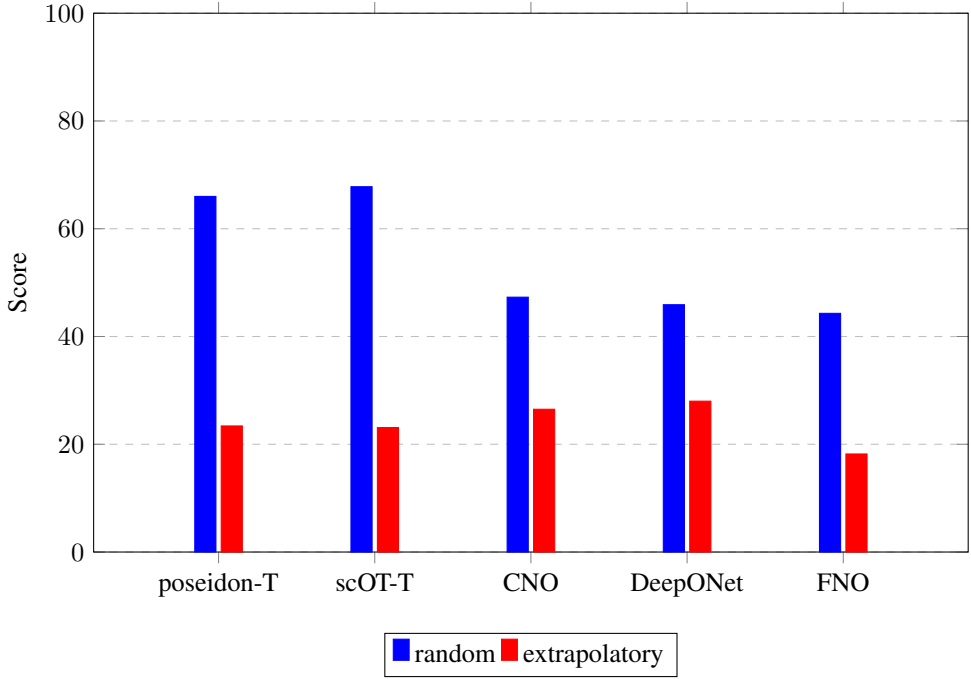
\begin{figure}[t!]
    \centering
    \begin{tikzpicture}
        \begin{axis}[
            ybar,
            bar width=8pt,
            width=0.8\linewidth,
            height=0.53\linewidth,
            enlarge x limits=0.2,
            symbolic x coords={poseidon-T, scOT-T, CNO, DeepONet, FNO},
            xtick=data,
            ylabel={\text{Score}},
            legend style={at={(0.5,-0.15)},anchor=north,legend columns=-1},
            ymin=0,  
            ymax=100,
            log origin=infty,  
            ymajorgrids=true,
            grid style=dashed
        ]
            \addplot+[ybar, fill=blue] coordinates {(poseidon-T,66) (scOT-T,67.8) (CNO,47.3) (DeepONet,45.9) (FNO,44.3)};
            \addplot+[ybar, fill=red] coordinates {(poseidon-T,23.4) (scOT-T,23.1) (CNO,26.5) (DeepONet,28) (FNO,18.2)};
            
            \legend{random, extrapolatory}
        \end{axis}
    \end{tikzpicture}
    \caption{Comparison of score values for different models using random and extrapolatory test/train splits. The bar plot shows the score for each model, indicating the difference in performance between random and extrapolatory data splits. Poseidon-T, scOT-T, and CNO use a binary mask representation of the geometry, while DeepONet and FNO use a Signed Distance Field (SDF).}
    \label{fig:split_error_comparison}
\end{figure}

We train models for each geometric representation using both random and extrapolatory datasets. As shown in \figref{fig:split_error_comparison}, notable differences exist among the models for the extrapolatory data split and substantial performance gaps between each model's random and extrapolatory splits. Specifically, models trained on the random split show stronger performance, with Poseidon and scOT achieving nearly an order of magnitude lower error than other models. For neural operators such as FNO, DeepONet, Geometric-DeepONet, and WNO, we observe that using the SDF as a geometric representation provides marginal but consistent improvements for the extrapolatory data split. This can be attributed to the SDF's ability to encode richer geometric information, including the precise location and structure of objects within the domain, compared to the binary mask's simpler representation. These results highlight the ongoing challenge of accurately extrapolating to out-of-distribution complex fluid dynamics simulations. To further illustrate this, we provide sample field predictions and error comparisons between models trained on the random and extrapolatory splits, focusing on y-velocity for an example sample, as shown in \figref{fig:random-vs-extrapolatory} in \appendixref{sec:field-predictions}.

\begin{tcolorbox}[colframe=white!70!black, colback=blue!10, coltitle=black, title=Key takeaways on extrapolation capabilities]
\textbf{Extrapolation Challenges:} Testing on extreme Reynolds numbers (top and bottom 10\%) reveals the difficulty of generalizing to out-of-distribution scenarios. While Poseidon and scOT perform slightly better, geometric representations and dataset size have minimal impact.
\end{tcolorbox}

\subsection{Performance Metrics}\label{subsec:M2-M3}

We evaluate model performance using three metrics: global accuracy ($M1$), boundary layer accuracy ($M2$), and physical consistency ($M3$). Global accuracy ($M1$) measures overall prediction accuracy across the domain, excluding the geometry. Boundary layer accuracy ($M2$) focuses on errors within the boundary layer (SDF between 0 and 0.2), highlighting precision near surfaces. Physical (PDE) consistency ($M3$) assesses adherence to governing laws by evaluating momentum residuals, ensuring the physical plausibility of predictions.

Notably, our results consistently show that boundary layer MSE errors ($M2$) are lower than global MSE errors ($M1$) across all SciML models, as the velocity values are close to zero near the geometry, resulting in a lower absolute MSE within the boundary layer. This observation is consistent across all results in \Cref{tab:mse-512-sdf,tab:mse-512-mask,tab:mse-512-sdf-half,tab:mse-512-mask-half,tab:mse-512-sdf-third,tab:mse-512-mask-third,tab:mse-512-sdf-sixth,tab:mse-512-mask-sixth,tab:mse-512-sdf-tenth,tab:mse-512-mask-tenth}. Performance in the boundary layer is critical for downstream tasks such as calculating the coefficients of lift and drag using SciML predictions. 

The $M3$ metric, defined as the $L_2$ norm of the momentum residuals $\left( \sqrt{r_x^2 + r_y^2}\right)$, evaluates the models’ ability to satisfy underlying physical laws (PDEs) in fluid dynamics simulations. Analysis of the $M3$ metric reveals that DeepONet consistently achieve the lowest $M3$ error across all dataset configurations, as shown in \Cref{tab:mse-512-sdf,tab:mse-512-mask,tab:mse-512-sdf-half,tab:mse-512-mask-half,tab:mse-512-sdf-third,tab:mse-512-mask-third,tab:mse-512-sdf-sixth,tab:mse-512-mask-sixth,tab:mse-512-sdf-tenth,tab:mse-512-mask-tenth}. Vision transformer-based foundation models excel in minimizing global MSE, leveraging their image-focused architecture for efficient feature extraction. In contrast, neural operators like DeepONet are specifically tailored to model physical behaviors, utilizing basis functions to approximate continuous spatial fields. This inherent alignment with the governing equations allows neural operators to predict smoother solutions, leading to superior performance on the $M3$ metric. A detailed explanation of the momentum residual calculation can be found in \appendixref{sec:residual-calculation}.

Detailed information about the computational characteristics of the models, including their sizes, training times, and inference times, is provided in \appendixref{sec:model-size}. The appendix highlights the significant trade-offs between model size and computational efficiency. For example, while FNO stands out for its rapid training time of just 2.3 hours, DeepONet's smaller model size enables it to achieve the fastest inference time, requiring less than a second per 600 samples. These distinctions emphasize the versatility of these models in balancing computational demands with performance requirements for different applications. In contrast, larger models such as scOT-L and Poseidon-L demand substantially longer training times due to their large architecture, making them better suited for tasks where training time is less critical.

As illustrated in \figref{fig:residuals-mask-vs-sdf}, which compares residuals in the $x$ and $y$ directions for mask and SDF representations in Poseidon-T, CNO, and DeepONet, there is no significant difference in residual values between the SDF and mask representations. However, DeepONet outperforms the other models in the $M3$ metric, achieving the lowest residuals for the sample shown. In particular, the element-wise momentum residual of DeepONet is only non-zero near the top boundary and the surface of the geometry, where the velocity and pressure gradients are the highest. Conversely, Poseidon-T, and to a lesser extent, CNO, exhibit relatively high residuals throughout the domain. This finding suggests that, although it does not have the lowest MSE, DeepONet achieves a more accurate approximation of the solution gradient (see~\figref{fig:vel-error-mask-vs-sdf}), effectively satisfying the underlying PDE behavior more robustly. Additionally, residual values are generally lower in the extrapolatory split than in the random split, likely due to the higher proportion of samples with low Reynolds numbers in the extrapolatory split, simplifying the enforcement of PDE constraints for SciML models.

\begin{tcolorbox}[colframe=white!70!black, colback=blue!10, coltitle=black, title=Key takeaways on model performance metrics]
\textbf{Boundary vs. Global Accuracy ($M1$, $M2$):} Boundary layer MSE ($M2$) is consistently lower than global MSE ($M1$), as the reduced velocity near the geometry leads to naturally lower absolute errors. This indicates that models effectively learn the near-zero velocity conditions around the geometry.

\textbf{PDE Consistency ($M3$):} DeepONet achieves the lowest $M3$ error by leveraging its basis-function based architecture to model continuous fields. This enables DeepONet to capture smoother solutions and thus better satisfy governing equations compared to vision transformer-based models, which excel in feature extraction.

\textbf{Computational Efficiency:} Neural operators are more efficient in training and inference time, while vision-based transformer models demand more time and computational resources.
\end{tcolorbox}

\section{Discussion}\label{sec:discussion_and_conclusion}

This study addresses the challenge of simulating fluid dynamics around complex geometries. Traditional simulation based methods, while accurate, are computationally expensive, prompting the exploration of scientific machine learning (SciML) for faster, scalable solutions. By benchmarking neural operators and vision transformer-based models, this work evaluates geometric representations, data efficiency, and out-of-distribution generalization, offering insights into SciML’s capabilities and limitations for complex flow prediction. Key observations from our experiments include:

\paragraph{Geometric Representation:} The choice of geometric representation influences model performance. The SDF representation generally improves the performance of neural operators such as geometric-DeepONet by offering detailed boundary information and a continuous field representation. In contrast, Poseidon, scOT, and CNO models tend to perform better with the binary mask, leveraging its sharp contrasts to focus on key geometric features.

\paragraph{Data Sufficiency:} The impact of data sufficiency depends on geometric representation and model type. For the mask representation, neural operators saturate at a score of around $40$. With only 300 training samples, Poseidon and scOT models using the mask representation achieve a score of around $50$ (MSE ~$10^{-4}$), highlighting their efficiency in low-data scenarios. Interestingly, in the extrapolatory split, MSE marginally changes across dataset sizes, suggesting that training data size has minimal impact on performance in out-of-distribution predictions.

\paragraph{Model Comparison:} Across all experiments, the scOT and Poseidon models consistently deliver the best performance, often by a significant margin. This superiority stems from the scOT architecture's resemblance to vision transformer (ViT) models, which have demonstrated remarkable success across various image-based tasks. Our findings indicate that \emph{scalable Operator Transformers} outperform other SciML models in modeling fluid dynamics over complex geometries. Notably, scOT and its pre-trained foundation model counterpart, Poseidon, have minimal performance differences. Investigating the impact—or lack thereof—of pretraining in foundation SciML models remains an avenue for future research. Among neural operators, results vary across metrics. DeepONet and FNO excel on the $M3$ metric, reflecting superior physical consistency, while CNO achieves better results on the $M1$ and $M2$ metric, highlighting precision in global/boundary layer field predictions.

\paragraph{Open Questions:} While our study provides insights, several open questions remain: (a) {\textit{Incorporating Physics losses}:} Incorporating physics-based losses, such as the Navier-Stokes momentum or continuity equations, could help penalize high-gradient regions and improve overall model accuracy. (b) {\textit{Robust Foundation Models}:} Throughout all experiments, scOT models and their Poseidon counterparts demonstrated comparable performance, with scOT outperforming Poseidon in several scenarios. This observation highlights the need to further investigate the underlying factors contributing to this discrepancy, offering an opportunity to refine and optimize foundation models for fine-tuning on fluid dynamics across complex geometries. (c) {\textit{Multiphysics Simulations}:} The combination of complex geometries with multiphysics flow phenomena (like buoyancy-driven flows (thermal or concentration), electrokinetic flows, or multi-phase flows) is the next frontier. Our results suggest opportunities and promise for these SciML models, suggesting the need for the creation and curation of such datasets.

\section{Methods}
\subsection{Training Data}\label{sec:background_data}
In this paper, we utilize the FlowBench dataset publicly available on Hugging Face at \url{https://huggingface.co/datasets/BGLab/FlowBench/tree/main} ~\citep{Tali2024}. FlowBench is specifically designed for complex fluid dynamics and heat transfer problems, providing high-fidelity simulations for challenging scientific ML models. The dataset includes 300 diverse parametric and non-parametric geometries that capture intricate flow patterns and transitions around complex shapes. Each geometry is paired with 10 Reynolds numbers, randomly selected between 10 and 1000, to allow for studying flow behavior under varying conditions. Data generation uses a validated framework based on the Navier-Stokes equations using the shifted boundary method to impose boundary conditions on surrogate boundaries~\citep{main2018shifted, yang2024optimal}.

While most research has focused on flow around simple geometries, FlowBench targets more complex shapes, categorized into three distinct types. The first category consists of parametric shapes created using Non-Uniform Rational B-Splines (NURBS) curves, commonly used in computer graphics and CAD for flexible modeling of complex shapes~\citep{piegl2012nurbs}. These shapes are generated by varying control points to produce diverse, smooth curves. The second category features parametric shapes defined by spherical harmonics, allowing for the creation of smooth, radial geometries~\citep{Wei2018}. The third category includes non-parametric shapes sourced from the SkelNetOn dataset, which provides grayscale images of various objects~\citep{demir2019skelneton, Atienza2019}. These categories offer a broad range of shapes for evaluating model performance on fluid dynamics tasks.

This study focuses on the 2D lid-driven cavity flow (LDC) problem using the Navier-Stokes equations. The LDC setup, a canonical problem in fluid dynamics, features a square cavity with three stationary walls and one moving lid, generating complex internal flow structures such as vortices, recirculation zones, and transitions from laminar to turbulent regimes~\citep{Johnston2005, Launder2010}. The FlowBench dataset is formatted as compressed numpy (\texttt{.npz}) files, with input fields including Reynolds numbers, geometry masks, and Signed Distance Fields (SDF) to provide geometric and physical data. The 2D LDC dataset in FlowBench contains 300 geometries with 10 simulations per geometry totaling 3000 samples, with each sample providing output fields such as velocity components ($u$ and $v$) and pressure at $512 \times 512$ resolution. The Reynolds numbers are randomly selected for each geometry, with five values ranging from 10 to 100 and another five from 100 to 1000. By varying Reynolds numbers across different geometries, FlowBench offers an extensive platform for benchmarking scientific machine-learning models on flow prediction across diverse flow regimes.

\subsection{SciML Models}\label{subsec:background_models}
We evaluate 11 different SciML models; six scalable Operator Transformers with different sizes (with/without pretraining), Convolutional Neural Operators, Fourier Neural Operator, DeepONet, Geometric-DeepONet, and the Wavelet Neural Operator. For clarity, we will refer to the spectral-based neural operators, including the Convolution Neural Operator, Fourier Neural Operator, DeepONet, Geometric-DeepONet, and the Wavelet Neural Operator as \emph{neural operators}.

The \emph{Scalable Operator Transformers (scOT)}, the base model for the Poseidon foundation models~\citep{herde2024poseidon}, is a hierarchical, multiscale vision transformer combining Swin Transformer V2~\citep{Liu2021SwinTH, liu2022swinv2scaling} blocks with ConvNeXt~\citep{liu2022convnet2020s} residual blocks in a U-Net encoder-decoder style. The Poseidon models are scOT models pretrained on a dataset of compressible Euler and incompressible Navier-Stokes equations, available in three sizes: tiny (20M parameters), base (160M parameters), and large (630M parameters). We also include randomly initialized versions, totaling six scOT-based models. For clarity, we refer to these as scOT for the randomly initialized models and Poseidon for the pretrained models, along with their model size designation, e.g., scOT-T for the tiny randomly initialized configuration or Poseidon-B for the pretrained base configuration.

The \emph{Fourier Neural Operators (FNO)}~\citep{li2021} leverages Fourier transforms to efficiently capture global interactions within the data, achieving state-of-the-art performance in problems with long-range dependencies, such as turbulence and complex flow dynamics. \emph{Convolutional Neural Operators (CNO)}~\citep{raonic2023}, which extends standard convolutional architectures to learn operator mappings through a U-shaped architecture. CNO employs convolutional kernels to capture localized phenomena while preserving the function's continuous-discrete equivalence, making it effective for modeling fluid flow problems. \emph{Wavelet Neural Operator (WNO)}~\citep{TRIPURA2023115783}, employs wavelet transformations to capture multiscale interactions by decomposing input functions into frequency bands, making it effective for handling problems with both local and global variations.

The \emph{DeepONet}~\citep{lu2020} employs a dual-network architecture, consisting of branch and trunk networks, to separately model input functions. Inspired by the universal approximation theorem for arbitrary continuous functions~\citep{Chen1995}, this architecture enables DeepONet to flexibly represent nonlinear operator relationships. \emph{Geometric-DeepONet}~\citep{he2024} builds on the standard DeepONet by integrating information about geometric representations into the trunk network, enhancing accuracy in scenarios where geometric shapes are included in the domain.

\subsection{Evaluation Metrics}\label{subsec:metrics}
We evaluate the performance of a suite of SciML models in simulating the complex fluid dynamics of the lid-driven cavity flow dataset from FlowBench. Our experimental framework is structured around three core aspects: \emph{geometric representation}, \emph{data sufficiency}, and \emph{ability to extrapolate}. All models were tuned and trained on a single A100 80GB GPU using the Adam optimizer for 200 epochs. The list of hyperparameters used for each model is provided in \appendixref{sec:hyperparameter}. The training and validation loss for four representative models is shown in \figref{fig:loss-plots} in \appendixref{sec:loss-plots}.

\paragraph{Evaluation metrics}: To thoroughly assess the performance of trained models, we introduce a hierarchical framework of evaluation metrics ($M1$, $M2$, $M3$), each designed to assess distinct aspects of model accuracy.
\begin{itemize}[itemsep=0pt]
    \item \textit{$M1$: Global accuracy}: This metric evaluates the overall prediction accuracy across the entire domain (excluding geometry). It assesses how closely the model’s predicted velocity and pressure fields align with actual values, providing a comprehensive view of its general performance across the domain.    
    \item \textit{$M2$: Boundary layer accuracy}: This metric measures errors in the boundary layer, a critical region surrounding the object. Defined by the Signed Distance Field in the range $0 \leq SDF \leq 0.2$, this metric focuses on the model's ability to capture near-surface dynamics, essential for high-fidelity applications such as 3D manufacturing, design optimization, and control. By evaluating errors in this narrow zone, $M2$ is a more challenging test of the precision of the model in handling complex boundary phenomena.
    \item \textit{$M3$: Physical consistency}: This metric assesses the model's adhesion to the governing physical laws (momentum equation), focusing on conservation and residual errors. Evaluating the momentum residual ensures consistency with the governing equations, which is essential to validate the physical plausibility of the predictions of the model.
\end{itemize}

The errors presented are calculated per pixel and normalized to the range of 0 to 100 using the following score-based equation:
\[
\text{score} = 100 \times \left( 1 - \frac{\log(\text{MSE}) - \log(\text{MSE}_{\min})}{\log(\text{MSE}_{\max}) - \log(\text{MSE}_{\min})} \right)
\]
where $\text{MSE}_{\max} = 1$ and $\text{MSE}_{\min} = 10^{-6}$. The choice of $\text{MSE}_{\max} = 1$ represents a scenario in which the predicted solution is essentially meaningless, corresponding to the SciML model that predicts zero everywhere. Conversely, $\text{MSE}_{\min} = 10^{-6}$ reflects the numerical accuracy of the CFD simulations, which were solved using methods with residuals of a similar order. These bounds ensure that the normalized error metric aligns with both the worst-case scenario and the expected precision of the ground truth, providing an intuitive and meaningful scale for model evaluation.

First, we assess the impact of geometric representations---Signed Distance Field (SDF) and binary mask---on the models' ability to capture intricate flow attributes near the geometry. Next, we evaluate data sufficiency by varying the size of the training dataset, providing insights into each model's efficiency in learning the flow operator in data-limited regimes. This is particularly important in scientific machine learning, where data is often difficult and computationally expensive. Lastly, we examine the ability of the models to extrapolate to out-of-distribution test datasets. Specifically, we evaluate how well the models predict field solutions for problems with Reynolds numbers that are either larger or smaller than those encountered during training. In practice, scientific machine learning that can generalize to unseen data is needed for real scientific applications. This multifaceted set of experiments aims to address questions relevant to various partial differential equations (PDEs), extending beyond the lid-driven cavity problem.

\section*{Acknowledgements}
We gratefully acknowledge support from the NAIRR pilot program for computational access. We also acknowledge computational resources from TACC Frontera. This work is supported by the AI Research Institutes program supported by NSF and USDA-NIFA under AI Institute for Resilient Agriculture, Award No. 2021-67021-35329. We also acknowledge partial support through NSF awards CMMI-2053760 and DMREF-2323716

\section*{Data Availability}
\sloppy
This study utilizes the FlowBench 2D Lid-Driven Cavity (LDC) dataset, which is publicly accessible on HuggingFace at \url{https://huggingface.co/datasets/BGLab/FlowBench/tree/main/LDC_NS_2D/512x512}. The dataset is licensed under a CC-BY-NC-4.0 license and serves as a benchmark for the development and evaluation of scientific machine learning (SciML) models. The data on Hugging Face is divided into three geometry sets—nurbs, harmonics, and skeleton—each comprising 1,000 samples. For this study, these sets were combined to create a total dataset of 3,000 samples. The code for the models, along with training procedures and visualization scripts, is available at \url{https://github.com/baskargroup/GeometryMatters}.

\clearpage

\bibliography{references}

\begin{thebibliography}{52}
\providecommand{\natexlab}[1]{#1}
\providecommand{\url}[1]{\texttt{#1}}
\expandafter\ifx\csname urlstyle\endcsname\relax
  \providecommand{\doi}[1]{doi: #1}\else
  \providecommand{\doi}{doi: \begingroup \urlstyle{rm}\Url}\fi

\bibitem[Agarwal(1999)]{Agarwal1999}
Ramesh Agarwal.
\newblock Computational fluid dynamics of whole-body aircraft.
\newblock \emph{Annual review of fluid mechanics}, 31\penalty0 (1):\penalty0 125--169, 1999.

\bibitem[Iaccarino and Mittal(2005)]{Iaccarino2005}
G~Iaccarino and R~Mittal.
\newblock Immersed boundary methods.
\newblock \emph{Annu. Rev. Fluid Mech}, 37:\penalty0 239--261, 2005.

\bibitem[Sudin et~al.(2014)Sudin, Abdullah, Shamsuddin, Ramli, and Tahir]{Sudin2014}
Mohd~Nizam Sudin, Mohd~Azman Abdullah, Shamsul~Anuar Shamsuddin, Faiz~Redza Ramli, and Musthafah~Mohd Tahir.
\newblock Review of research on vehicles aerodynamic drag reduction methods.
\newblock \emph{International Journal of Mechanical and Mechatronics Engineering}, 14\penalty0 (02):\penalty0 37--47, 2014.

\bibitem[Balogh and Bagchi(2017)]{Balogh2017}
Peter Balogh and Prosenjit Bagchi.
\newblock A computational approach to modeling cellular-scale blood flow in complex geometry.
\newblock \emph{Journal of computational physics}, 334:\penalty0 280--307, 2017.

\bibitem[Antiga(2002)]{Antiga2002}
Luca Antiga.
\newblock Patient-specific modeling of geometry and blood flow in large arteries.
\newblock \emph{Politecnico di Milano}, 2002.

\bibitem[Wallisser(2009)]{Wallisser2009}
Tobias Wallisser.
\newblock Other geometries in architecture: bubbles, knots and minimal surfaces.
\newblock In \emph{Mathknow: Mathematics, Applied Sciences and Real Life}, pages 91--111. Springer, 2009.

\bibitem[Pant et~al.(2021)Pant, Doshi, Bahl, and Barati~Farimani]{Pant2021}
Pranshu Pant, Ruchit Doshi, Pranav Bahl, and Amir Barati~Farimani.
\newblock Deep learning for reduced order modelling and efficient temporal evolution of fluid simulations.
\newblock \emph{Physics of Fluids}, 33\penalty0 (10), 2021.

\bibitem[Choudhary et~al.(2022)Choudhary, DeCost, Chen, Jain, Tavazza, Cohn, Park, Choudhary, Agrawal, Billinge, et~al.]{Choudhary2022}
Kamal Choudhary, Brian DeCost, Chi Chen, Anubhav Jain, Francesca Tavazza, Ryan Cohn, Cheol~Woo Park, Alok Choudhary, Ankit Agrawal, Simon~JL Billinge, et~al.
\newblock Recent advances and applications of deep learning methods in materials science.
\newblock \emph{npj Computational Materials}, 8\penalty0 (1):\penalty0 59, 2022.

\bibitem[Balakrishnan et~al.(2024)Balakrishnan, Rider, Barone, and Parish]{Balakrishnan2024}
Uma Balakrishnan, William Rider, Matthew Barone, and Eric Parish.
\newblock Robust data-driven turbulence modeling for rans closures using a sciml approach for validation.
\newblock \emph{Bulletin of the American Physical Society}, 2024.

\bibitem[Kumar et~al.(2023)Kumar, Gleyzer, Kahana, Shukla, and Karniadakis]{Kumar2023}
Varun Kumar, Leonard Gleyzer, Adar Kahana, Khemraj Shukla, and George~Em Karniadakis.
\newblock Mycrunchgpt: A llm assisted framework for scientific machine learning.
\newblock \emph{Journal of Machine Learning for Modeling and Computing}, 4\penalty0 (4), 2023.

\bibitem[Waheed(2023)]{Waheed2023}
Umair~bin Waheed.
\newblock The emergence and impact of scientific machine learning in geophysical exploration.
\newblock In \emph{Third International Meeting for Applied Geoscience \& Energy}, pages 1807--1812. Society of Exploration Geophysicists and American Association of Petroleum~…, 2023.

\bibitem[Sun(2023)]{Sun2023}
Luning Sun.
\newblock \emph{Scientific Machine Learning for Modeling and Discovery of Physical Systems with Quantified Uncertainty}.
\newblock University of Notre Dame, 2023.

\bibitem[Bhattacharjee(2023)]{Bhattacharjee2023}
Shayan Bhattacharjee.
\newblock Integrating scientific machine learning and physics-based models for quantification of uncertainty in thermal properties of silica aerogel.
\newblock Master's thesis, State University of New York at Buffalo, 2023.

\bibitem[Psaros et~al.(2023)Psaros, Meng, Zou, Guo, and Karniadakis]{Psaros2023}
Apostolos~F Psaros, Xuhui Meng, Zongren Zou, Ling Guo, and George~Em Karniadakis.
\newblock Uncertainty quantification in scientific machine learning: Methods, metrics, and comparisons.
\newblock \emph{Journal of Computational Physics}, 477:\penalty0 111902, 2023.

\bibitem[Bonnet et~al.(2022)Bonnet, Mazari, Cinnella, and Gallinari]{Bonnet2022}
Florent Bonnet, Jocelyn Mazari, Paola Cinnella, and Patrick Gallinari.
\newblock Airfrans: High fidelity computational fluid dynamics dataset for approximating reynolds-averaged navier--stokes solutions.
\newblock \emph{Advances in Neural Information Processing Systems}, 35:\penalty0 23463--23478, 2022.

\bibitem[Luo et~al.(2024)Luo, Chen, and Zhang]{Luo2024}
Yining Luo, Yingfa Chen, and Zhen Zhang.
\newblock {CFDBench}: {A} large-scale benchmark for machine learning methods in fluid dynamics, 2024.
\newblock arXiv preprint.

\bibitem[Xu et~al.(2023)Xu, Grande~Gutierrez, and McComb]{Xu2023}
Wenzhuo Xu, Noelia Grande~Gutierrez, and Christopher McComb.
\newblock {MegaFlow2D}: {A} parametric dataset for machine learning super-resolution in computational fluid dynamics simulations.
\newblock In \emph{Proceedings of Cyber-Physical Systems and Internet of Things Week 2023}. Association for Computing Machinery, 2023.
\newblock URL \url{https://doi.org/10.1145/3576914.3587552}.

\bibitem[Collins et~al.(2023)Collins, New, Darragh, Damit, and Stiles]{Collins2023}
Gary~Lynn Collins, Alexander New, Ryan~A Darragh, Brian~E Damit, and Christopher~D Stiles.
\newblock Rapid prediction of two-dimensional airflow in an operating room using scientific machine learning.
\newblock In \emph{NeurIPS 2023 AI for Science Workshop}, 2023.

\bibitem[Tali et~al.(2024)Tali, Rabeh, Yang, Shadkhah, Karki, Upadhyaya, Dhakshinamoorthy, Saadati, Sarkar, Krishnamurthy, et~al.]{Tali2024}
Ronak Tali, Ali Rabeh, Cheng-Hau Yang, Mehdi Shadkhah, Samundra Karki, Abhisek Upadhyaya, Suriya Dhakshinamoorthy, Marjan Saadati, Soumik Sarkar, Adarsh Krishnamurthy, et~al.
\newblock Flowbench: A large scale benchmark for flow simulation over complex geometries.
\newblock \emph{arXiv preprint arXiv:2409.18032}, 2024.

\bibitem[Yang et~al.(2024{\natexlab{a}})Yang, Saurabh, Scovazzi, Canuto, Krishnamurthy, and Ganapathysubramanian]{Yang2024}
Cheng-Hau Yang, Kumar Saurabh, Guglielmo Scovazzi, Claudio Canuto, Adarsh Krishnamurthy, and Baskar Ganapathysubramanian.
\newblock Optimal surrogate boundary selection and scalability studies for the shifted boundary method on octree meshes.
\newblock \emph{Computer Methods in Applied Mechanics and Engineering}, 419:\penalty0 116686, 2024{\natexlab{a}}.

\bibitem[Zhang et~al.(2019)Zhang, Wu, and Nandakumar]{Zhang2019}
Chenguang Zhang, Chunliang Wu, and Krishnaswamy Nandakumar.
\newblock Effective geometric algorithms for immersed boundary method using signed distance field.
\newblock \emph{Journal of Fluids Engineering}, 141\penalty0 (6):\penalty0 061401, 2019.

\bibitem[Lai et~al.(2023)Lai, Zhao, Zhao, and Huang]{Lai2023}
Zhengshou Lai, Jidong Zhao, Shiwei Zhao, and Linchong Huang.
\newblock Signed distance field enhanced fully resolved cfd-dem for simulation of granular flows involving multiphase fluids and irregularly shaped particles.
\newblock \emph{Computer Methods in Applied Mechanics and Engineering}, 414:\penalty0 116195, 2023.

\bibitem[Li et~al.(2024)Li, Spelman, and Sansalone]{Li2024unit}
Haochen Li, David Spelman, and John Sansalone.
\newblock Unit operation and process modeling with physics-informed machine learning.
\newblock \emph{Journal of Environmental Engineering}, 150\penalty0 (4):\penalty0 04024002, 2024.

\bibitem[Gkimisis et~al.(2023)Gkimisis, Dias, Scoggins, Magin, Mendez, and Turchi]{Gkimisis2023}
Leonidas Gkimisis, Bruno Dias, James~B Scoggins, Thierry Magin, Miguel~A Mendez, and Alessandro Turchi.
\newblock Data-driven modeling of hypersonic reentry flow with heat and mass transfer.
\newblock \emph{AIAA Journal}, 61\penalty0 (8):\penalty0 3269--3286, 2023.

\bibitem[Kwak et~al.(2005)Kwak, Kiris, and Kim]{Kwak2005}
Dochan Kwak, Cetin Kiris, and Chang~Sung Kim.
\newblock Computational challenges of viscous incompressible flows.
\newblock \emph{Computers \& fluids}, 34\penalty0 (3):\penalty0 283--299, 2005.

\bibitem[Zhu et~al.(2022)Zhu, Chen, Ouyang, Yan, Lei, Chen, and Luo]{Zhu2022}
Li-Tao Zhu, Xi-Zhong Chen, Bo~Ouyang, Wei-Cheng Yan, He~Lei, Zhe Chen, and Zheng-Hong Luo.
\newblock Review of machine learning for hydrodynamics, transport, and reactions in multiphase flows and reactors.
\newblock \emph{Industrial \& Engineering Chemistry Research}, 61\penalty0 (28):\penalty0 9901--9949, 2022.

\bibitem[Subramanian et~al.(2024)Subramanian, Harrington, Keutzer, Bhimji, Morozov, Mahoney, and Gholami]{Subramanian2024}
Shashank Subramanian, Peter Harrington, Kurt Keutzer, Wahid Bhimji, Dmitriy Morozov, Michael~W Mahoney, and Amir Gholami.
\newblock Towards foundation models for scientific machine learning: Characterizing scaling and transfer behavior.
\newblock \emph{Advances in Neural Information Processing Systems}, 36, 2024.

\bibitem[Muckley et~al.(2023)Muckley, Saal, Meredig, Roper, and Martin]{Muckley2023}
Eric~S Muckley, James~E Saal, Bryce Meredig, Christopher~S Roper, and John~H Martin.
\newblock Interpretable models for extrapolation in scientific machine learning.
\newblock \emph{Digital Discovery}, 2\penalty0 (5):\penalty0 1425--1435, 2023.

\bibitem[Goswami et~al.(2024)Goswami, Jagtap, Babaee, Susi, and Karniadakis]{Goswami2024}
Somdatta Goswami, Ameya~D Jagtap, Hessam Babaee, Bryan~T Susi, and George~Em Karniadakis.
\newblock Learning stiff chemical kinetics using extended deep neural operators.
\newblock \emph{Computer Methods in Applied Mechanics and Engineering}, 419:\penalty0 116674, 2024.

\bibitem[Papamakarios et~al.(2021)Papamakarios, Nalisnick, Rezende, Mohamed, and Lakshminarayanan]{Papamakarios2021}
George Papamakarios, Eric Nalisnick, Danilo~Jimenez Rezende, Shakir Mohamed, and Balaji Lakshminarayanan.
\newblock Normalizing flows for probabilistic modeling and inference.
\newblock \emph{Journal of Machine Learning Research}, 22\penalty0 (57):\penalty0 1--64, 2021.

\bibitem[Shen(2018)]{Shen2018}
Chaopeng Shen.
\newblock A transdisciplinary review of deep learning research and its relevance for water resources scientists.
\newblock \emph{Water Resources Research}, 54\penalty0 (11):\penalty0 8558--8593, 2018.

\bibitem[Rabeh et~al.(2024)Rabeh, Khanwale, Lee, and Ganapathysubramanian]{Rabeh2024}
Ali Rabeh, Makrand~A Khanwale, Jonghyun Lee, and Baskar Ganapathysubramanian.
\newblock Modeling and simulations of high-density two-phase flows using projection-based cahn-hilliard navier-stokes equations.
\newblock \emph{arXiv preprint arXiv:2406.17933}, 2024.

\bibitem[Khara et~al.(2024{\natexlab{a}})Khara, Herron, Balu, Gamdha, Yang, Saurabh, Jignasu, Jiang, Sarkar, Hegde, et~al.]{Khara2024}
Biswajit Khara, Ethan Herron, Aditya Balu, Dhruv Gamdha, Chih-Hsuan Yang, Kumar Saurabh, Anushrut Jignasu, Zhanhong Jiang, Soumik Sarkar, Chinmay Hegde, et~al.
\newblock Neural pde solvers for irregular domains.
\newblock \emph{Computer-Aided Design}, 172:\penalty0 103709, 2024{\natexlab{a}}.

\bibitem[Khara et~al.(2024{\natexlab{b}})Khara, Balu, Joshi, Sarkar, Hegde, Krishnamurthy, and Ganapathysubramanian]{NeuFENetKhara2024}
Biswajit Khara, Aditya Balu, Ameya Joshi, Soumik Sarkar, Chinmay Hegde, Adarsh Krishnamurthy, and Baskar Ganapathysubramanian.
\newblock Neufenet: Neural finite element solutions with theoretical bounds for parametric pdes.
\newblock \emph{Engineering with Computers}, 40\penalty0 (5):\penalty0 2761--2783, October 2024{\natexlab{b}}.
\newblock \doi{10.1007/s00366-024-01955-7}.
\newblock URL \url{https://doi.org/10.1007/s00366-024-01955-7}.

\bibitem[Main and Scovazzi(2018)]{main2018shifted}
Alex Main and Guglielmo Scovazzi.
\newblock The shifted boundary method for embedded domain computations. part {I}: Poisson and stokes problems.
\newblock \emph{Journal of Computational Physics}, 372:\penalty0 972--995, 2018.

\bibitem[Yang et~al.(2024{\natexlab{b}})Yang, Saurabh, Scovazzi, Canuto, Krishnamurthy, and Ganapathysubramanian]{yang2024optimal}
Cheng-Hau Yang, Kumar Saurabh, Guglielmo Scovazzi, Claudio Canuto, Adarsh Krishnamurthy, and Baskar Ganapathysubramanian.
\newblock Optimal surrogate boundary selection and scalability studies for the shifted boundary method on octree meshes.
\newblock \emph{Computer Methods in Applied Mechanics and Engineering}, 419:\penalty0 116686, 2024{\natexlab{b}}.

\bibitem[Piegl and Tiller(2012)]{piegl2012nurbs}
Les Piegl and Wayne Tiller.
\newblock \emph{The NURBS book}.
\newblock Springer Science \& Business Media, 2012.

\bibitem[Wei et~al.(2018)Wei, Wang, and Zhao]{Wei2018}
Deheng Wei, Jianfeng Wang, and Budi Zhao.
\newblock A simple method for particle shape generation with spherical harmonics.
\newblock \emph{Powder Technology}, 330:\penalty0 284--291, 2018.
\newblock ISSN 0032-5910.
\newblock \doi{https://doi.org/10.1016/j.powtec.2018.02.006}.
\newblock URL \url{https://www.sciencedirect.com/science/article/pii/S0032591018301189}.

\bibitem[Demir et~al.(2019)Demir, Hahn, Leonard, Morin, Rahbani, Panotopoulou, Fondevilla, Balashova, Durix, and Kortylewski]{demir2019skelneton}
Ilke Demir, Camilla Hahn, Kathryn Leonard, Geraldine Morin, Dana Rahbani, Athina Panotopoulou, Amelie Fondevilla, Elena Balashova, Bastien Durix, and Adam Kortylewski.
\newblock Skelneton 2019: Dataset and challenge on deep learning for geometric shape understanding.
\newblock In \emph{Proceedings of the IEEE/CVF conference on computer vision and pattern recognition workshops}, pages 0--0, 2019.

\bibitem[Atienza(2019)]{Atienza2019}
Rowel Atienza.
\newblock Pyramid u-network for skeleton extraction from shape points.
\newblock In \emph{The IEEE Conference on Computer Vision and Pattern Recognition (CVPR) Workshops}, June 2019.

\bibitem[Johnston(2005)]{Johnston2005}
JP~Johnston.
\newblock Internal flows.
\newblock \emph{Turbulence}, pages 109--169, 2005.

\bibitem[Launder et~al.(2010)Launder, Poncet, and Serre]{Launder2010}
Brian Launder, S{\'e}bastien Poncet, and Eric Serre.
\newblock Laminar, transitional, and turbulent flows in rotor-stator cavities.
\newblock \emph{Annual review of fluid mechanics}, 42\penalty0 (1):\penalty0 229--248, 2010.

\bibitem[Herde et~al.(2024)Herde, Raonić, Rohner, Käppeli, Molinaro, de~Bézenac, and Mishra]{herde2024poseidon}
Maximilian Herde, Bogdan Raonić, Tobias Rohner, Roger Käppeli, Roberto Molinaro, Emmanuel de~Bézenac, and Siddhartha Mishra.
\newblock Poseidon: Efficient foundation models for pdes, 2024.

\bibitem[Liu et~al.(2021)Liu, Lin, Cao, Hu, Wei, Zhang, Lin, and Guo]{Liu2021SwinTH}
Ze~Liu, Yutong Lin, Yue Cao, Han Hu, Yixuan Wei, Zheng Zhang, Stephen Lin, and Baining Guo.
\newblock Swin transformer: Hierarchical vision transformer using shifted windows.
\newblock \emph{2021 IEEE/CVF International Conference on Computer Vision (ICCV)}, pages 9992--10002, 2021.
\newblock URL \url{https://api.semanticscholar.org/CorpusID:232352874}.

\bibitem[Liu et~al.(2022{\natexlab{a}})Liu, Hu, Lin, Yao, Xie, Wei, Ning, Cao, Zhang, Dong, Wei, and Guo]{liu2022swinv2scaling}
Ze~Liu, Han Hu, Yutong Lin, Zhuliang Yao, Zhenda Xie, Yixuan Wei, Jia Ning, Yue Cao, Zheng Zhang, Li~Dong, Furu Wei, and Baining Guo.
\newblock Swin transformer v2: Scaling up capacity and resolution, 2022{\natexlab{a}}.
\newblock URL \url{https://arxiv.org/abs/2111.09883}.

\bibitem[Liu et~al.(2022{\natexlab{b}})Liu, Mao, Wu, Feichtenhofer, Darrell, and Xie]{liu2022convnet2020s}
Zhuang Liu, Hanzi Mao, Chao-Yuan Wu, Christoph Feichtenhofer, Trevor Darrell, and Saining Xie.
\newblock A convnet for the 2020s, 2022{\natexlab{b}}.
\newblock URL \url{https://arxiv.org/abs/2201.03545}.

\bibitem[Li et~al.(2021)Li, Kovachki, Azizzadenesheli, Liu, Bhattacharya, Stuart, and Anandkumar]{li2021}
Zongyi Li, Nikola Kovachki, Kamyar Azizzadenesheli, Burigede Liu, Kaushik Bhattacharya, Andrew Stuart, and Anima Anandkumar.
\newblock Fourier neural operator for parametric partial differential equations, 2021.

\bibitem[Raonić et~al.(2023)Raonić, Molinaro, Ryck, Rohner, Bartolucci, Alaifari, Mishra, and de~Bézenac]{raonic2023}
Bogdan Raonić, Roberto Molinaro, Tim~De Ryck, Tobias Rohner, Francesca Bartolucci, Rima Alaifari, Siddhartha Mishra, and Emmanuel de~Bézenac.
\newblock Convolutional neural operators for robust and accurate learning of pdes, 2023.

\bibitem[Tripura and Chakraborty(2023)]{TRIPURA2023115783}
Tapas Tripura and Souvik Chakraborty.
\newblock Wavelet neural operator for solving parametric partial differential equations in computational mechanics problems.
\newblock \emph{Computer Methods in Applied Mechanics and Engineering}, 404:\penalty0 115783, 2023.
\newblock ISSN 0045-7825.
\newblock \doi{https://doi.org/10.1016/j.cma.2022.115783}.
\newblock URL \url{https://www.sciencedirect.com/science/article/pii/S0045782522007393}.

\bibitem[Lu et~al.(2021)Lu, Jin, Pang, Zhang, and Karniadakis]{lu2020}
Lu~Lu, Pengzhan Jin, Guofei Pang, Zhongqiang Zhang, and George~Em Karniadakis.
\newblock Learning nonlinear operators via deeponet based on the universal approximation theorem of operators.
\newblock \emph{Nature Machine Intelligence}, 3:\penalty0 218--–229, 2021.

\bibitem[Chen and Chen(1995)]{Chen1995}
Tianping Chen and Hong Chen.
\newblock Universal approximation to nonlinear operators by neural networks with arbitrary activation functions and its application to dynamical systems.
\newblock \emph{IEEE transactions on neural networks}, 6\penalty0 (4):\penalty0 911--917, 1995.

\bibitem[He et~al.(2024)He, Koric, Abueidda, Najafi, and Jasiuk]{he2024}
Junyan He, Seid Koric, Diab Abueidda, Ali Najafi, and Iwona Jasiuk.
\newblock Geom-deeponet: A point-cloud-based deep operator network for field predictions on 3d parameterized geometries.
\newblock \emph{Computer Methods in Applied Mechanics and Engineering}, 429:\penalty0 117130, September 2024.
\newblock ISSN 0045-7825.
\newblock \doi{10.1016/j.cma.2024.117130}.
\newblock URL \url{http://dx.doi.org/10.1016/j.cma.2024.117130}.

\end{thebibliography}

\appendix
\newpage
\setcounter{page}{1}
\renewcommand\thefigure{A.\arabic{figure}}    
\renewcommand\thetable{A.\arabic{table}}    
\setcounter{figure}{0}
\setcounter{table}{0}

\section{Data Sufficiency} \label{sec:data-sufficiency}

In this appendix, we explore the data sufficiency requirements for accurate performance of scientific machine learning (SciML) models across varying conditions. We assess model performance on subsets of one-third and one-tenth of the original data using the signed distance field or the binary mask geometry representation. Tables \ref{tab:mse-512-sdf-half} through \ref{tab:mse-512-mask-tenth} summarize the mean squared errors (MSE) across two difficulty levels, random and exrapolatory, allowing for a detailed comparison of how reduced data availability impacts model accuracy. By examining these results, we aim to highlight the trade-offs in error rates and overall model robustness under limited training data, guiding the selection of optimal data requirements for efficient and reliable SciML model applications.

\begin{table}[h!]
\centering
\small
\setlength\extrarowheight{2pt}
\caption{The score of SciML models trained on a subset of half of the dataset using the signed distance field at two different difficulty levels (random and extrapolatory). All errors are reported on the testing dataset.}
\label{tab:mse-512-sdf-half}
\begin{tabular}{c c c c c c c}
\hline
 & \multicolumn{3}{c}{\textbf{Random}} & \multicolumn{3}{c}{\textbf{Extrapolatory}} \\ \hline
\textbf{Model} & \textbf{M1} & \textbf{M2} & \textbf{M3} & \textbf{M1} & \textbf{M2} & \textbf{M3} \\ \hline
\textbf{poseidon-L} & $56.4$ & $65.3$ & $24.0$ & $\mathbf{23.9}$ & $\mathbf{36.9}$ & $28.8$ \\ \hline
\textbf{poseidon-B} & $51.9$ & $62.1$ & $23.9$ & $22.7$ & $35.9$ & $30.4$ \\ \hline
\textbf{poseidon-T} & $\mathbf{58.6}$ & $\mathbf{65.5}$ & $24.1$ & $21.8$ & $36.4$ & $27.6$ \\ \hline
\textbf{scOT-L} & $52.2$ & $60.4$ & $25.3$ & $20.9$ & $34.6$ & $30.4$ \\ \hline
\textbf{scOT-B} & $48.6$ & $60.5$ & $23.5$ & $20.3$ & $35.2$ & $31.0$ \\ \hline
\textbf{scOT-T} & $56.5$ & $63.9$ & $24.3$ & $22.5$ & $37.1$ & $30.1$ \\ \hline
\textbf{CNO} & $36.5$ & $47.2$ & $22.5$ & $21.6$ & $33.9$ & $29.4$ \\ \hline
\textbf{FNO} & $37.6$ & $56.2$ & $29.1$ & $15.3$ & $28.9$ & $44.4$ \\ \hline
\textbf{WNO} & $24.6$ & $40.0$ & $32.4$ & $12.1$ & $25.8$ & $0.0$ \\ \hline
\textbf{Deeponet} & $43.3$ & $50.5$ & $\mathbf{39.3}$ & $19.6$ & $31.0$ & $\mathbf{44.7}$ \\ \hline
\textbf{geometric-deeponet} & $47.1$ & $56.4$ & $35.2$ & $19.7$ & $33.1$ & $42.4$ \\ \hline
\end{tabular}
\end{table}

\begin{table}[h!]
\centering
\small
\setlength\extrarowheight{2pt}
\caption{The score of SciML models trained on a subset of half of the dataset using the binary mask at two different difficulty levels (random and extrapolatory). All errors are reported on the testing dataset.}
\label{tab:mse-512-mask-half}
\begin{tabular}{c c c c c c c}
\hline
 & \multicolumn{3}{c}{\textbf{Random}} & \multicolumn{3}{c}{\textbf{Extrapolatory}} \\ \hline
\textbf{Model} & \textbf{M1} & \textbf{M2} & \textbf{M3} & \textbf{M1} & \textbf{M2} & \textbf{M3} \\ \hline
\textbf{poseidon-L} & $63.4$ & $72.3$ & $24.6$ & $26.1$ & $\mathbf{41.0}$ & $15.9$ \\ \hline
\textbf{poseidon-B} & $56.4$ & $66.5$ & $24.6$ & $21.1$ & $33.6$ & $31.8$ \\ \hline
\textbf{poseidon-T} & $\mathbf{65.7}$ & $\mathbf{73.1}$ & $24.6$ & $\mathbf{26.8}$ & $39.8$ & $30.4$ \\ \hline
\textbf{scOT-L} & $62.7$ & $70.9$ & $24.5$ & $20.2$ & $33.7$ & $32.4$ \\ \hline
\textbf{scOT-B} & $57.0$ & $67.4$ & $24.1$ & $20.0$ & $32.9$ & $33.2$ \\ \hline
\textbf{scOT-T} & $63.7$ & $70.7$ & $24.5$ & $20.5$ & $33.5$ & $33.6$ \\ \hline
\textbf{CNO} & $40.0$ & $50.4$ & $26.7$ & $22.6$ & $35.1$ & $32.0$ \\ \hline
\textbf{FNO} & $34.1$ & $55.5$ & $30.5$ & $18.5$ & $33.4$ & $\mathbf{39.1}$ \\ \hline
\textbf{WNO} & $24.3$ & $38.8$ & $29.8$ & $13.2$ & $27.4$ & $0.0$ \\ \hline
\textbf{Deeponet} & $43.9$ & $53.7$ & $32.6$ & $21.3$ & $32.9$ & $37.1$ \\ \hline
\textbf{geometric-deeponet} & $45.2$ & $54.0$ & $\mathbf{32.9}$ & $19.7$ & $33.3$ & $38.6$ \\ \hline
\end{tabular}
\end{table}

\begin{table}[t!]
\centering
\small
\setlength\extrarowheight{2pt}
\caption{The score of SciML models trained on a subset of one-third of the dataset using the signed distance field at two different difficulty levels (random and extrapolatory). All errors are reported on the testing dataset.}
\label{tab:mse-512-sdf-third}
\begin{tabular}{c c c c c c c}
\hline
 & \multicolumn{3}{c}{\textbf{Random}} & \multicolumn{3}{c}{\textbf{Extrapolatory}} \\ \hline
\textbf{Model} & \textbf{M1} & \textbf{M2} & \textbf{M3} & \textbf{M1} & \textbf{M2} & \textbf{M3} \\ \hline
\textbf{poseidon-L} & $53.3$ & $62.0$ & $24.0$ & $\mathbf{23.9}$ & $36.8$ & $27.2$ \\ \hline
\textbf{poseidon-B} & $39.9$ & $48.7$ & $27.0$ & $21.1$ & $34.7$ & $31.1$ \\ \hline
\textbf{poseidon-T} & $\mathbf{56.3}$ & $\mathbf{63.7}$ & $24.7$ & $20.7$ & $35.9$ & $21.5$ \\ \hline
\textbf{scOT-L} & $52.0$ & $60.0$ & $24.8$ & $20.0$ & $34.2$ & $30.5$ \\ \hline
\textbf{scOT-B} & $50.8$ & $59.3$ & $24.8$ & $20.4$ & $34.1$ & $31.8$ \\ \hline
\textbf{scOT-T} & $54.4$ & $61.6$ & $24.5$ & $22.0$ & $\mathbf{37.6}$ & $29.1$ \\ \hline
\textbf{CNO} & $37.8$ & $45.4$ & $27.5$ & $20.4$ & $30.8$ & $28.8$ \\ \hline
\textbf{FNO} & $34.2$ & $53.6$ & $34.5$ & $15.1$ & $28.8$ & $45.8$ \\ \hline
\textbf{WNO} & $23.4$ & $38.4$ & $36.2$ & $12.7$ & $26.3$ & $24.4$ \\ \hline
\textbf{Deeponet} & $42.4$ & $50.5$ & $\mathbf{44.9}$ & $22.1$ & $31.4$ & $\mathbf{51.0}$ \\ \hline
\textbf{geometric-deeponet} & $45.2$ & $56.3$ & $34.7$ & $19.5$ & $32.4$ & $42.6$ \\ \hline
\end{tabular}
\end{table}

\begin{table}[t!]
\centering
\small
\setlength\extrarowheight{2pt}
\caption{The score of SciML models trained on a subset of one-third of the dataset using the binary mask at two different difficulty levels (random and extrapolatory). All errors are reported on the testing dataset.}
\label{tab:mse-512-mask-third}
\begin{tabular}{c c c c c c c}
\hline
 & \multicolumn{3}{c}{\textbf{Random}} & \multicolumn{3}{c}{\textbf{Extrapolatory}} \\ \hline
\textbf{Model} & \textbf{M1} & \textbf{M2} & \textbf{M3} & \textbf{M1} & \textbf{M2} & \textbf{M3} \\ \hline
\textbf{poseidon-L} & $50.7$ & $62.7$ & $21.9$ & $28.2$ & $42.5$ & $27.6$ \\ \hline
\textbf{poseidon-B} & $49.1$ & $59.8$ & $24.4$ & $20.8$ & $33.4$ & $32.5$ \\ \hline
\textbf{poseidon-T} & $\mathbf{55.6}$ & $63.7$ & $26.3$ & $\mathbf{29.3}$ & $\mathbf{46.8}$ & $23.2$ \\ \hline
\textbf{scOT-L} & $54.8$ & $\mathbf{64.3}$ & $25.2$ & $20.5$ & $33.4$ & $32.3$ \\ \hline
\textbf{scOT-B} & $53.1$ & $62.3$ & $24.8$ & $19.7$ & $33.0$ & $33.7$ \\ \hline
\textbf{scOT-T} & $52.3$ & $62.7$ & $23.9$ & $20.4$ & $33.1$ & $32.9$ \\ \hline
\textbf{CNO} & $42.5$ & $52.5$ & $28.3$ & $18.5$ & $30.6$ & $34.4$ \\ \hline
\textbf{FNO} & $32.3$ & $53.8$ & $33.7$ & $17.2$ & $31.6$ & $\mathbf{43.1}$ \\ \hline
\textbf{WNO} & $24.1$ & $39.1$ & $27.3$ & $13.1$ & $26.6$ & $15.3$ \\ \hline
\textbf{Deeponet} & $41.2$ & $52.4$ & $\mathbf{36.4}$ & $20.8$ & $32.6$ & $40.3$ \\ \hline
\textbf{geometric-deeponet} & $45.5$ & $54.4$ & $33.7$ & $18.1$ & $31.3$ & $\mathbf{43.1}$ \\ \hline
\end{tabular}
\end{table}

\begin{table}[t!]
\centering
\small
\setlength\extrarowheight{2pt}
\caption{The score of SciML models trained on a subset of one-sixth of the dataset using the signed distance field at two different difficulty levels (random and extrapolatory). All errors are reported on the testing dataset.}
\label{tab:mse-512-sdf-sixth}
\begin{tabular}{c c c c c c c}
\hline
 & \multicolumn{3}{c}{\textbf{Random}} & \multicolumn{3}{c}{\textbf{Extrapolatory}} \\ \hline
\textbf{Model} & \textbf{M1} & \textbf{M2} & \textbf{M3} & \textbf{M1} & \textbf{M2} & \textbf{M3} \\ \hline
\textbf{poseidon-L} & $\mathbf{46.8}$ & $\mathbf{57.9}$ & $24.5$ & $\mathbf{27.1}$ & $38.0$ & $24.0$ \\ \hline
\textbf{poseidon-B} & $43.6$ & $52.6$ & $25.1$ & $21.4$ & $34.9$ & $31.3$ \\ \hline
\textbf{poseidon-T} & $43.5$ & $53.7$ & $25.2$ & $25.7$ & $\mathbf{42.0}$ & $19.0$ \\ \hline
\textbf{scOT-L} & $45.9$ & $54.5$ & $27.1$ & $19.4$ & $33.3$ & $31.7$ \\ \hline
\textbf{scOT-B} & $45.7$ & $55.4$ & $24.3$ & $19.3$ & $31.5$ & $33.5$ \\ \hline
\textbf{scOT-T} & $45.1$ & $53.7$ & $28.0$ & $19.3$ & $32.6$ & $33.9$ \\ \hline
\textbf{CNO} & $36.4$ & $47.3$ & $27.6$ & $20.2$ & $32.3$ & $33.4$ \\ \hline
\textbf{FNO} & $32.4$ & $50.5$ & $37.7$ & $15.7$ & $29.7$ & $46.1$ \\ \hline
\textbf{WNO} & $22.2$ & $38.6$ & $0.0$ & $11.9$ & $25.9$ & $0.0$ \\ \hline
\textbf{Deeponet} & $37.7$ & $47.9$ & $\mathbf{50.3}$ & $17.2$ & $28.8$ & $\mathbf{58.9}$ \\ \hline
\textbf{geometric-deeponet} & $40.3$ & $52.9$ & $43.7$ & $19.5$ & $32.6$ & $43.6$ \\ \hline
\end{tabular}
\end{table}

\begin{table}[t!]
\centering
\small
\setlength\extrarowheight{2pt}
\caption{The score of SciML models trained on a subset of one-sixth of the dataset using the binary mask at two different difficulty levels (random and extrapolatory). All errors are reported on the testing dataset.}
\label{tab:mse-512-mask-sixth}
\begin{tabular}{c c c c c c c}
\hline
 & \multicolumn{3}{c}{\textbf{Random}} & \multicolumn{3}{c}{\textbf{Extrapolatory}} \\ \hline
\textbf{Model} & \textbf{M1} & \textbf{M2} & \textbf{M3} & \textbf{M1} & \textbf{M2} & \textbf{M3} \\ \hline
\textbf{poseidon-L} & $50.4$ & $60.1$ & $25.0$ & $\mathbf{34.5}$ & $\mathbf{50.6}$ & $12.6$ \\ \hline
\textbf{poseidon-B} & $44.8$ & $54.1$ & $25.3$ & $26.3$ & $39.9$ & $25.1$ \\ \hline
\textbf{poseidon-T} & $\mathbf{54.7}$ & $\mathbf{62.9}$ & $26.3$ & $30.2$ & $48.0$ & $27.9$ \\ \hline
\textbf{scOT-L} & $51.3$ & $60.1$ & $24.5$ & $17.5$ & $32.4$ & $28.0$ \\ \hline
\textbf{scOT-B} & $47.8$ & $58.0$ & $23.5$ & $20.4$ & $32.7$ & $33.8$ \\ \hline
\textbf{scOT-T} & $45.1$ & $55.7$ & $27.2$ & $22.7$ & $36.5$ & $29.8$ \\ \hline
\textbf{CNO} & $32.7$ & $44.9$ & $26.9$ & $24.5$ & $36.2$ & $28.6$ \\ \hline
\textbf{FNO} & $30.9$ & $51.2$ & $35.9$ & $16.4$ & $31.1$ & $45.0$ \\ \hline
\textbf{WNO} & $22.4$ & $38.4$ & $0.0$ & $11.5$ & $26.2$ & $0.0$ \\ \hline
\textbf{Deeponet} & $39.2$ & $52.0$ & $37.0$ & $17.6$ & $29.9$ & $\mathbf{49.8}$ \\ \hline
\textbf{geometric-deeponet} & $40.0$ & $51.2$ & $\mathbf{37.7}$ & $18.9$ & $32.1$ & $41.7$ \\ \hline
\end{tabular}
\end{table}

\begin{table}[t!]
\centering
\small
\setlength\extrarowheight{2pt}
\caption{The score of SciML models trained on a subset of one-tenth of the dataset using the signed distance field at two different difficulty levels (random and extrapolatory). All errors are reported on the testing dataset.}
\label{tab:mse-512-sdf-tenth}
\begin{tabular}{c c c c c c c}
\hline
 & \multicolumn{3}{c}{\textbf{Random}} & \multicolumn{3}{c}{\textbf{Extrapolatory}} \\ \hline
\textbf{Model} & \textbf{M1} & \textbf{M2} & \textbf{M3} & \textbf{M1} & \textbf{M2} & \textbf{M3} \\ \hline
\textbf{poseidon-L} & $\mathbf{45.0}$ & $\mathbf{56.6}$ & $22.0$ & $\mathbf{28.7}$ & $\mathbf{39.5}$ & $16.5$ \\ \hline
\textbf{poseidon-B} & $44.7$ & $54.2$ & $23.0$ & $25.9$ & $38.5$ & $26.9$ \\ \hline
\textbf{poseidon-T} & $43.8$ & $54.4$ & $26.4$ & $24.5$ & $39.4$ & $22.8$ \\ \hline
\textbf{scOT-L} & $42.9$ & $52.2$ & $28.6$ & $18.3$ & $33.6$ & $27.8$ \\ \hline
\textbf{scOT-B} & $43.1$ & $53.6$ & $24.8$ & $19.5$ & $32.8$ & $32.0$ \\ \hline
\textbf{scOT-T} & $42.7$ & $52.2$ & $30.4$ & $19.8$ & $33.9$ & $33.9$ \\ \hline
\textbf{CNO} & $32.7$ & $44.6$ & $25.0$ & $21.9$ & $34.6$ & $32.3$ \\ \hline
\textbf{FNO} & $32.0$ & $51.9$ & $35.6$ & $14.3$ & $27.9$ & $51.4$ \\ \hline
\textbf{WNO} & $17.9$ & $37.5$ & $0.0$ & $12.1$ & $25.6$ & $0.0$ \\ \hline
\textbf{Deeponet} & $35.0$ & $46.4$ & $\mathbf{51.8}$ & $18.2$ & $29.4$ & $\mathbf{57.0}$ \\ \hline
\textbf{geometric-deeponet} & $34.9$ & $47.9$ & $44.5$ & $18.6$ & $32.1$ & $48.7$ \\ \hline
\end{tabular}
\end{table}

\begin{table}[t!]
\centering
\small
\setlength\extrarowheight{2pt}
\caption{The score of SciML models trained on a subset of one-tenth of the dataset using the binary mask at two different difficulty levels (random and extrapolatory). All errors are reported on the testing dataset.}
\label{tab:mse-512-mask-tenth}
\begin{tabular}{c c c c c c c}
\hline
 & \multicolumn{3}{c}{\textbf{Random}} & \multicolumn{3}{c}{\textbf{Extrapolatory}} \\ \hline
\textbf{Model} & \textbf{M1} & \textbf{M2} & \textbf{M3} & \textbf{M1} & \textbf{M2} & \textbf{M3} \\ \hline
\textbf{poseidon-L} & $45.9$ & $57.5$ & $21.8$ & $\mathbf{30.2}$ & $47.0$ & $13.6$ \\ \hline
\textbf{poseidon-B} & $\mathbf{50.9}$ & $60.5$ & $25.0$ & $26.9$ & $42.7$ & $23.3$ \\ \hline
\textbf{poseidon-T} & $50.8$ & $\mathbf{60.9}$ & $30.3$ & $29.6$ & $\mathbf{47.3}$ & $27.1$ \\ \hline
\textbf{scOT-L} & $44.2$ & $54.4$ & $29.6$ & $18.0$ & $32.6$ & $26.2$ \\ \hline
\textbf{scOT-B} & $46.4$ & $56.0$ & $27.1$ & $19.8$ & $33.8$ & $30.3$ \\ \hline
\textbf{scOT-T} & $38.7$ & $50.7$ & $29.5$ & $18.9$ & $31.7$ & $35.5$ \\ \hline
\textbf{CNO} & $30.6$ & $42.5$ & $40.3$ & $18.4$ & $30.3$ & $40.8$ \\ \hline
\textbf{FNO} & $29.8$ & $50.9$ & $\mathbf{36.7}$ & $14.6$ & $28.4$ & $51.1$ \\ \hline
\textbf{WNO} & $21.4$ & $37.5$ & $0.0$ & $10.7$ & $25.5$ & $0.0$ \\ \hline
\textbf{Deeponet} & $37.1$ & $52.0$ & $36.1$ & $16.2$ & $28.5$ & $\mathbf{53.7}$ \\ \hline
\textbf{geometric-deeponet} & $38.6$ & $53.0$ & $36.0$ & $18.8$ & $32.1$ & $42.2$ \\ \hline
\end{tabular}
\end{table}

\clearpage

\section{Field Predictions} \label{sec:field-predictions}
\begin{figure}[h!]
    \begin{minipage}[b]{0.33\linewidth}
        \centering
        \includegraphics[width=\linewidth]{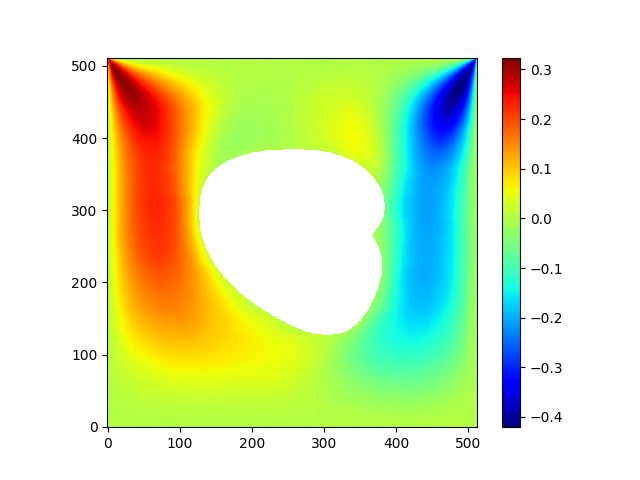}
        \textbf{Poseidon-T}
    \end{minipage}
    \begin{minipage}[b]{0.33\linewidth}
        \centering
        \includegraphics[width=\linewidth]{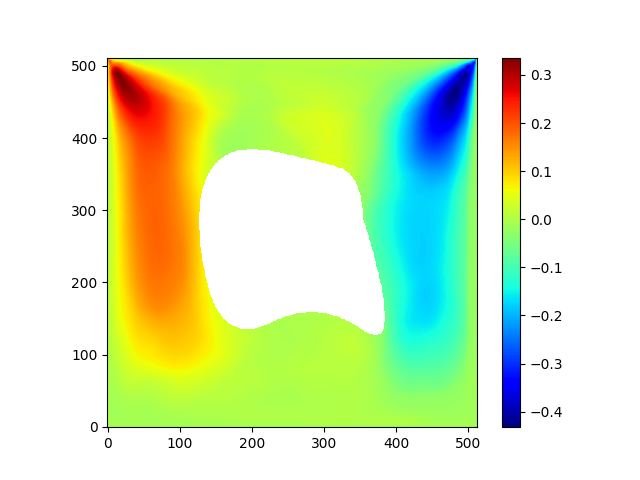}
        \textbf{CNO}
    \end{minipage}
    \begin{minipage}[b]{0.33\linewidth}
        \centering
        \includegraphics[width=\linewidth]{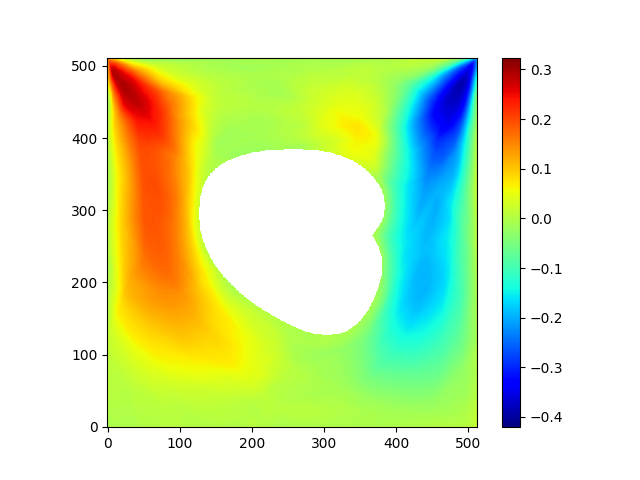}
        \textbf{geo-DeepONet}
    \end{minipage} \\
    \begin{minipage}[b]{0.33\linewidth}
        \centering
        \includegraphics[width=\linewidth]{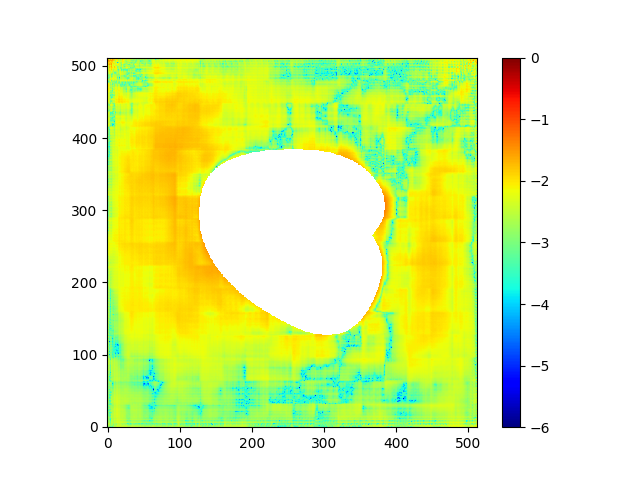}
        \textbf{error}
    \end{minipage}
    \begin{minipage}[b]{0.33\linewidth}
        \centering
        \includegraphics[width=\linewidth]{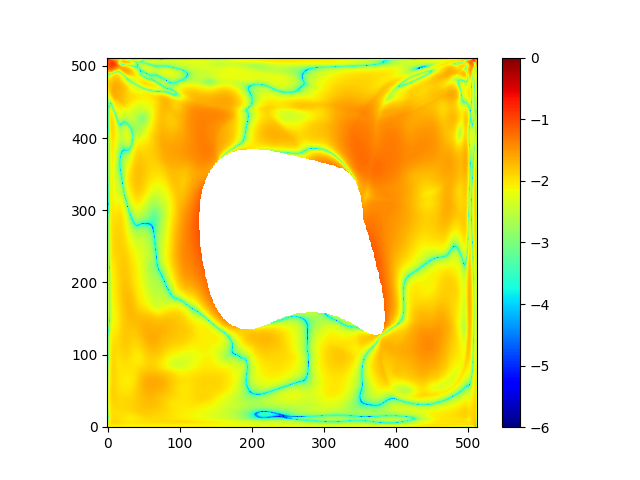}
        \textbf{error}
    \end{minipage}
    \begin{minipage}[b]{0.33\linewidth}
        \centering
        \includegraphics[width=\linewidth]{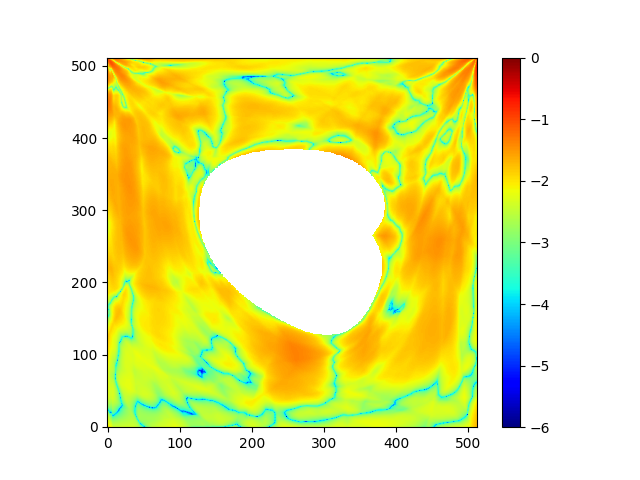}
        \textbf{error}
    \end{minipage} \\
    \begin{minipage}[b]{0.33\linewidth}
        \centering
        \includegraphics[width=\linewidth]{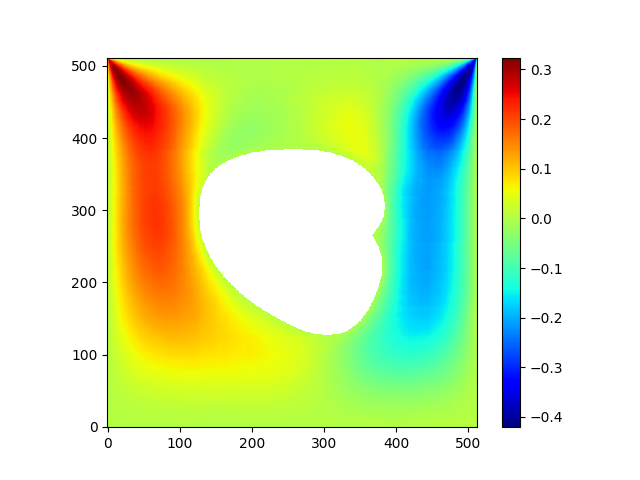}
        \textbf{Poseidon-T}
    \end{minipage}
    \begin{minipage}[b]{0.33\linewidth}
        \centering
        \includegraphics[width=\linewidth]{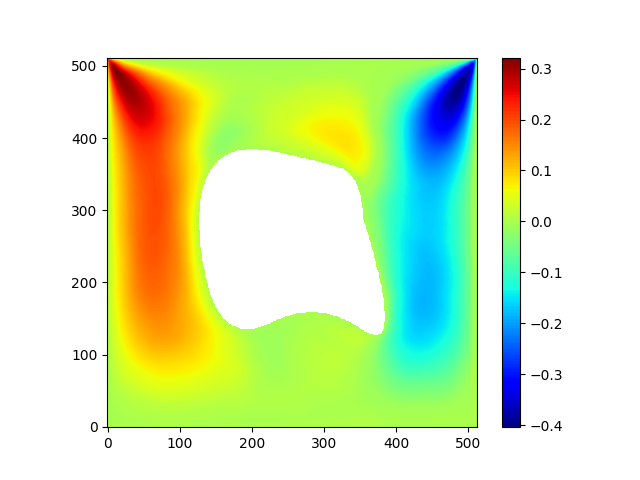}
        \textbf{CNO}
    \end{minipage}
    \begin{minipage}[b]{0.33\linewidth}
        \centering
        \includegraphics[width=\linewidth]{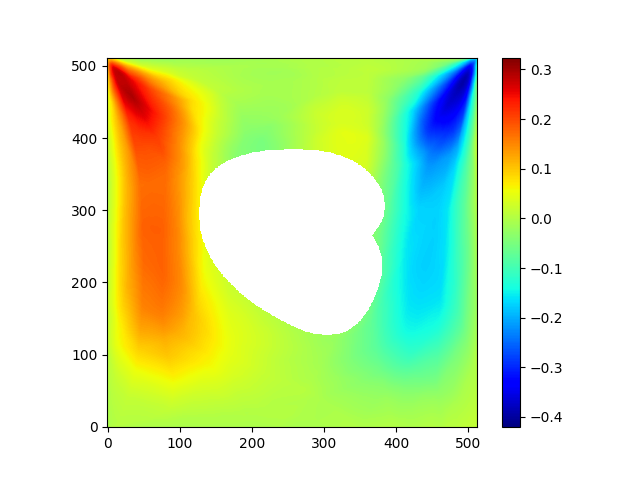}
        \textbf{geo-DeepONet}
    \end{minipage} \\
    \begin{minipage}[b]{0.33\linewidth}
        \centering
        \includegraphics[width=\linewidth]{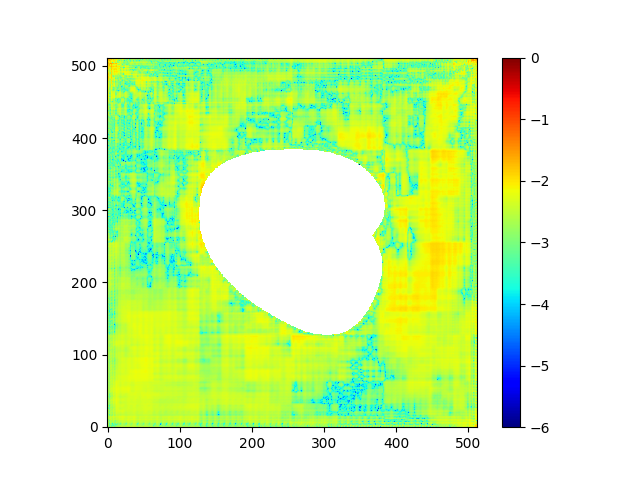}
        \textbf{error}
    \end{minipage}
    \begin{minipage}[b]{0.33\linewidth}
        \centering
        \includegraphics[width=\linewidth]{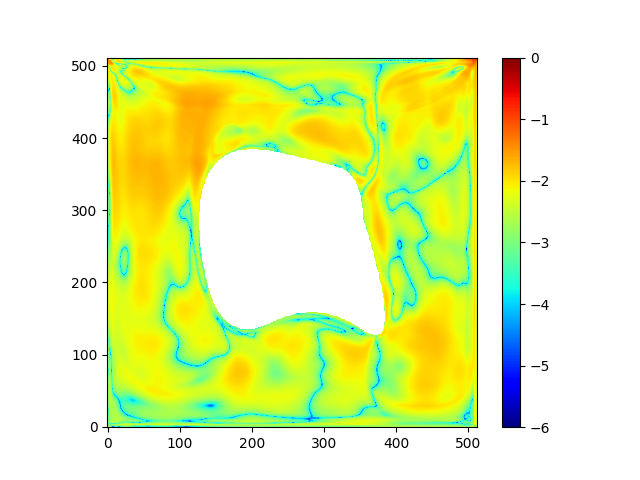}
        \textbf{error}
    \end{minipage}
    \begin{minipage}[b]{0.33\linewidth}
        \centering
        \includegraphics[width=\linewidth]{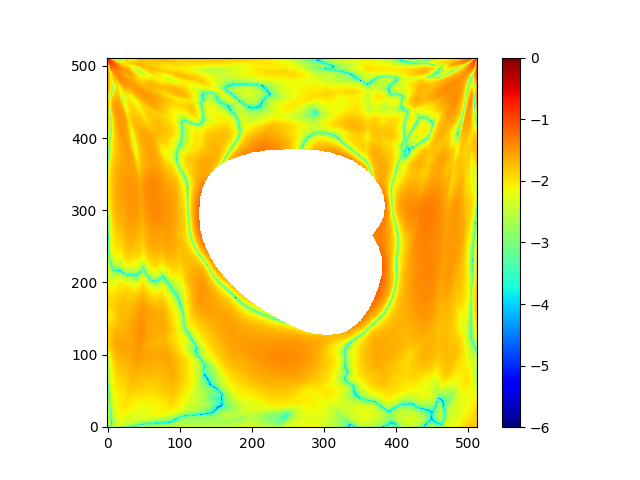}
        \textbf{error}
    \end{minipage}
    \caption{Comparison of model predictions and error distributions using SDF and mask representations. The first row shows predictions based on the SDF representation, followed by the second row displaying corresponding errors in the log scale relative to the ground truth. The third row presents predictions using the mask representation, with the fourth row displaying the corresponding log-scale errors relative to the ground truth.}
    \label{fig:sdf-vs-mask} 
\end{figure}

\begin{figure}[t!]
    \begin{minipage}[b]{0.33\linewidth}
        \centering
        \includegraphics[width=\linewidth]{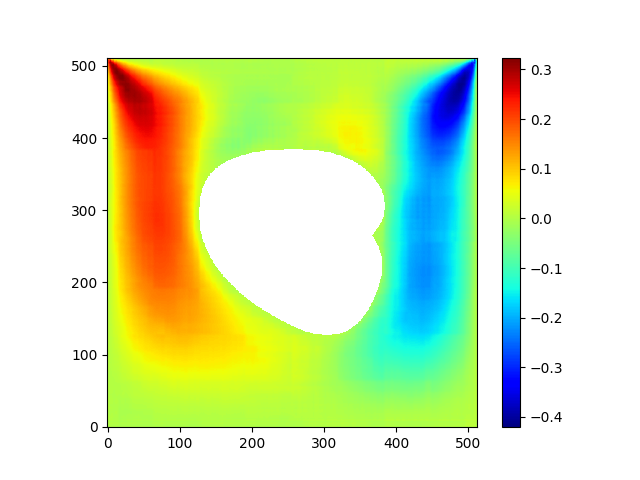}
        \textbf{Poseidon-T}
    \end{minipage}
    \begin{minipage}[b]{0.33\linewidth}
        \centering
        \includegraphics[width=\linewidth]{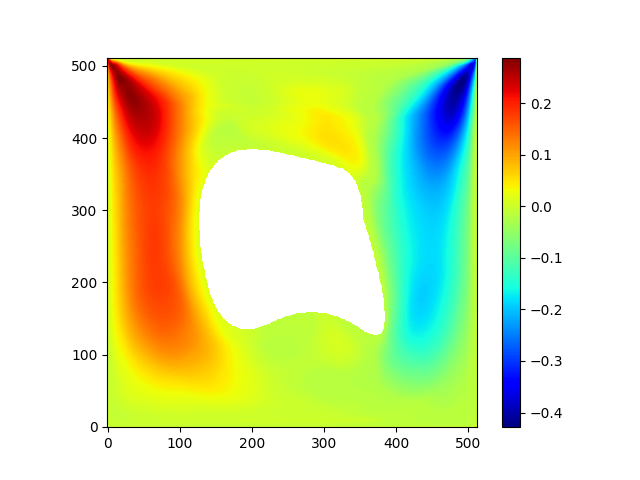}
        \textbf{CNO}
    \end{minipage}
    \begin{minipage}[b]{0.33\linewidth}
        \centering
        \includegraphics[width=\linewidth]{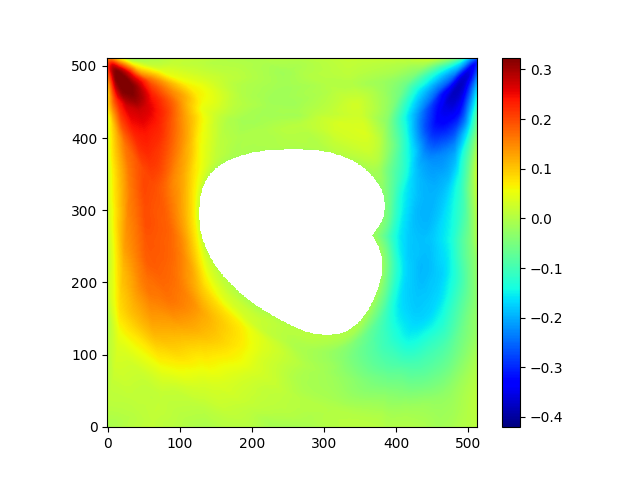}
        \textbf{geo-DeepONet}
    \end{minipage} \\
    \begin{minipage}[b]{0.33\linewidth}
        \centering
        \includegraphics[width=\linewidth]{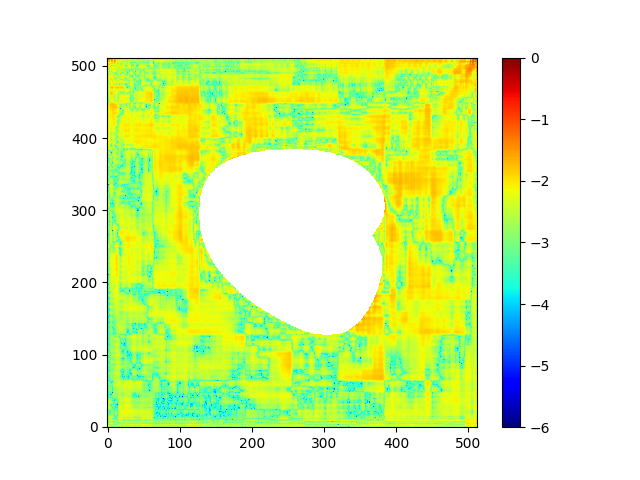}
        \textbf{error}
    \end{minipage}
    \begin{minipage}[b]{0.33\linewidth}
        \centering
        \includegraphics[width=\linewidth]{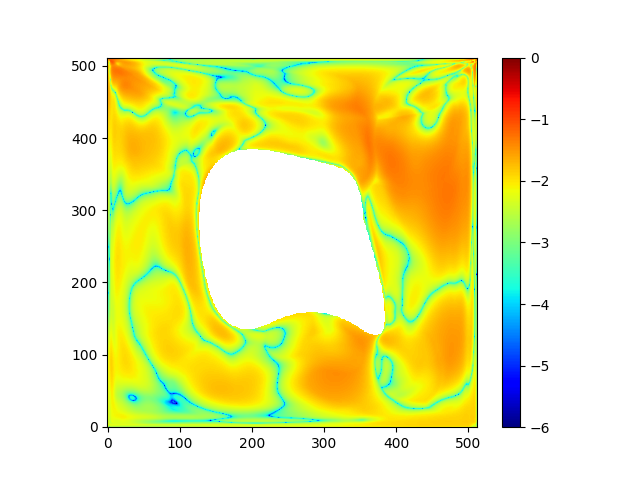}
        \textbf{error}
    \end{minipage}
    \begin{minipage}[b]{0.33\linewidth}
        \centering
        \includegraphics[width=\linewidth]{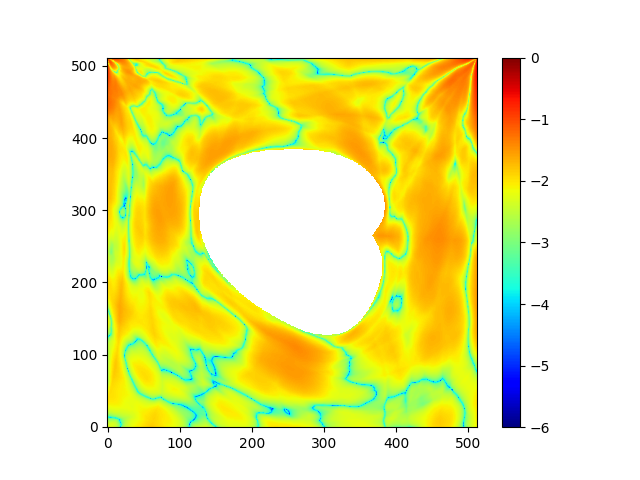}
        \textbf{error}
    \end{minipage} \\
    \begin{minipage}[b]{0.33\linewidth}
        \centering
        \includegraphics[width=\linewidth]{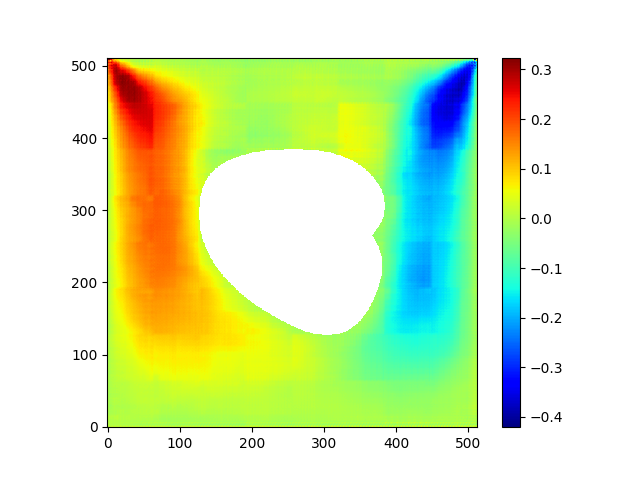}
        \textbf{Poseidon-T}
    \end{minipage}
    \begin{minipage}[b]{0.33\linewidth}
        \centering
        \includegraphics[width=\linewidth]{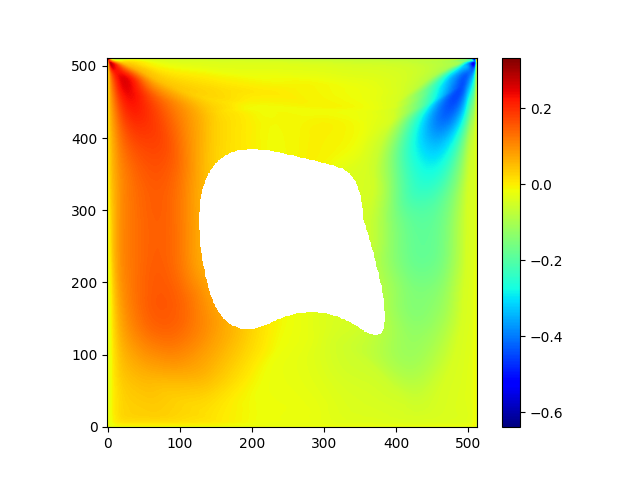}
        \textbf{CNO}
    \end{minipage}
    \begin{minipage}[b]{0.33\linewidth}
        \centering
        \includegraphics[width=\linewidth]{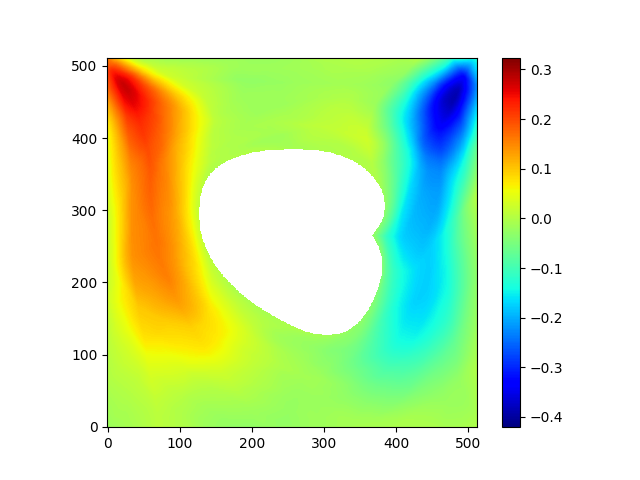}
        \textbf{geo-DeepONet}
    \end{minipage} \\
    \begin{minipage}[b]{0.33\linewidth}
        \centering
        \includegraphics[width=\linewidth]{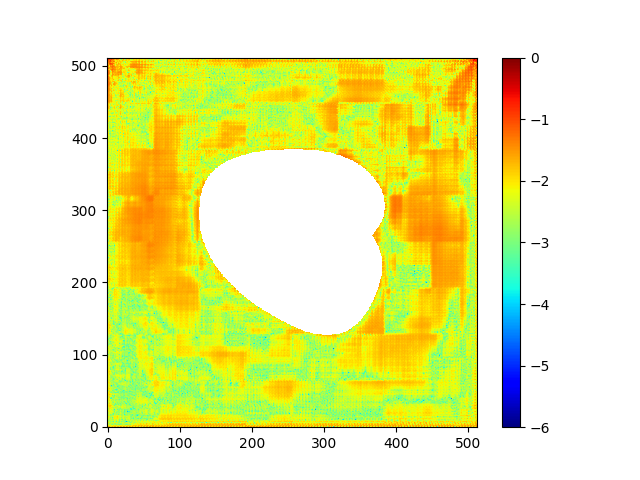}
        \textbf{error}
    \end{minipage}
    \begin{minipage}[b]{0.33\linewidth}
        \centering
        \includegraphics[width=\linewidth]{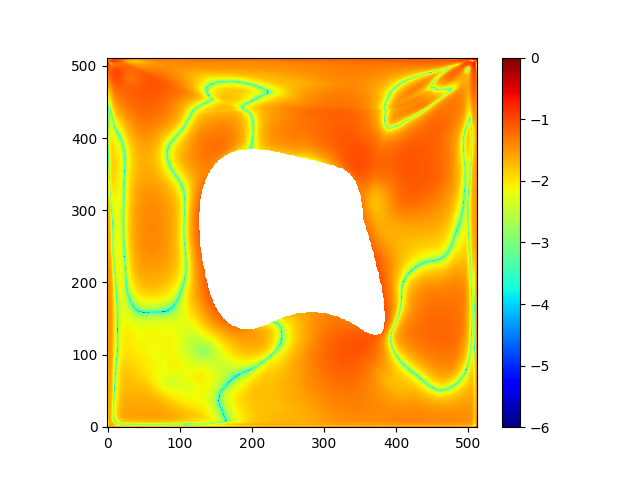}
        \textbf{error}
    \end{minipage}
    \begin{minipage}[b]{0.33\linewidth}
        \centering
        \includegraphics[width=\linewidth]{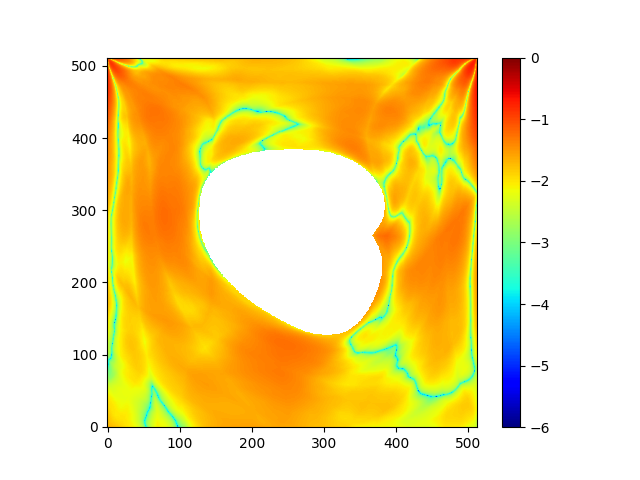}
        \textbf{error}
    \end{minipage}
    \caption{Comparison of model predictions and error distributions for training sample sizes of 800 and 240, highlighting data sufficiency. The first row shows predictions using a training sample size of 800, featuring Geometric-DeepONet with the Signed Distance Field (SDF) representation and CNO and Poseidon-T using the binary mask, followed by the second row displaying corresponding errors in the log scale relative to the ground truth. The third row presents predictions using a sample size of 300, featuring Geometric-DeepONet with the Signed Distance Field (SDF) representation and CNO and Poseidon-T using the binary mask, with the fourth row displaying the corresponding log-scale errors relative to the ground truth.}
    \label{fig:1000-vs-300} 
\end{figure}

\begin{figure}[t!]
    \begin{minipage}[b]{0.33\linewidth}
        \centering
        \includegraphics[width=\linewidth]{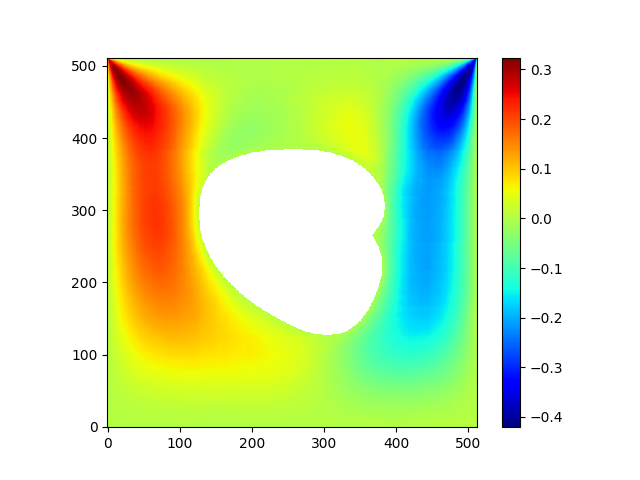}
        \textbf{Poseidon-T}
    \end{minipage}
    \begin{minipage}[b]{0.33\linewidth}
        \centering
        \includegraphics[width=\linewidth]{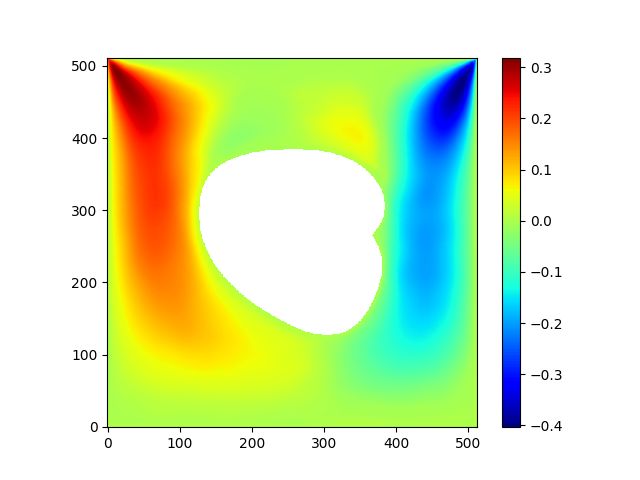}
        \textbf{CNO}
    \end{minipage}
    \begin{minipage}[b]{0.33\linewidth}
        \centering
        \includegraphics[width=\linewidth]{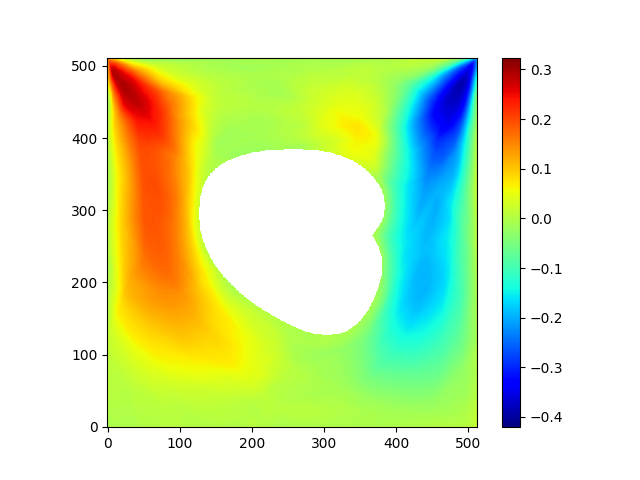}
        \textbf{geo-DeepONet}
    \end{minipage} \\
    \begin{minipage}[b]{0.33\linewidth}
        \centering
        \includegraphics[width=\linewidth]{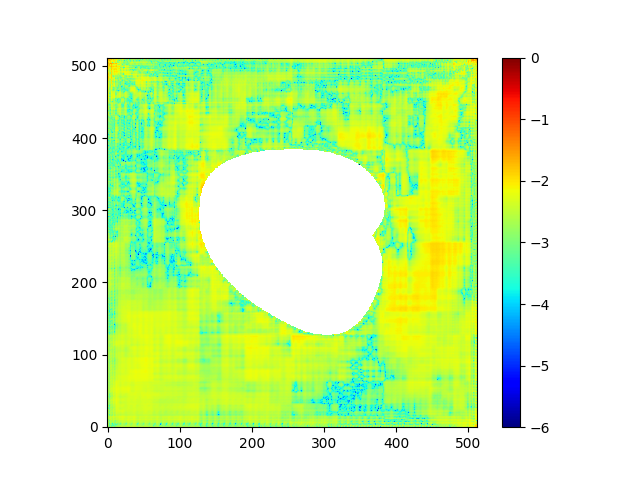}
        \textbf{error}
    \end{minipage}
    \begin{minipage}[b]{0.33\linewidth}
        \centering
        \includegraphics[width=\linewidth]{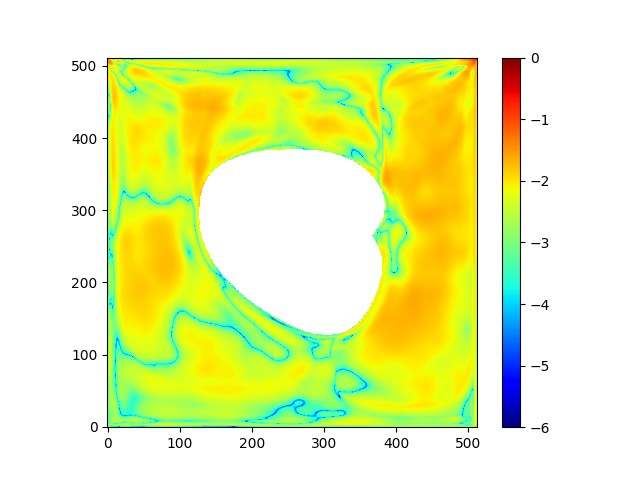}
        \textbf{error}
    \end{minipage}
    \begin{minipage}[b]{0.33\linewidth}
        \centering
        \includegraphics[width=\linewidth]{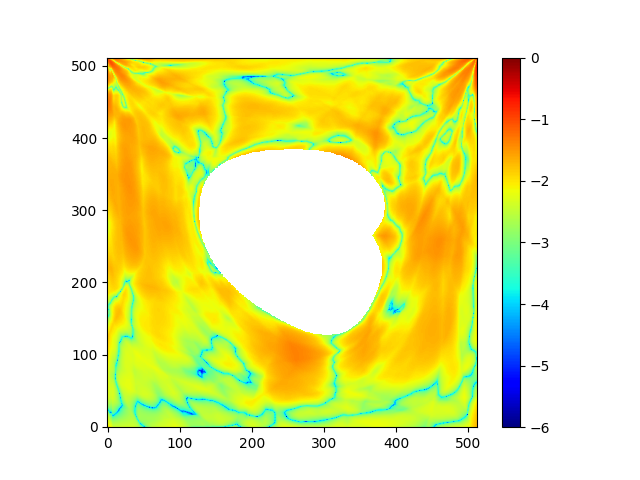}
        \textbf{error}
    \end{minipage} \\
    \begin{minipage}[b]{0.33\linewidth}
        \centering
        \includegraphics[width=\linewidth]{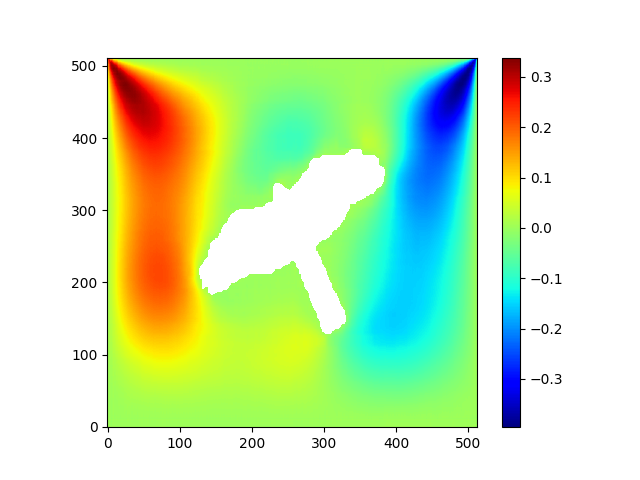}
        \textbf{Poseidon-T}
    \end{minipage}
    \begin{minipage}[b]{0.33\linewidth}
        \centering
        \includegraphics[width=\linewidth]{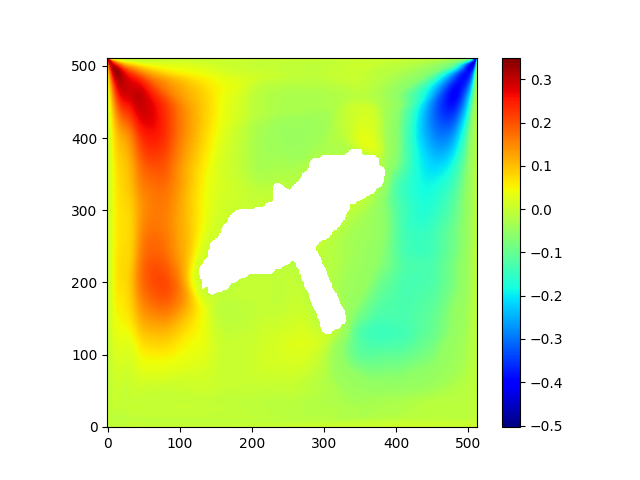}
        \textbf{CNO}
    \end{minipage}
    \begin{minipage}[b]{0.33\linewidth}
        \centering
        \includegraphics[width=\linewidth]{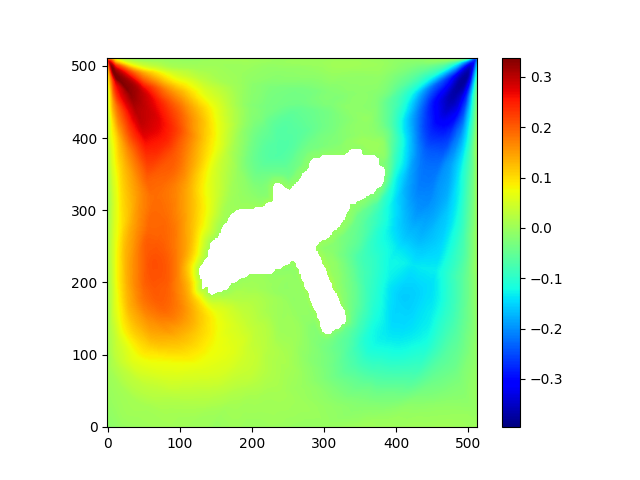}
        \textbf{geo-DeepONet}
    \end{minipage} \\
    \begin{minipage}[b]{0.33\linewidth}
        \centering
        \includegraphics[width=\linewidth]{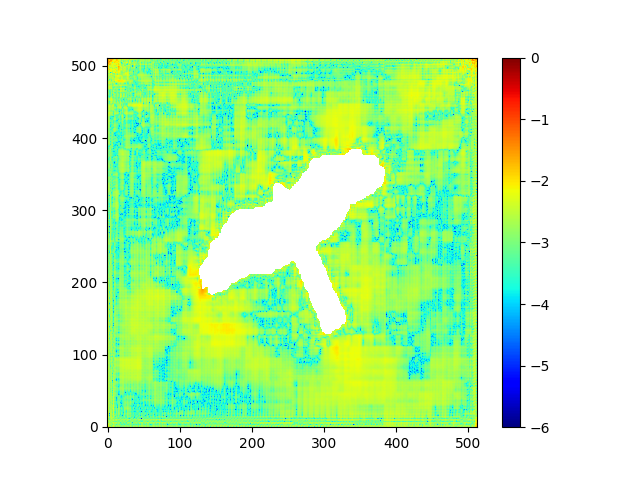}
        \textbf{error}
    \end{minipage}
    \begin{minipage}[b]{0.33\linewidth}
        \centering
        \includegraphics[width=\linewidth]{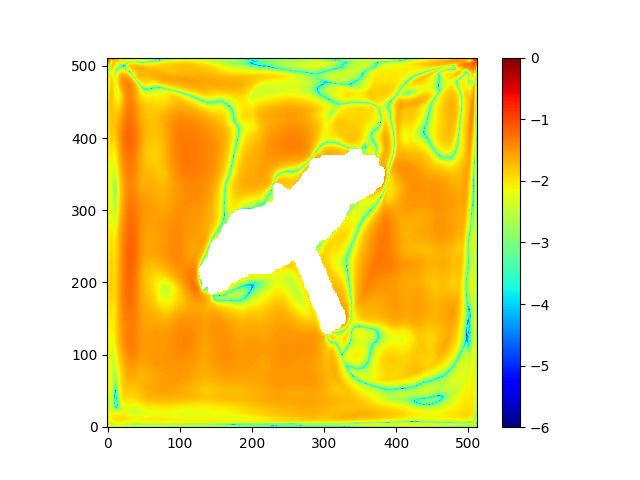}
        \textbf{error}
    \end{minipage}
    \begin{minipage}[b]{0.33\linewidth}
        \centering
        \includegraphics[width=\linewidth]{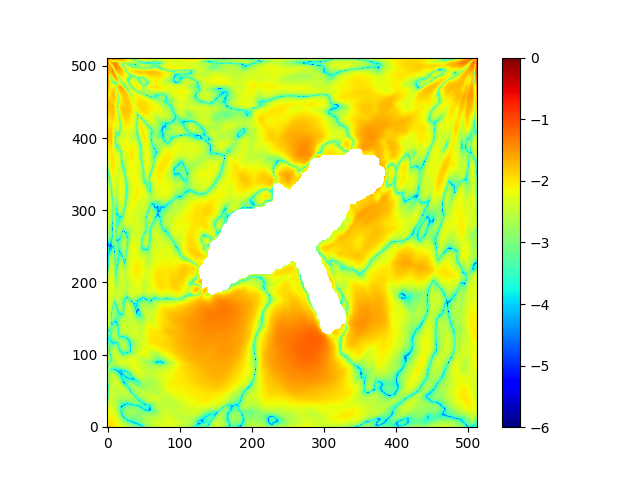}
        \textbf{error}
    \end{minipage}
   \caption{Comparison of model predictions and error distributions for random train/test splitting and extrapolatory train/test splitting, highlighting the impact of different data splits. The first row shows predictions using the random split, featuring Geometric-DeepONet with the Signed Distance Field (SDF) representation and CNO and Poseidon-T using the binary mask, followed by the second row displaying corresponding errors in the log scale relative to the ground truth. The third row presents predictions using the extrapolatory split, featuring Geometric-DeepONet with the Signed Distance Field (SDF) representation and CNO and Poseidon-T using the binary mask, with the fourth row displaying the corresponding log-scale errors relative to the ground truth.}
    \label{fig:random-vs-extrapolatory} 
\end{figure}

In \figref{fig:sdf-vs-mask}, \figref{fig:1000-vs-300}, and \figref{fig:random-vs-extrapolatory}, we present field velocity predictions in the y-direction $v$ for a representative sample using scientific machine learning models (CNO, geometric-DeepONet, Poseidon-T). 

\section{Residual Calculation} \label{sec:residual-calculation}

The finite element mesh, denoted by \( \mathcal{K}_h \), defines the computational discretization of the domain \( \Omega \). Integrations over \( \Omega \) are computed by integrating within each finite element \( e \) and summing over all elements in \( \mathcal{K}_h \). For instance, the \( L_2 \)-norm of the velocity component \( u \) over \( \Omega \) is defined as:
\[
\| u \|_{L_2(\Omega)} = \left( \int_\Omega u^2 \, d\Omega \right)^{1/2} = \left( \sum_{e \in \mathcal{K}_h} \int_e u^2 \, de \right)^{1/2}
\]

Similarly, for the Navier-Stokes momentum residuals, we compute the element-wise residuals. These residuals, denoted by \( r_{\text{x}} \) and \( r_{\text{y}} \), represent the momentum conservation equations in the \( x \)- and \( y \)-directions, respectively. 

\[
r_{\text{x}} = \frac{\partial u_x}{\partial t} + u \cdot \nabla u_x - \eta \nabla^2 u_x + \frac{\partial p}{\partial x} 
\]
\[
r_{\text{y}} = \frac{\partial u_y}{\partial t} + u \cdot \nabla u_y - \eta \nabla^2 u_y + \frac{\partial p}{\partial y} 
\]

These element-wise residuals are calculated as follows:

\[
\| r_{\text{x}} \|_{L_2(e)} = \left( \int_e r_{\text{x}}^2 \, de \right)^{1/2}
\]
\[
\| r_{\text{y}} \|_{L_2(e)} = \left( \int_e r_{\text{y}}^2 \, de \right)^{1/2}
\]
The total residual \( r_{\text{Total}} \) (denoted $M3$) is then obtained by adding the contributions of each element:
\[
r_{\text{Total}} = \sum_{e \in \mathcal{K}_h} \left( \| r_{\text{x}} \|^2_{L_2(e)} + \| r_{\text{y}} \|^2_{L_2(e)} \right)
\]

In addition to the residuals, we compute the errors in the spatial derivatives of the velocity components, \( u \) and \( v \), by comparing the predicted derivatives with their ground truth gradient values. These errors, denoted \( u'_{\text{error}} \) and \( v'_{\text{error}} \), respectively, are defined as:

\[
u'_{\text{error}} = \left( \frac{\partial u}{\partial x} - \frac{\partial u_{\text{true}}}{\partial x} \right)^2 + \left( \frac{\partial u}{\partial y} - \frac{\partial u_{\text{true}}}{\partial y} \right)^2
\]
\[
v'_{\text{error}} = \left( \frac{\partial v}{\partial x} - \frac{\partial v_{\text{true}}}{\partial x} \right)^2 + \left( \frac{\partial v}{\partial y} - \frac{\partial v_{\text{true}}}{\partial y} \right)^2
\]

These errors are integrated element-wise to calculate their \( L_2 \)-norms over each element \( e \):

\[
\| u'_{\text{error}} \|_{L_2(e)} = \left( \int_e u'_{\text{error}} \, de \right)^{1/2}, \quad \| v'_{\text{error}} \|_{L_2(e)} = \left( \int_e v'_{\text{error}} \, de \right)^{1/2}
\]

The residual metric assesses the accuracy of the velocity and pressure fields within each element and measures the adherence to momentum conservation throughout the domain, while the derivative error quantifies discrepancies in velocity gradients compared to ground truth gradient values.

\begin{figure}[t!]
    \begin{minipage}[b]{0.33\linewidth}
        \centering
        \includegraphics[width=\linewidth]{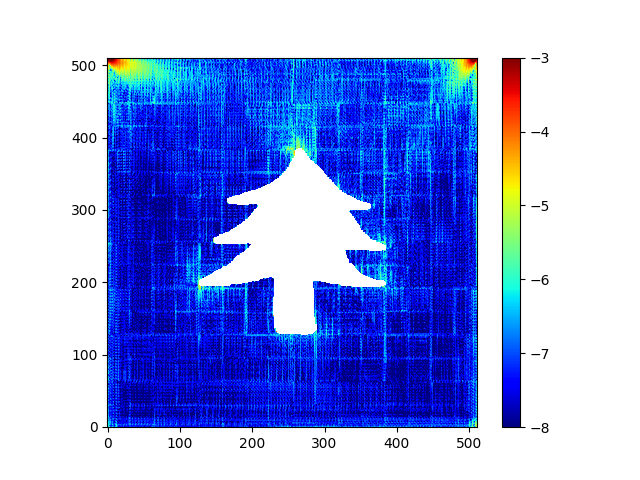}
        \textbf{Poseidon-T using SDF ($r_x$)}
    \end{minipage}
    \begin{minipage}[b]{0.33\linewidth}
        \centering
        \includegraphics[width=\linewidth]{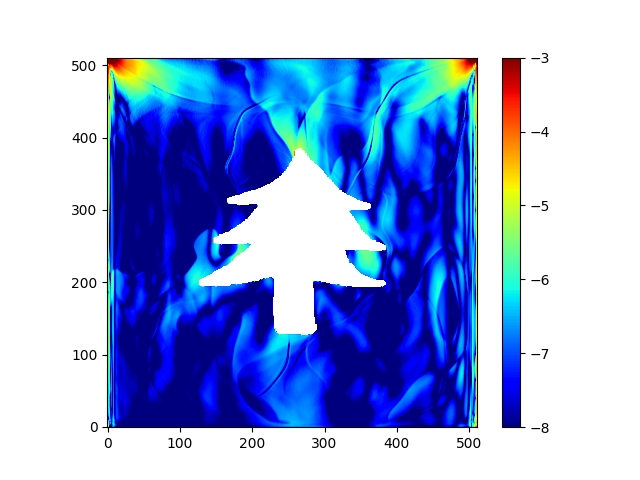}
        \textbf{CNO using SDF ($r_x$)}
    \end{minipage}
    \begin{minipage}[b]{0.33\linewidth}
        \centering
        \includegraphics[width=\linewidth]{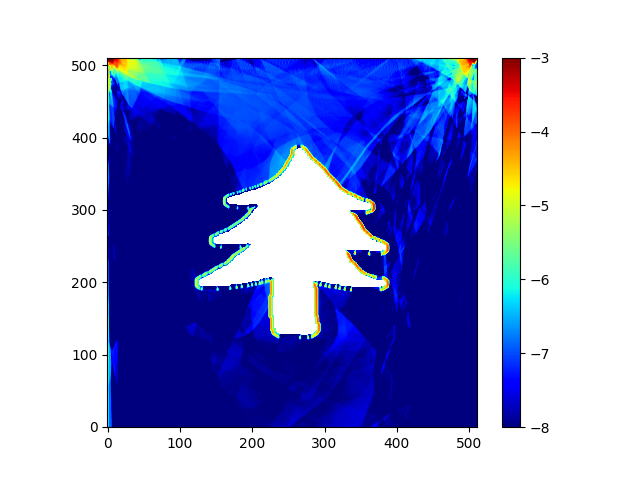}
        \textbf{DeepONet using SDF ($r_x$)}
    \end{minipage} \\
    \begin{minipage}[b]{0.33\linewidth}
        \centering
        \includegraphics[width=\linewidth]{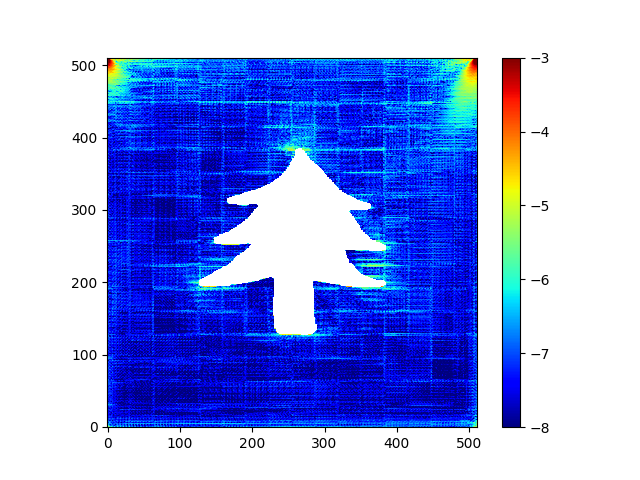}
        \textbf{Poseidon-T using SDF ($r_y$)}
    \end{minipage}
    \begin{minipage}[b]{0.33\linewidth}
        \centering
        \includegraphics[width=\linewidth]{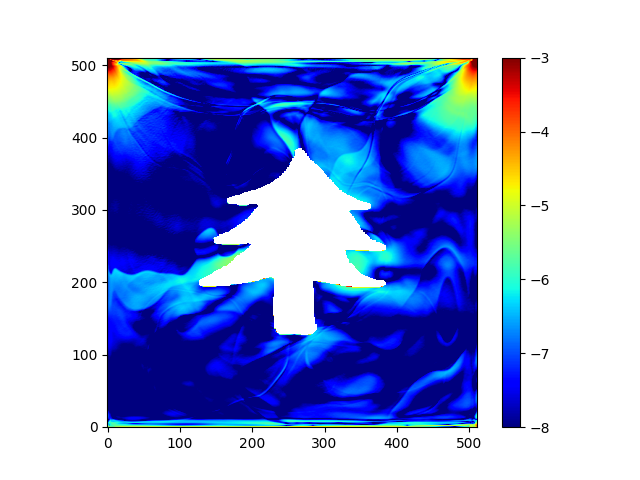}
        \textbf{CNO using SDF ($r_y$)}
    \end{minipage}
    \begin{minipage}[b]{0.33\linewidth}
        \centering
        \includegraphics[width=\linewidth]{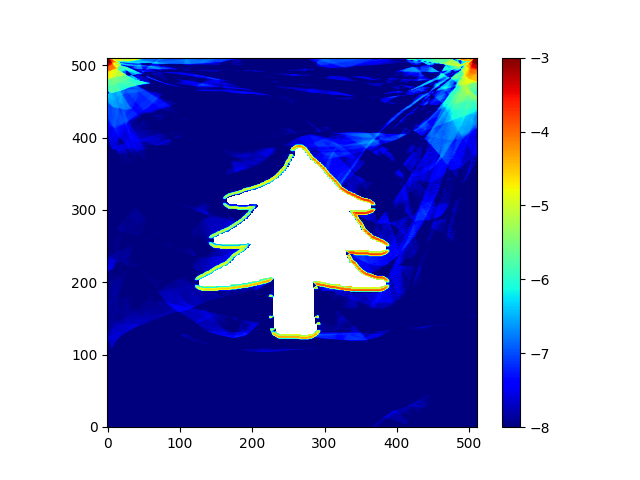}
        \textbf{DeepONet using SDF ($r_y$)}
    \end{minipage} \\
    \begin{minipage}[b]{0.33\linewidth}
        \centering
        \includegraphics[width=\linewidth]{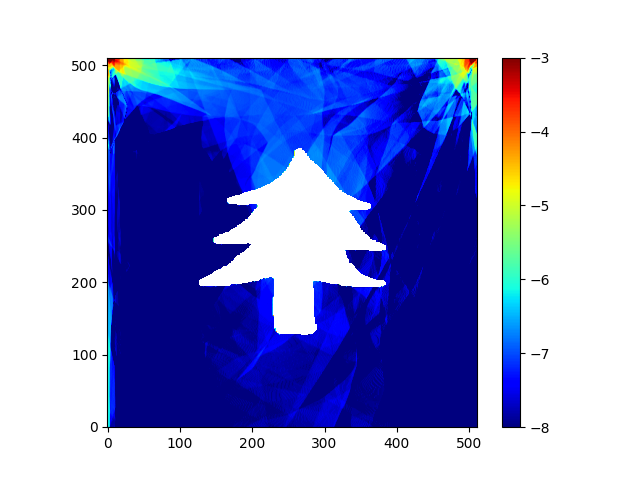}
        \textbf{Poseidon-T using mask ($r_x$)}
    \end{minipage}
    \begin{minipage}[b]{0.33\linewidth}
        \centering
        \includegraphics[width=\linewidth]{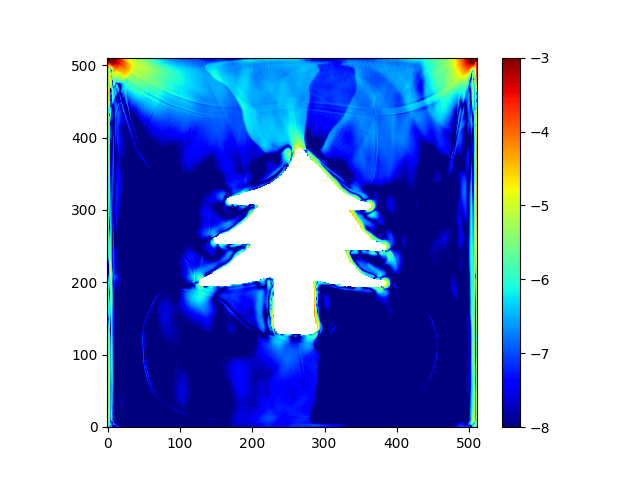}
        \textbf{CNO using mask ($r_x$)}
    \end{minipage}
    \begin{minipage}[b]{0.33\linewidth}
        \centering
        \includegraphics[width=\linewidth]{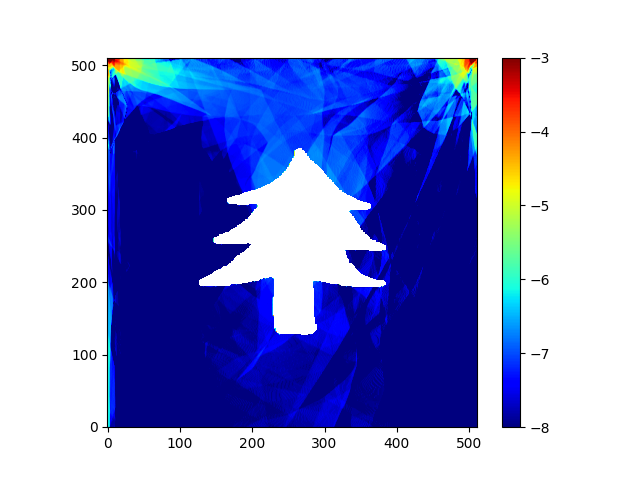}
        \textbf{DeepONet using mask ($r_x$)}
    \end{minipage} \\
    \begin{minipage}[b]{0.33\linewidth}
        \centering
        \includegraphics[width=\linewidth]{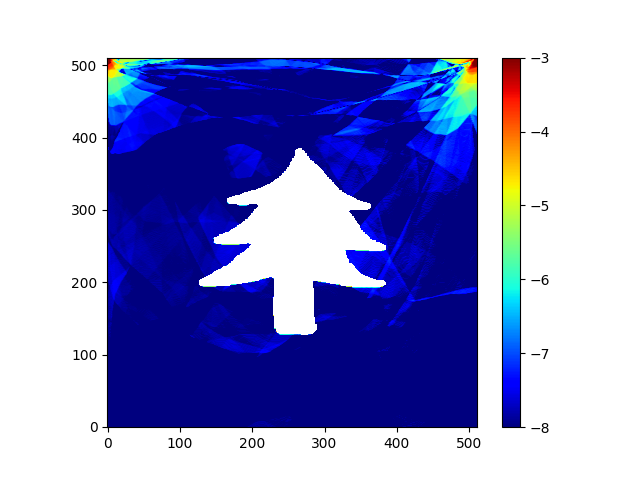}
        \textbf{Poseidon-T using mask ($r_y$)}
    \end{minipage}
    \begin{minipage}[b]{0.33\linewidth}
        \centering
        \includegraphics[width=\linewidth]{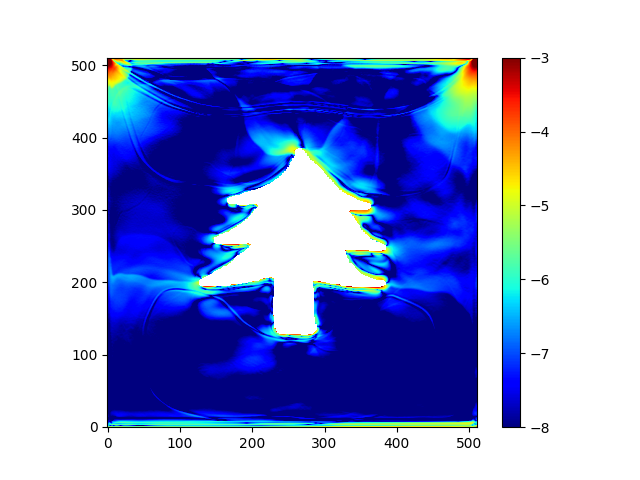}
        \textbf{CNO using mask ($r_y$)}
    \end{minipage}
    \begin{minipage}[b]{0.33\linewidth}
        \centering
        \includegraphics[width=\linewidth]{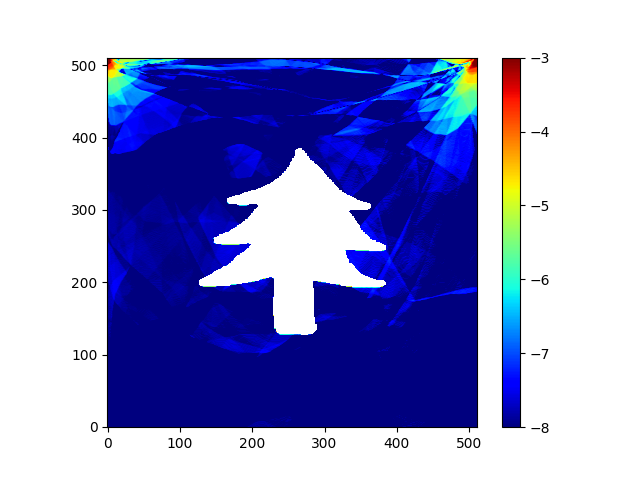}
        \textbf{DeepONet using mask ($r_y$)}
    \end{minipage}
    \caption{Comparison of model residual distributions in the x and y directions ($r_x$, $r_y$) for a single sample using random train/test splitting, highlighting the impact of different geometry representations. The first two rows show $r_x$ and $r_y$ (respectively) in the log scale using the signed distance field, featuring poseidon-T, CNO, and DeepONet. The third and fourth rows present $r_x$ and $r_y$ (respectively) in the log scale using the binary mask, featuring poseidon-T, CNO, and DeepONet.}
    \label{fig:residuals-mask-vs-sdf} 
\end{figure}

\begin{figure}[t!]
    \begin{minipage}[b]{0.33\linewidth}
        \centering
        \includegraphics[width=\linewidth]{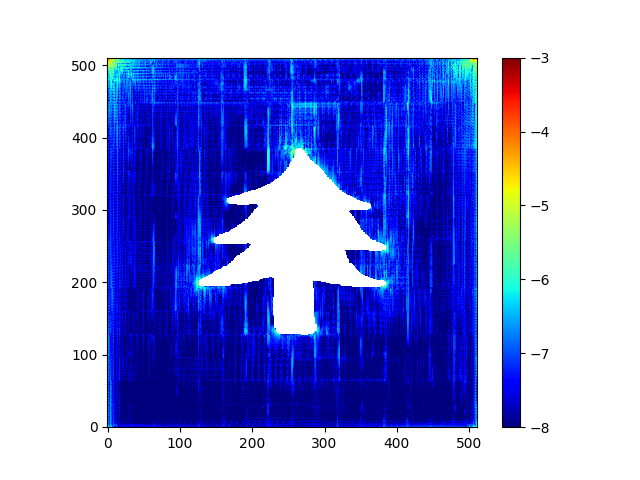}
        \textbf{Poseidon-T using SDF ($u_\text{error}$)}
    \end{minipage}
    \begin{minipage}[b]{0.33\linewidth}
        \centering
        \includegraphics[width=\linewidth]{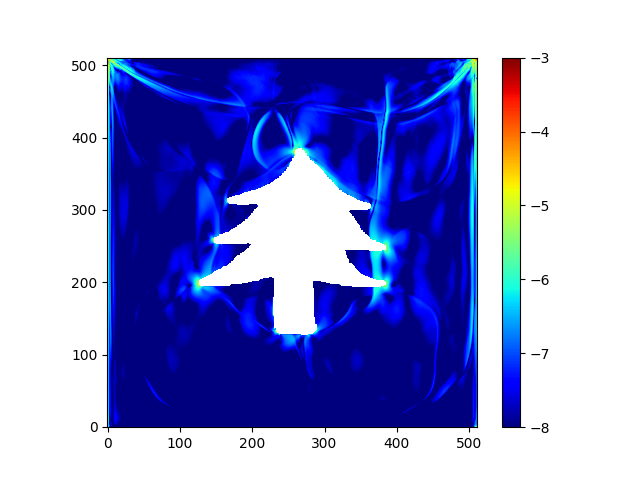}
        \textbf{CNO using SDF ($u_\text{error}$)}
    \end{minipage}
    \begin{minipage}[b]{0.33\linewidth}
        \centering
        \includegraphics[width=\linewidth]{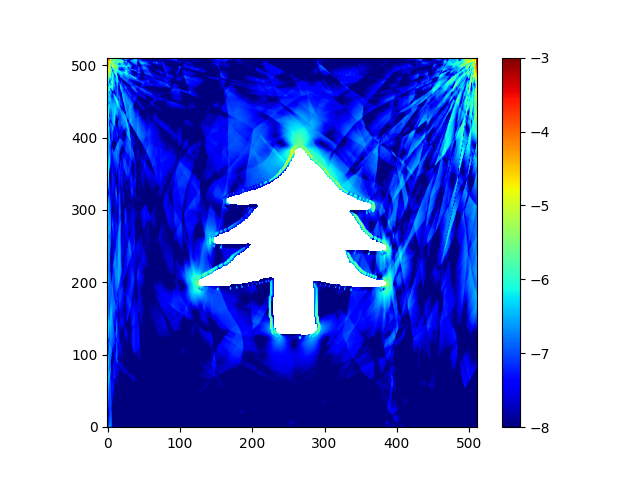}
        \textbf{DeepONet using SDF ($u_\text{error}$)}
    \end{minipage} \\
    \begin{minipage}[b]{0.33\linewidth}
        \centering
        \includegraphics[width=\linewidth]{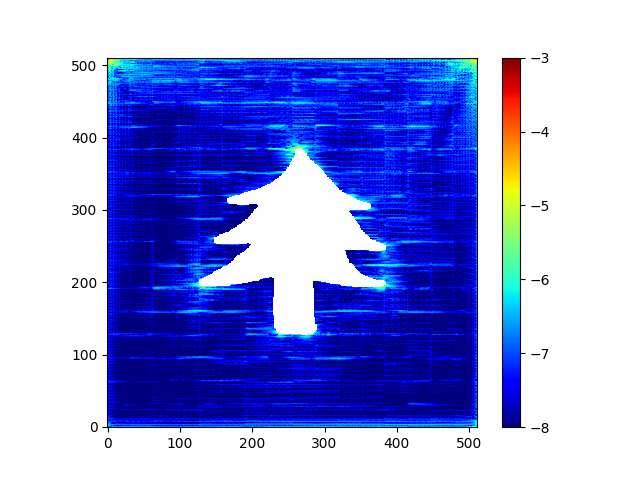}
        \textbf{Poseidon-T using SDF ($v_\text{error}$)}
    \end{minipage}
    \begin{minipage}[b]{0.33\linewidth}
        \centering
        \includegraphics[width=\linewidth]{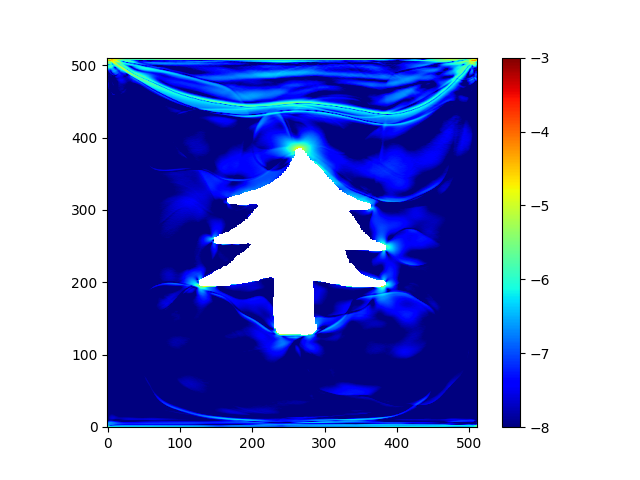}
        \textbf{CNO using SDF ($v_\text{error}$)}
    \end{minipage}
    \begin{minipage}[b]{0.33\linewidth}
        \centering
        \includegraphics[width=\linewidth]{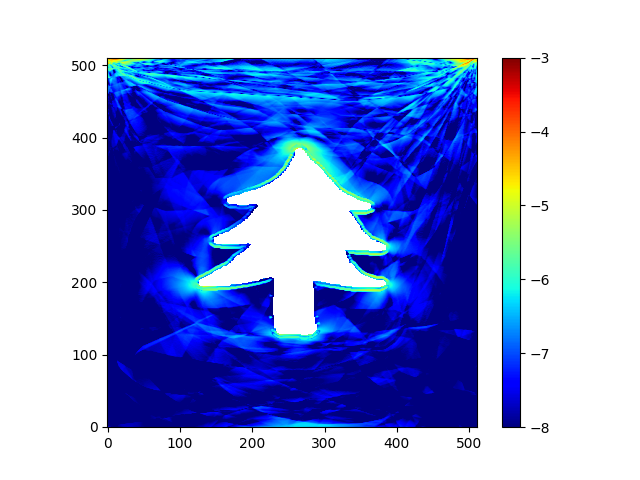}
        \textbf{DeepONet using SDF ($v_\text{error}$)}
    \end{minipage} \\
    \begin{minipage}[b]{0.33\linewidth}
        \centering
        \includegraphics[width=\linewidth]{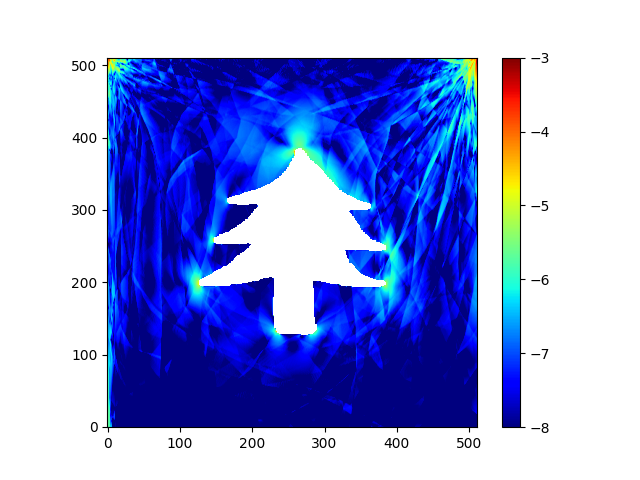}
        \textbf{Poseidon-T using mask ($u_\text{error}$)}
    \end{minipage}
    \begin{minipage}[b]{0.33\linewidth}
        \centering
        \includegraphics[width=\linewidth]{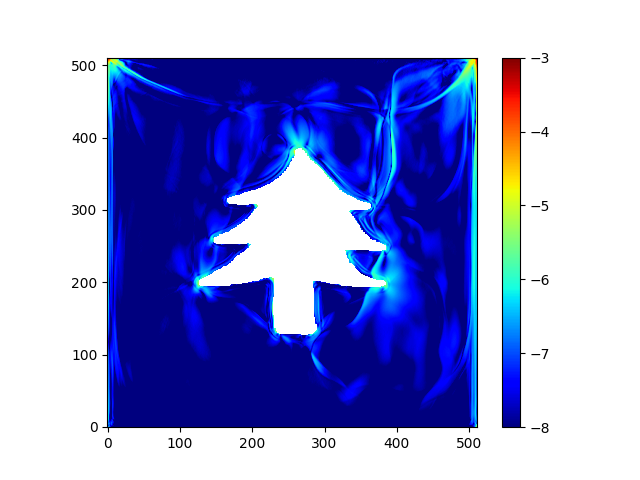}
        \textbf{CNO using mask ($u_\text{error}$)}
    \end{minipage}
    \begin{minipage}[b]{0.33\linewidth}
        \centering
        \includegraphics[width=\linewidth]{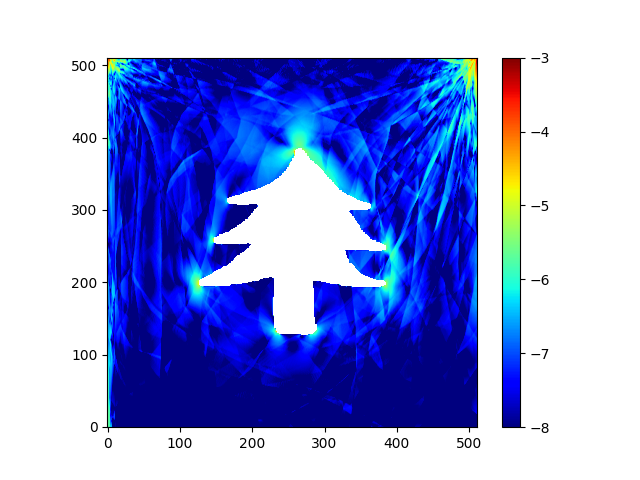}
        \textbf{DeepONet using mask ($u_\text{error}$)}
    \end{minipage} \\
    \begin{minipage}[b]{0.33\linewidth}
        \centering
        \includegraphics[width=\linewidth]{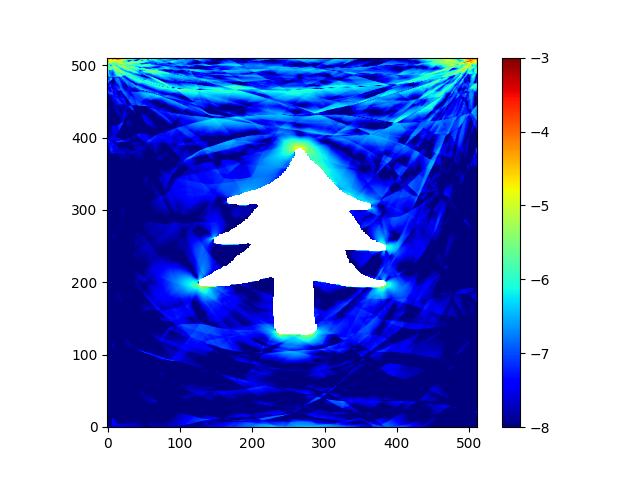}
        \textbf{Poseidon-T using mask ($v_\text{error}$)}
    \end{minipage}
    \begin{minipage}[b]{0.33\linewidth}
        \centering
        \includegraphics[width=\linewidth]{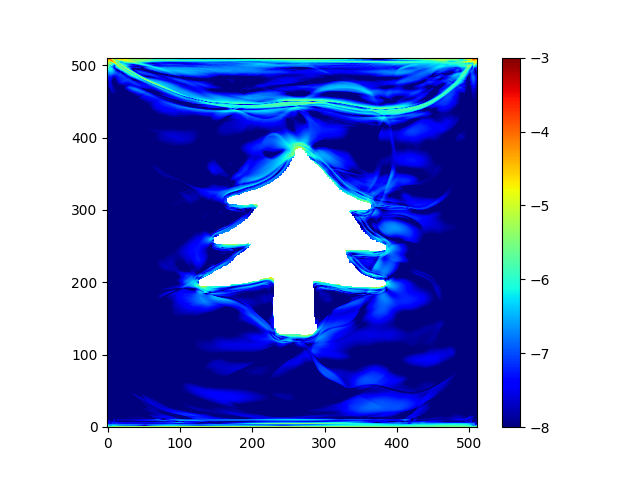}
        \textbf{CNO using mask ($v_\text{error}$)}
    \end{minipage}
    \begin{minipage}[b]{0.33\linewidth}
        \centering
        \includegraphics[width=\linewidth]{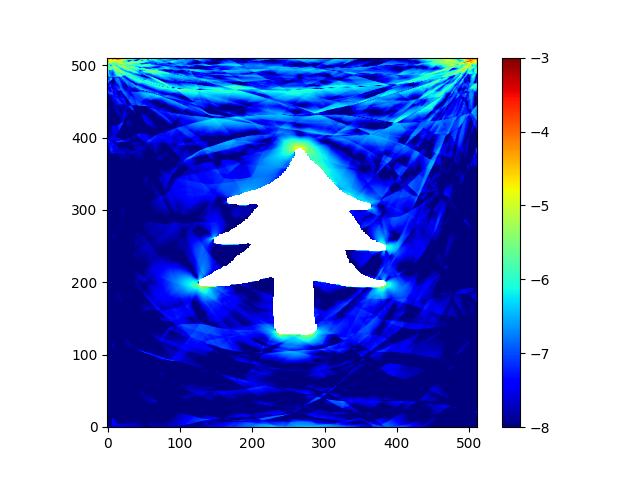}
        \textbf{DeepONet using mask ($v_\text{error}$)}
    \end{minipage}
    \caption{Comparison of model velocity gradient error distributions in the x and y directions ($u'_\text{error}$, $v'_\text{error}$) for a single sample using random train/test splitting, highlighting the impact of different geometry representations. The first two rows show $u'_\text{error}$ and $v'_\text{error}$ (respectively) in the log scale using the signed distance field, featuring poseidon-T, CNO, and DeepONet. The third and fourth rows present $u'_\text{error}$ and $v'_\text{error}$ (respectively) in the log scale using the binary mask, featuring poseidon-T, CNO, and DeepONet.}
    \label{fig:vel-error-mask-vs-sdf} 
\end{figure}

\clearpage

\section{Computational Performance} \label{sec:model-size}

The numerical efficiency of the different SciML models, including model size, training time, and inference time, are presented in \tabref{tab:model-performance}. The scOT and Poseidon models exhibit similar values across these metrics due to their shared architecture. Among all models, FNO stands out as the fastest model to train, completing training in just 2.29 hours, owing to its relatively simple architecture and smaller number of parameters compared to other large models. DeepONet and geometric-DeepONet are the fastest in inference time, with DeepONet requiring less than a second per sample due to its small model size. On the other hand, the larger models, such as WNO and the base and large versions of scOT and poseidon, take significantly longer to train due to their substantial number of parameters. This trade-off between model size and computational requirements highlights the varying computational efficiencies of these architectures and their suitability for different applications.

\begin{table}[!h]
\centering
\small
\setlength\extrarowheight{2pt}
\caption{Performance metrics of SciML models: Model size (in million parameters), training time (in hours) on the full dataset of 2400 samples, and inference time (in seconds) evaluated on the randomly split test dataset containing 600 samples.}
\label{tab:model-performance}
\begin{tabular}{c c c c}
\hline
\textbf{Model} & \textbf{Model Size (Million)} & \textbf{Training Time (hr)} & \textbf{Inference Time (sec)} \\ \hline
\textbf{poseidon-L} & $628$ & $21.94$ & $80.41$ \\ \hline
\textbf{poseidon-B} & $157$ & $13.25$ & $65.93$ \\ \hline
\textbf{poseidon-T} & $20.7$ & $6.25$ & $38.81$ \\ \hline
\textbf{scOT-L} & $628$ & $22.50$ & $81.30$ \\ \hline
\textbf{scOT-B} & $157$ & $12.93$ & $67.48$ \\ \hline
\textbf{scOT-T} & $20.7$ & $6.48$ & $49.25$ \\ \hline
\textbf{CNO} & $11.7$ & $6.04$ & $12.02$ \\ \hline
\textbf{FNO} & $10.9$ & $\mathbf{2.29}$ & $11.46$ \\ \hline
\textbf{WNO} & $94.7$ & $10.17$ & $29.93$ \\ \hline
\textbf{Deeponet} & $\mathbf{0.9}$ & $5.86$ & $\mathbf{0.61}$ \\ \hline
\textbf{geometric-deeponet} & $2.1$ & $6.67$ & $1.04$ \\ \hline
\end{tabular}
\end{table}

\clearpage

\section{Model Hyperparameters} 
\label{sec:hyperparameter}

This section provides a detailed overview of the hyperparameters used for training each of the 11 scientific machine learning (SciML) models. These hyperparameters were selected based on extensive tuning to optimize performance for the datasets used in this study.

\subsection{Convolution Neural Operator}
The Convolution Neural Operator (CNO) was trained with the following hyperparameters:
\begin{itemize}
    \item Number of residual blocks (\texttt{N\_res}): 4
    \item Learning rate (\texttt{lr}): 0.001
    \item Number of layers (\texttt{n\_layers}): 4
\end{itemize}

\subsection{Fourier Neural Operator}
The Fourier Neural Operator (FNO) was trained with the following hyperparameters:
\begin{itemize}
    \item Hidden channels (\texttt{hidden\_channels}): 16
    \item Learning rate (\texttt{lr}): 0.0001
    \item Number of layers (\texttt{n\_layers}): 10
    \item Number of Fourier modes (\texttt{n\_modes}): [64, 64]
    \item Projection channels (\texttt{projection\_channels}): 16
\end{itemize}

\subsection{DeepONet and Geometric DeepONet}
The hyperparameters for the standard DeepONet and its geometric extension are as follows:
\begin{itemize}
    \item Branch network layers (\texttt{branch\_net\_layers}): [512, 512, 512]
    \item Trunk network layers (\texttt{trunk\_net\_layers}): [256, 256, 256]
    \item Learning rate (\texttt{lr}): 0.0001
    \item Modes (\texttt{modes}): 128
\end{itemize}

\subsection{Wavelet Neural Operator}
The Wavelet Neural Operator (WNO) was trained with the following hyperparameters:
\begin{itemize}
    \item Input channels (\texttt{in\_channels}): 4
    \item Wavelet decomposition level (\texttt{level}): 4
    \item Learning rate (\texttt{lr}): 0.001
    \item Network width (\texttt{width}): 64
\end{itemize}

\subsection{scOT-T and poseidon-T}
The hyperparameters for the tiny versions of scOT and Poseidon are as follows:
\begin{itemize}
    \item Depths (\texttt{depths}): [4, 4, 4, 4]
    \item Embedding dimension (\texttt{embed\_dim}): 48
    \item Learning rate (\texttt{lr}): 0.0005
\end{itemize}

\subsection{scOT-B and poseidon-B}
The hyperparameters for the base versions of scOT and Poseidon are as follows:
\begin{itemize}
    \item Depths (\texttt{depths}): [8, 8, 8, 8]
    \item Embedding dimension (\texttt{embed\_dim}): 96
    \item Learning rate (\texttt{lr}): 0.0005
\end{itemize}

\subsection{scOT-L and poseidon-L}
The hyperparameters for the large versions of scOT and Poseidon are as follows:
\begin{itemize}
    \item Depths (\texttt{depths}): [8, 8, 8, 8]
    \item Embedding dimension (\texttt{embed\_dim}): 192
    \item Learning rate (\texttt{lr}): 0.0001
\end{itemize}

\section{Training and Validation Loss} \label{sec:loss-plots}

\begin{figure}[b!]
    \centering
    \begin{subfigure}[b]{0.48\textwidth}
        \centering
        \begin{tikzpicture}
        \begin{semilogyaxis}[
            width=0.9\textwidth,
            height=0.6\textwidth,
            xlabel={Epoch},
            ylabel={Loss},
            title={Poseidon-T},
            legend pos=north east,
            grid=major,
            xmin=0, xmax=200,
            ymin=1e-4, ymax=1
        ]
        \addplot[blue, thick] table[x=epoch, y=train_loss, col sep=comma] {figures/losses/poseidon-T.csv};
        \addlegendentry{Training}
        \addplot[orange, thick] table[x=epoch, y=val_loss_full, col sep=comma] {figures/losses/poseidon-T.csv};
        \addlegendentry{Validation}
        \end{semilogyaxis}
        \end{tikzpicture}
        \caption{Poseidon-T}
    \end{subfigure}
    \hfill
    \begin{subfigure}[b]{0.48\textwidth}
        \centering
        \begin{tikzpicture}
        \begin{semilogyaxis}[
            width=0.9\textwidth,
            height=0.6\textwidth,
            xlabel={Epoch},
            ylabel={Loss},
            title={scOT-T},
            legend pos=north east,
            grid=major,
            xmin=0, xmax=200,
            ymin=1e-4, ymax=1
        ]
        \addplot[blue, thick] table[x=epoch, y=train_loss, col sep=comma] {figures/losses/scot-T.csv};
        \addlegendentry{Training}
        \addplot[orange, thick] table[x=epoch, y=val_loss_full, col sep=comma] {figures/losses/scot-T.csv};
        \addlegendentry{Validation}
        \end{semilogyaxis}
        \end{tikzpicture}
        \caption{scOT-T}
    \end{subfigure}
    \\
    \begin{subfigure}[b]{0.48\textwidth}
        \centering
        \begin{tikzpicture}
        \begin{semilogyaxis}[
            width=0.9\textwidth,
            height=0.6\textwidth,
            xlabel={Epoch},
            ylabel={Loss},
            title={FNO},
            legend pos=north east,
            grid=major,
            xmin=0, xmax=200,
            ymin=1e-4, ymax=1
        ]
        \addplot[blue, thick] table[x=epoch, y=train_loss, col sep=comma] {figures/losses/fno.csv};
        \addlegendentry{Training}
        \addplot[orange, thick] table[x=epoch, y=val_loss_full, col sep=comma] {figures/losses/fno.csv};
        \addlegendentry{Validation}
        \end{semilogyaxis}
        \end{tikzpicture}
        \caption{FNO}
    \end{subfigure}
    \hfill
    \begin{subfigure}[b]{0.48\textwidth}
        \centering
        \begin{tikzpicture}
        \begin{semilogyaxis}[
            width=0.9\textwidth,
            height=0.6\textwidth,
            xlabel={Epoch},
            ylabel={Loss},
            title={DeepONet},
            legend pos=north east,
            grid=major,
            xmin=0, xmax=200,
            ymin=1e-4, ymax=1
        ]
        \addplot[blue, thick] table[x=epoch, y=train_loss, col sep=comma] {figures/losses/deeponet.csv};
        \addlegendentry{Training}
        \addplot[orange, thick] table[x=epoch, y=val_loss_full, col sep=comma] {figures/losses/deeponet.csv};
        \addlegendentry{Validation}
        \end{semilogyaxis}
        \end{tikzpicture}
        \caption{DeepONet}
    \end{subfigure}
    \caption{Training and validation loss (semi-log scale) for four models: Poseidon-T, scOT-T, FNO, and DeepONet.}
    \label{fig:loss-plots}
\end{figure}
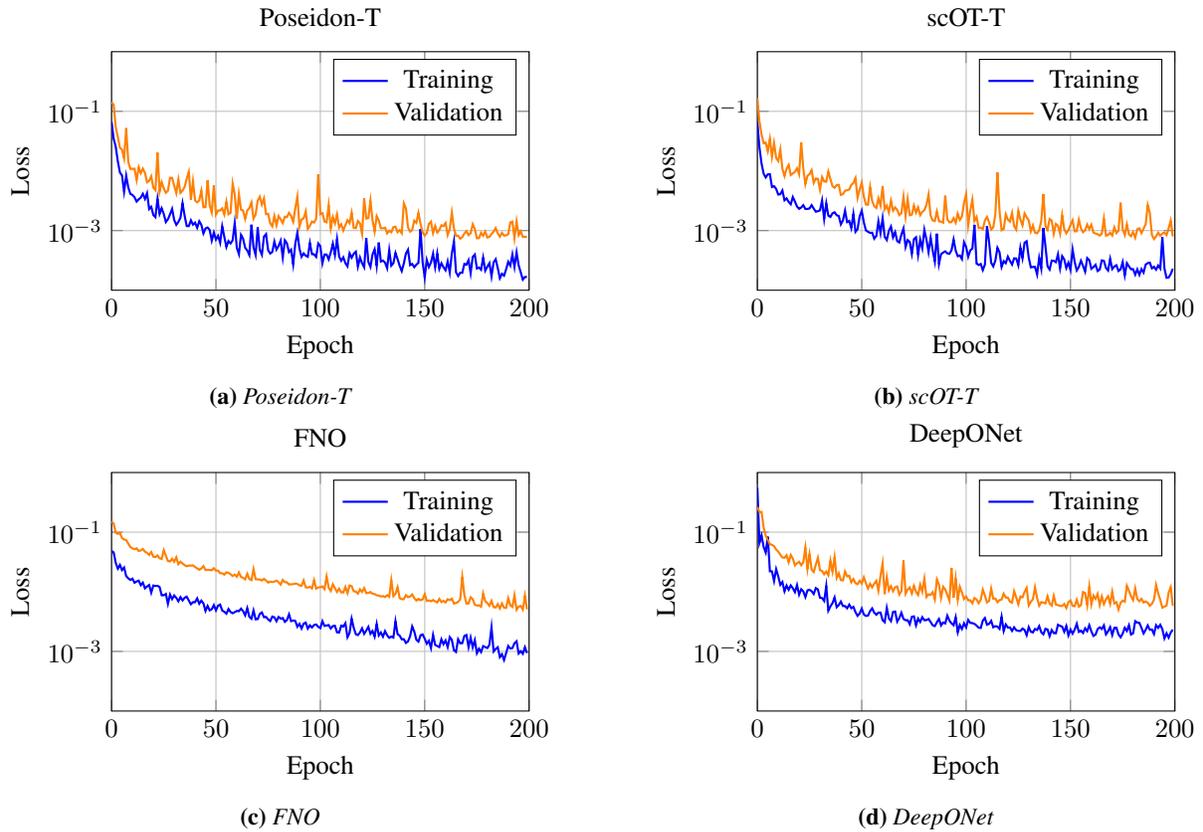

To further analyze the training performance, the evolution of training and validation loss for four representative models (Poseidon-T, scOT-T, FNO, and DeepONet) are shown in~\figref{fig:loss-plots}. 

\end{document}